\documentclass[dvipsnames,format=sigconf,nonacm]{acmart}

\usepackage{xcolor}
\usepackage[ruled,vlined,linesnumbered]{algorithm2e}
\DontPrintSemicolon
\SetAlFnt{\small}
\SetKwInOut{Input}{Input}
\SetKwInOut{Output}{Output}
\SetKw{KwAnd}{and}
\SetKw{KwOr}{or}
\SetKw{KwNot}{not}
\graphicspath{{figs/}}

\newcommand{\xD}[1]{$#1{\cdot}D$}

\setcitestyle{round,aysep={,},yysep={;}}

\begin{document}

\title{Adaptive Multi Mode: Budget-Aware Routing of Optimizers By Online Landscape Probes}

\author{Benjamin Jacob Bodner}
\affiliation{%
  \institution{}
  \city{}
  \country{}}

\begin{abstract}

The best classical optimizer for a continuous black-box problem depends on its budget and landscape. Existing routers decide from the dimension and budget, past improvement or features fitted to one benchmark, and few test an optimizer's implicit assumptions online. AMM (Adaptive Multi Mode) routes a known budget among quasi-Newton, trust-region, coordinate, multistart and evolution-strategy optimization engines with online probes of curvature, plateaus, separability, kinks, basins and fractal structure. Each probe has a default, a dimension-by-budget validity range and a cost charged to the budget. In its pre-registered development loop on bbob, 32 of 36 predictions held over the last 16 confirmations. AMM was then tested once on held-out bbob instances 101--110 ($D=2$--40, \xD{10}--\xD{1000}) against up to seven competitors. It has the highest target share in 382 of 600 function--dimension--budget cells (236 outright, 146 tied) and, by a descriptive posterior rule, is credibly behind in 35, mostly at \xD{10} and \xD{30}. At every budget it also leads the controls built from its own engines (the best engine alone, the best engine per dimension and budget, and a greedy schedule), narrowly at \xD{10} and \xD{30}, and nevergrad's NgIohTuned. In pre-registered runs its lead over IPOP-CMA-ES holds under affine and box transformations, on box-constrained bbob and, pooled, on rotated f2--f4, but rotating the separable f3 and f4 removes most of the lead there. On held-out MA-BBOB its final share at \xD{2000} is level with IPOP-CMA-ES's and its anytime area lower. We published AMM as an open-source Python package (\texttt{pip install amm-optimizer}; \url{https://github.com/BjBodner/amm-optimizer}).

\end{abstract}


\begin{teaserfigure}
  \centering
  \includegraphics[width=\textwidth,alt={Line chart of the share of targets reached against the evaluation budget, 10 to 1000 times the dimension, on bbob instances 101 to 110: AMM's thick black line above IPOP-CMA-ES, BIPOP-CMA-ES, SLSQP, NGOpt, DE, the 2025 MA-BBOB winner NADE and van Stein's 2024 algorithm ERADS, above HCMA's published results (dimensions 2 to 20), and above a single TPE point at 10 times the dimension.}]{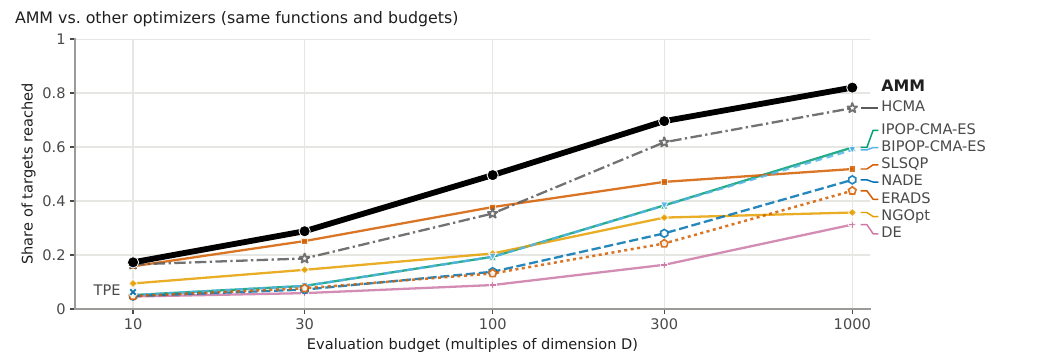}
  \caption{Held-out bbob instances 101--110 (f1--f24), equal budgets; field runs, SLSQP included, restart until the budget is spent or the final target is reached. Points: share of 51 targets reached per run, averaged over 10 instances and 120 function--dimension cells ($D=2$--40). HCMA (published COCO archive, own instances): $D=2$--20 only; there AMM is level at \xD{10} ($+.016$) and leads by $+.10$ to $+.17$ from \xD{30}. TPE ran every $D$ only at \xD{10}. NADE (2025 MA-BBOB winner) and ERADS (van Stein's 2024 algorithm): authors' code, run after the final test. Against the pre-registered field AMM is highest in 382 of 600 function--dimension--budget cells (236 outright, 146 tied; 380, 234 outright, with NADE and ERADS added) and leads IPOP-CMA-ES by $+.121$ to $+.312$.}
  \Description[Share of targets reached against the budget, AMM and other optimizers (final test)]{A line chart. The x-axis is the evaluation budget in multiples of the dimension D, on a logarithmic scale from 10 to 1000; the y-axis is the share of targets reached, from 0 to 1, where a target is a precision f minus f_opt between 10 to the 2 and 10 to the minus 8 (51 targets). It compares AMM with IPOP-CMA-ES, BIPOP-CMA-ES, SLSQP (restarted until the budget is spent, like every method of the pre-registered field), NGOpt and differential evolution, averaged over dimensions 2 to 40, each line labelled at its right end; with HCMA's published COCO archive results, drawn grey and dash-dotted with hollow stars, averaged over dimensions 2 to 20 only; with TPE as a single point at the smallest budget, the only one at which it ran every dimension; and with NADE, the 2025 MA-BBOB winner (blue, dashed), and ERADS, van Stein's 2024 algorithm (vermillion, dotted), run from their authors' code after the final test, both with hollow markers. AMM's thick black line is the highest at every budget, rising from .173 to .820; SLSQP is close to it at the smallest budget and reaches .518 at the largest; the IPOP- and BIPOP-CMA-ES lines lie on top of each other, IPOP-CMA-ES rising from .052 to .598; NADE and ERADS start at .048 and end at .478 and .438, close to IPOP-CMA-ES at the two smallest budgets and below it from 100 times the dimension; HCMA's line starts level with AMM's and stays below it. On the same dimension-2-to-20 cells AMM is level with HCMA at the smallest budget (+.016) and leads it by +.10 to +.17 from 30 times the dimension. Against the pre-registered field AMM is highest in 382 of the 600 function, dimension and budget cells (236 outright, 146 tied), and in 380 (234 outright) with NADE and ERADS added; its lead over IPOP-CMA-ES is +.121, +.203, +.303, +.312 and +.222 from the smallest to the largest budget.}
  \label{fig:teaser}
\end{teaserfigure}

\maketitle

\section{Introduction}\label{sec:intro}
Which classical algorithm is best for a continuous black-box problem depends on how many evaluations it may spend \cite{hansen2010comparing}. Around \xD{10}, model-based trust-region methods such as COBYLA lead \cite{raponi2023optimizing}. On smooth functions, quasi-Newton methods with finite-difference gradients converge fast \cite{varelas2019benchmarking,brockhoff2025benchmarking}. With thousands of evaluations per dimension, restarted CMA-ES with growing populations finds global structure that local methods miss \cite{hansen2009benchmarking,auger2005restart}. A user often knows the budget before the run; the landscape, which moves these bands, is unknown.

Existing routers decide without testing $f$, or test it only through generic features fitted to a training suite. Wizards such as Nevergrad's NGOpt follow rules on dimension, budget and variable types \cite{meunier2020black}, so at a given $D$ and $B$ every function gets the same chain. Races such as HCMA's keep the engine that improves and pay for the loser \cite{loshchilov2013bi}. Offline schedules fix one budget split for all problems \cite{schapermeier2025greedy}, and selectors built on exploratory landscape analysis (ELA) fall back towards the single best solver on held-out problems \cite{petelin2025pitfalls}.

AMM (Adaptive Multi Mode) decides by testing $f$. An \emph{online probe} is a short test, paid from the budget, of whether the assumption behind one classical engine holds near the incumbent. A mirrored loss that grows quadratically with the step calls for a quasi-Newton method, additivity over coordinate pairs for a coordinate search, converged descents in separate basins for restarts. A probe that does not fire leaves the run on a general path, so a misreading costs evaluations, not the run. Every rule reads only $D$, $B$, the box and evaluated $f$-values, as COCO (Comparing Continuous Optimizers) requires \cite{hansen2021coco,hansen2016coco}.

None of these statistics is new, and Section~\ref{sec:related} credits each one's ancestor. To our knowledge, AMM is the first black-box optimizer that routes a known evaluation budget among quasi-Newton, trust-region, coordinate, population and multistart solvers through a decision list of cheap mechanism tests, each a known statistic with a default outcome, a calibrated dimension-by-budget validity range and a cost charged to the budget.

A rule set developed on the 24 bbob functions invites the objection that it is a decision tree fitted to them \cite{meunier2020black,bartzbeielstein2020benchmarking}. In its final form the loop kept a fix only if a prediction written before its run held on unseen instances, and the held-out tests measure robustness (tested regime in Section~\ref{sec:heldout}, limits in Section~\ref{sec:limits}) to new instances, new affine combinations and relabelled, rotated and box-constrained versions of the 24 functions and to unseen dimensions and budget multipliers, not to an unseen function (an unregistered check on 20 other functions is in Appendix~\ref{app:fresh}).

\paragraph{Contributions.} The claims concern one configuration of AMM, run once on bbob instances 101--110, which no design decision used. The paper makes four contributions.
\begin{itemize}
\item AMM, a router whose online probes test an engine's assumption before the engine runs (Section~\ref{sec:method}).
\item Held-out evidence that from \xD{100} up its routing, not only its engine pool, carries its pooled lead, though not in every cell (Section~\ref{sec:res-losses}). AMM has the highest target share among up to seven competitors (Section~\ref{sec:baselines}) in 382 of 600 function--dimension--budget cells and is credibly behind in 35 (Figure~\ref{fig:teaser}; Sections~\ref{sec:res-final} and~\ref{sec:res-losses}). Against a ladder of controls that separates the routing from the engine pool, every rung except a post-hoc oracle trails AMM at every budget, its best engine alone, a per-dimension, per-budget selector and a greedy schedule only narrowly at \xD{10} and \xD{30}, and AMM exceeds the oracle only at \xD{1000} (Section~\ref{sec:res-ladder}); a leave-one-route-out ablation shows every route family paying on its own slice, and route-firing logs show where each route fires (Section~\ref{sec:res-parts}).
\item Pre-registered tests of how far the lead carries, losses named (Sections~\ref{sec:res-general} and~\ref{sec:res-losses}): the lead over IPOP-CMA-ES holds under affine and box transformations, on the box-constrained suite and, pooled, on rotated f2--f4, but rotating f3 and f4 removes most of the lead on those two functions (on rotated f4 it is level), and on MA-BBOB AMM is level with IPOP-CMA-ES in final share at \xD{2000} and behind in anytime area.
\item A development loop that can say no: in its last 16 confirmations 32 of 36 pre-registered predictions held, and no failed fix was promoted (Section~\ref{sec:loop}).
\end{itemize}
AMM is an open-source Python package (Software availability).

\section{Related work}\label{sec:related}

\begin{table*}[t]
  \caption{How AMM differs from the closest routing approaches (from the cited sources; Section~\ref{sec:related}).}
  \Description[Positioning table: eight methods compared on four properties]{An eight-row table. For BIPOP-CMA-ES, HCMA, SLSQP+lq-CMA-ES, Nevergrad's wizards, greedy restart schedules, per-run algorithm selection, LLaMEA and AMM it lists what the method routes on, when it decides, what it reads about the objective before choosing an engine, and the benchmark it was evaluated on. Only AMM reads mechanism probes, with default outcomes and costs charged to the budget, before each engine runs.}
  \label{tab:position}
  \scriptsize
  \begin{tabular}{@{}l p{0.19\textwidth} p{0.16\textwidth} p{0.22\textwidth} p{0.19\textwidth}@{}}
    \toprule
    Method & Routes on & Decides & Reads about $f$ before choosing & Evaluated on \\
    \midrule
    BIPOP-CMA-ES & two fixed restart regimes & at each restart & nothing & bbob (archive) \\
    HCMA & a fixed opening, then the engine ahead & NEWUOA for \xD{10}, then a race & which engine improved & bbob (archive, 2013) \\
    SLSQP+lq-CMA-ES & a fixed sequence & fixed hand-over & nothing & bbob (archive, 2020) \\
    NGOpt39, NgIohTuned & $D$, $B$, bounds, noise & before the run & nothing & Nevergrad suites \\
    Greedy restart schedules & training-set statistics per $D$ & before the run & nothing & bbob, $D\le10$ \\
    Per-run selection & features and a trained model & after a \xD{30} CMA-ES run & generic landscape features & bbob; YABBOB \\
    LLaMEA & a generated program & at design time & nothing at run time & bbob 5-D (anytime area) \\
    AMM & $D$, $B$ and online probes & before each engine & mechanism probes with defaults, paid from $B$ & bbob 101--110; MA-BBOB rows 900--999; transformed, rotated and bbob-boxed bbob, ids 810--849 \\
    \bottomrule
  \end{tabular}
\end{table*}

Table~\ref{tab:position} contrasts AMM with the closest routing approaches. Single classical solvers and restarted or surrogate-assisted CMA-ES variants each own a budget band (Appendix~\ref{app:related}); AMM uses such solvers unchanged and claims only a better allocation of a known budget.

\paragraph{Fixed hand-overs and races.} HCMA opens with NEWUOA for \xD{10}, then races STEP against surrogate-assisted BIPOP-CMA-ES \cite{loshchilov2013bi}. \citet{hansen2019global} compared lq-CMA-ES with a portfolio that runs SLSQP (sequential least-squares quadratic programming) before CMA-ES; AMM's prefix (the route for budgets up to \xD{100}, Section~\ref{sec:routing}), which picks restarted SLSQP or, on a rugged reading, an evolution strategy (at \xD{100}, surrogate-ranked populations), descends from it. COCOpf, MOS and EGO-CMA switch or chain engines by realised improvement or a fixed rule \cite{baudis2014cocopf,latorre2010mos,mohammadi2015making}. So do memetic schemes such as MTS and SHADE-ILS, whose choice between a coordinate search and L-BFGS-B is closest to AMM's \cite{linyutseng2008multiple,molina2018shade}. All decide by schedule or by which engine improved.

\paragraph{A-priori wizards.} Shiwa and NGOpt select by hand-written rules on problem metadata \cite{liu2020versatile,meunier2020black}. Nevergrad's current wizards (NGOpt39, the NgIoh family, NgIohTuned and Wiz) refine them into finer rules on $D$ and $B$, with Cobyla at small budgets and a fixed Cobyla, MetaModel and SQP chain at middle budgets \cite{rapin2018nevergrad}. They share AMM's inputs and much of its pool.

\paragraph{Schedules, selection and switching.} Greedy restart schedules and sequential portfolios learn one budget split from bbob data \cite{schapermeier2025greedy,dinu2026how}. The greedy schedule closed 95.6\% of the gap between the single and the virtual best solver across instances, but about 27\% when whole problems were held out \cite{schapermeier2025greedy}. Exploratory landscape analysis reads $f$ through generic features \cite{mersmann2011exploratory} for cost-sensitive \cite{bischl2012algorithm} and fixed-budget \cite{jankovic2020landscape} selection, and probing trajectories select from short runs of the candidate solvers \cite{renau2024utility}. A prototype chooses its operators, not engines, online from a separability analysis \cite{epitropakis2014separability}, and dynamic algorithm configuration learns policies that change a configuration during a run \cite{adriaensen2022automated}. Feature-based, warm-started, learned and image-based selectors and switchers \cite{kerschke2019automated,tanabe2021benchmarking,vermetten2020dynamic,kostovska2022per,vermetten2023switch,schelling2026reinforcement,wang2026geometric} lose most of their advantage under leave-problem-out validation and on MA-BBOB \cite{petelin2025pitfalls,cenikj2025landscape,dietrich2024impact}. \citet{tanabe2021benchmarking} found that SLSQP as a pre-solver significantly improved such selectors, as in AMM's prefix.

\paragraph{Probes inside optimizers.} The curvature probe reads a mirrored second difference, as in the finite-difference intervals of \citet{gill1983computing} and in HE-ES \cite{glasmachers2020hessian}, and the plateau probe pre-empts CMA-ES's flat-fitness restart \cite{hansen2016cma}. The pair test samples the additivity checks of LINC-R and differential grouping \cite{tezuka2004linkage,omidvar2014cooperative} and SPAM's separability index \cite{caraffini2014analysis}; the hill-valley test comes from \citet{ursem1999multinational} and HillVallEA \cite{maree2018real}. The kink probe relates to finite-difference noise estimation \cite{more2012estimating}, and the fractal probe's slope comes from \citet{wang2023fractal}, who in policy optimization read a log-log slope below 2 as non-differentiability. AMM contributes the routing semantics, not the statistics; its fractal route to restarted SLSQP runs against that reading and rests on its own confirmation.

\paragraph{Automated and LLM design.} AMM's rules, proposed by AI coding-agent sessions, are a form of automated design, as in LLaMEA \cite{stein2024llamea} and the related work of Appendix~\ref{app:related}, but a supervised one: the author approved every pre-registration and every promotion of a fix into the method (Section~\ref{sec:loop}). Accordingly, AMM reports its design budget, audits its inputs and meets the 2025 MA-BBOB winner in the competition's setting on held-out public rows \cite{vermetten2023ma,vermetten2023mab,vermetten2025gecco,stein2025neighborhood}.

\paragraph{What AMM changes.} Unlike these routers, AMM reads $f$ \emph{before} an engine runs and maps each reading to an engine by a stated mechanism. Its thresholds and validity ranges were hand-calibrated on bbob f1--f24 at the development dimensions, not learned by a model fitted to a training suite; Sections~\ref{sec:res-ladder} and~\ref{sec:res-losses} test whether this pays and where earlier routers are stronger.

\section{AMM: routing by online probes}\label{sec:method}
\begin{figure*}[t]
  \centering
  \includegraphics[width=\textwidth,alt={Flowchart of AMM: three whole-budget gates, a budget gate, the prefix, the ATM stage, local probes, eight engines and the finisher.}]{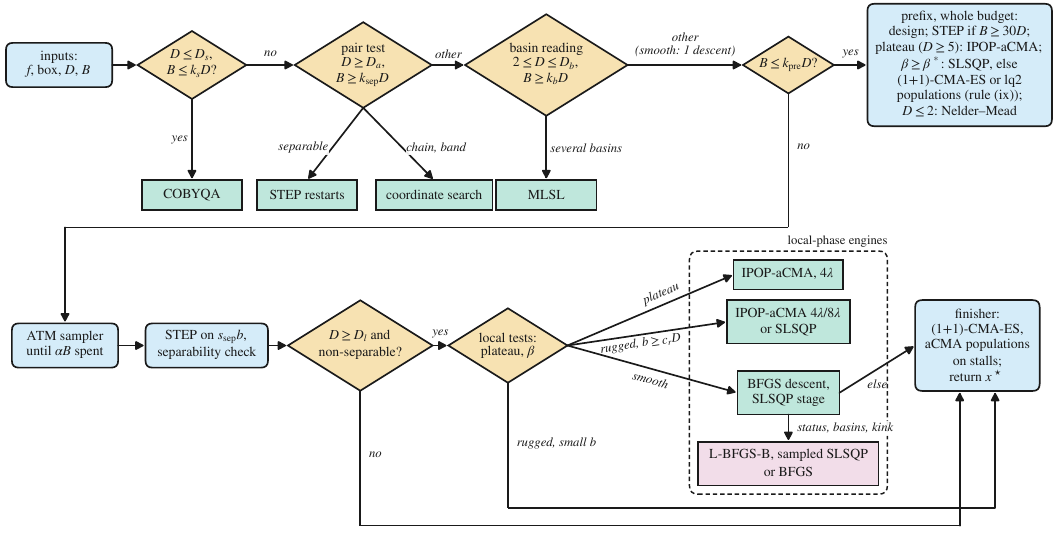}
  \caption{Routing of AMM (simplified). Rounded boxes are phases, diamonds are probes or gates, square boxes are engines; a dashed box groups the engines that spend the rest of the budget. Guards use the symbols of Table~\ref{tab:params}; ``lq2 populations'' are populations ranked by a quadratic surrogate. Not shown: the prefix's single SLSQP descents (at $D\le2$ up to \xD{30}; below $\min(k_{\mathrm{des}}D,200)$), its COBYQA hand-overs and the conditions of its lq2 populations (Algorithm~\ref{alg:prefix}); and, on a rugged reading with $b<c_rD$, the capped BFGS try of \xD{100} at $D\le9$, $B\le500D$ and $b\ge200D$, which rule (vii) replaces at $0.5\le\beta<1$ by an immediate start of the finisher's populations and rule (vi) continues to the end of the budget at a kink (Algorithm~\ref{alg:main}, lines~\ref{ln:direct}--\ref{ln:finish}).}
  \Description[Flowchart: whole-budget gates, a budget gate, the prefix, the ATM stage, local probes and engines]{A left-to-right flowchart in two rows. Top row: the inputs, then three gates in sequence (small dimension and budget; additivity pair test; low-dimensional basin reading), each with a branch down to an engine that spends the whole budget (COBYQA; STEP restarts or a decoupled coordinate search; MLSL), then a budget gate. The budget gate's yes branch leads to the prefix, a box that runs a design, STEP, a plateau probe that chooses IPOP-aCMA-ES populations and a curvature probe that chooses SLSQP, a (1+1)-CMA-ES or lq2 populations. Its no branch leads along the bottom row to the ATM sampler, a STEP hand-over with a separability check, a diamond of local probes (plateau, curvature, then exit status, basins and kink), a dashed group of engines (IPOP populations, restarted SLSQP, L-BFGS-B, sampled SLSQP restarts, scaled BFGS) and the finisher, a (1+1)-CMA-ES that hands stalls to aCMA-ES populations. All paths end at the incumbent.}
  \label{fig:overview}
\end{figure*}

Figure~\ref{fig:overview} shows the routing, Algorithms~\ref{alg:main} and~\ref{alg:prefix} its control loop, Table~\ref{tab:params} every gate and probe and Table~\ref{tab:rules} the nine rules added by composites c44--c46. The benchmarked configuration is the code's preset \texttt{c46}, which sets 67 of the 160 configuration fields away from their defaults (Appendix~\ref{app:codemap}); it is the last \emph{composite}, a numbered configuration of AMM from the development loop (Section~\ref{sec:loop}).

\paragraph{Inputs and compliance.} One configuration serves all functions. AMM reads only the dimension $D$, the budget $B$, the box $[\ell,u]$ and the $f$-values of points it evaluates, never the function or instance identifier, the optimal value or any other property of the problem. Its thresholds depend on $D$ and $B$ only, as COCO allows. Every evaluation, probes included, counts against $B$, and every evaluated point lies in the box. The benchmark harness enforces $B$; the package does not (Software availability). A code audit confirms that the entry point receives only the objective, a start point, the budget, the box and a seed (or a random generator), with the configuration and a profiling switch as options, and that no module reads anything else (Appendix~\ref{app:codemap}).

\SetKwFunction{Run}{Run}
\SetKwFunction{PairTest}{PairTest}
\SetKwFunction{BandCheck}{BandCheck}
\SetKwFunction{BasinReading}{BasinReading}
\SetKwFunction{Prefix}{Prefix}
\SetKwFunction{Plateau}{Plateau}
\SetKwFunction{Curvature}{Curvature}
\SetKwFunction{Kink}{Kink}
\SetKwFunction{LocalTests}{LocalTests}
\SetKwFunction{Finisher}{Finisher}

\SetAlFnt{\footnotesize}
\begin{algorithm}[t]
\caption{Control loop of AMM, preset \texttt{c46}. \textsf{Run}(E, $x$) runs engine E from $x$ until it converges, stalls or spends its share (unless stated, all of $b$, the budget left); the incumbent $x^\star$ never gets worse. Roman numerals: rules in Table~\ref{tab:rules}; (viii) first maps the box onto $[-5,5]^D$. \textsf{Prefix}: Algorithm~\ref{alg:prefix}; probes, \textsf{LocalTests}, \textsf{Finisher}: Appendix~\ref{app:routing}; symbols: Table~\ref{tab:params}.}
\label{alg:main}
\Input{objective $f$; box $[\ell,u]\subset\mathbb{R}^D$; budget $B$; start point $x_0$}
$x^\star\gets x_0$; $b\gets B$\;
\If{$D\le D_s$ \KwAnd $B\le k_sD$\nllabel{ln:cobyqa}}{\KwRet{\Run{\textsc{Cobyqa}, $x^\star$}}\;}
\If{$D\ge D_a$ \KwAnd $B\ge k_{\mathrm{sep}}D$}{
  $r\gets$ \PairTest{$x_0$}\nllabel{ln:pair}\tcp*{$3D$ evaluations}
  \lIf{$r=$ separable}{\KwRet{\Run{\textsc{Step} restarts, $x^\star$}}}
  \lIf{$r=$ chain \KwAnd $B\ge k_{\mathrm{ch}}D$ \KwAnd \BandCheck{$x_0$} $<\tau_{\mathrm{band}}$}{\KwRet{\Run{\textsc{CoordSearch}, $x^\star$}}}
}
\If{$2\le D\le D_b$ \KwAnd $B\ge k_bD$}{
  $(m,s,x^\star)\gets$ \BasinReading{}; $x_0\gets x^\star$\nllabel{ln:basin}\tcp*{Sobol design}
  \lIf{$m$}{\KwRet{\Run{\textsc{Mlsl}, $x^\star$}}}
  \lIf{$s$ \KwAnd $B\le k_dD$}{$x^\star\gets$ \Run{\textsc{Slsqp} descent, $x_0$}\nllabel{ln:basin2}\tcp*[f]{not in $\alpha B$}}
}
\lIf{$B\le k_{\mathrm{pre}}D$}{\KwRet{\Prefix{$x_0$}}\nllabel{ln:prefix}\tcp*[f]{Algorithm~\ref{alg:prefix}}}
$x^\star\gets$ \Run{\textsc{Atm}, $x_0$} until $\alpha B$ are spent (at most $c_AD$ if $D\ge10$)\nllabel{ln:atm}\;
$(x^\star,\mathit{sep})\gets$ \Run{\textsc{Step}, $x^\star$} on $s_{\mathrm{sep}}b$, with its separability check\nllabel{ln:step}\;
\If{$D\ge D_l$ \KwAnd \KwNot $\mathit{sep}$}{
  $r\gets$ \LocalTests{$x^\star$}\nllabel{ln:local}\tcp*{plateau, then $\beta$}
  \Switch{$r$}{
    \Case{plateau}{\KwRet{\Run{IPOP-aCMA $4\lambda$, uniform point}}\;}
    \Case{rugged \KwAnd $b\ge c_rD$}{\leIf{$D\ge20$ \KwAnd fractal}{\KwRet{\Run{\textsc{Slsqp} restarts}}}{\KwRet{\Run{IPOP-aCMA $4\lambda$ or $8\lambda$}}}}
    \Case{smooth}{
      \lIf{$B\le k_{\mathrm{bf}}D$}{$x^\star\gets$ \Run{\textsc{Bfgs}, one descent}}
      $x^\star\gets$ \Run{\textsc{Slsqp} stage, $\min(b/3,c_qD)$, stall monitor at the cap}\;
      \lIf{exit status, basins or kink fire}{\KwRet{\Run{\textsc{L-Bfgs-B} $\mid$ sampled \textsc{Slsqp} $\mid$ \textsc{Bfgs}}}\nllabel{ln:early}}
    }
    \Other{
      \lIf{$0.5\le\beta<1$}{$\mathit{direct}\gets(b\ge gD)$\nllabel{ln:direct}\tcp*[f]{(vii)}}
      \lElseIf{$D\le D_{\mathrm{try}}$ \KwAnd $B\le k_{\mathrm{rt}}D$ \KwAnd $b\ge2k_{\mathrm{try}}D$}{$x^\star\gets$ \Run{\textsc{Bfgs}, $k_{\mathrm{try}}D$; on to the end at a \Kink{}}\nllabel{ln:try}\tcp*[f]{(vi)}}
    }
  }
}
\KwRet{\Finisher{$x^\star$}}\nllabel{ln:finish}\tcp*{(1+1)-CMA-ES; aCMA on stalls or if $\mathit{direct}$}
\end{algorithm}

\begin{algorithm}[t]
\caption{\textsf{Prefix} of preset \texttt{c46} (Algorithm~\ref{alg:main}, line~\ref{ln:prefix}), which spends the whole budget at $B\le k_{\mathrm{pre}}D$. Roman numerals as in Algorithm~\ref{alg:main}.}
\label{alg:prefix}
$x^\star\gets$ best of $x_0$ and $\max(4,D)$ uniform points; from \xD{30}, $(x^\star,\mathit{sep})\gets$ \Run{\textsc{Step}, $x^\star$} on $b/2$\nllabel{ln:pdesign}\;
\lIf{$D\le2$}{up to \xD{30}, $x^\star\gets$ one \textsc{Slsqp} descent (v); \KwRet{\Run{Nelder--Mead restarts, $x^\star$}}}
\If{$B<\min(k_{\mathrm{des}}D,200)$}{
  $x^\star\gets$ \Run{\textsc{Slsqp}, one descent}\tcp*{retried (v) or read by (ii)}
  \KwRet{\Run{(1+1)-CMA-ES, $x^\star$}}\tcp*{COBYQA under (ii)}
}
\lIf{$D\ge5$ \KwAnd $B>30D$ \KwAnd \Plateau{$x^\star$}}{\KwRet{\Run{IPOP-aCMA, $x^\star$}}\nllabel{ln:pplat}\tcp*[f]{(i)}}
$\beta\gets$ \Curvature{$x^\star$}, smooth if undefined unless \Plateau{$x^\star$}\tcp*{(iv)}
\lIf{$\beta\ge\beta^\ast$}{\KwRet{\Run{\textsc{Slsqp} restarts, $x^\star$; \textsc{L-Bfgs-B} after a failed majority}}\tcp*[f]{COBYQA: (ii), (iii)}}
\lIf{$B=k_{\mathrm{pre}}D$ \KwAnd \KwNot $\mathit{sep}$ \KwAnd ($D<10$ \KwOr $\beta\ge0.15$ \KwOr $\beta$ undefined)}{\KwRet{\Run{lq2 populations, $x^\star$}}\tcp*[f]{(ix)}}
\KwRet{\Run{(1+1)-CMA-ES, $x^\star$}}\nllabel{ln:pcma}\tcp*{COBYQA under (ii)}
\end{algorithm}

\subsection{Routing by dimension and budget}\label{sec:routing}
Algorithm~\ref{alg:main} first applies three \emph{whole-budget gates} (lines~\ref{ln:cobyqa}--\ref{ln:basin2}), which can give the whole budget to one engine before any general search. Restarted COBYQA takes it at $D\le3$ and $B\le10D$ (line~\ref{ln:cobyqa}), where no probe pays for itself. From $D=5$ and \xD{100}, the pair test at the start point (line~\ref{ln:pair}) sends a separable reading to STEP (select the easiest point) restarts, the \emph{separable route}, and, from \xD{300}, a coupling of only adjacent coordinates that the band check confirms to a decoupled coordinate search, the \emph{chain route}. At $D=2$--3 from \xD{100}, a basin reading on a Sobol design (line~\ref{ln:basin}) sends several smooth basins to MLSL (multi-level single linkage) and, up to \xD{300}, a single smooth basin to one SLSQP descent first (line~\ref{ln:basin2}). A gate that does not fire costs only its evaluations, and its best point stays a candidate for the result; the run continues from the start point or, after a basin reading, from the design's best point, also when the descent of line~\ref{ln:basin2} ran.

Up to \xD{100} the \emph{prefix} spends the whole budget (line~\ref{ln:prefix}; Algorithm~\ref{alg:prefix}) on a design, STEP and the curvature probe, which picks restarted SLSQP or a (1+1)-CMA-ES; from $D=5$ above \xD{30} a plateau reading sends the rest to IPOP-aCMA-ES populations, and up to \xD{30} two rules can hand it to restarted COBYQA (Table~\ref{tab:rules}, rules (i)--(iii)). Above \xD{100}, the ATM sampler, a population-based random search (Section~\ref{sec:engines}), takes a fifth of the budget (at most \xD{60} from $D=10$; line~\ref{ln:atm}), and STEP, with a separability check, half of the rest (line~\ref{ln:step}). If the check fails, from $D=5$ the local probes (line~\ref{ln:local}) send plateau and rugged readings to IPOP-aCMA-ES populations, smooth readings to an SLSQP stage and, from $D=20$, fractal readings to restarted SLSQP. Exit-status, basin and kink readings can end that stage in favour of L-BFGS-B, sampled SLSQP restarts or a scaled finite-difference BFGS (line~\ref{ln:early}). Two of these routes appear unpublished: the kink reading to the scaled BFGS, and the fractal reading to restarted SLSQP, whose ablation effect is null (Section~\ref{sec:res-parts}). A run that no route has ended goes to the \emph{finisher} (line~\ref{ln:finish}), a restarted (1+1)-CMA-ES that hands stalls to aCMA-ES populations; Appendix~\ref{app:routing} gives every rule.

\def\paramcols{@{}p{0.15\textwidth} p{0.20\textwidth} p{0.06\textwidth} p{0.08\textwidth} p{0.10\textwidth} p{0.13\textwidth} p{0.13\textwidth}@{}}
\begin{table*}[t]
  \caption{Gates and probes of AMM (one setting for all functions; values as benchmarked). Columns: the thresholds, with the symbols of Figure~\ref{fig:overview} and Algorithms~\ref{alg:main} and~\ref{alg:prefix}; what a positive reading selects; cost in evaluations (charged to $B$); default; where it acts ($b$ is the budget left); the function and composite (cNN) whose confirmation introduced it (Section~\ref{sec:built}; Table~\ref{tab:confirm} for c32--c46, Appendix~\ref{app:build} for the earlier composites; rows marked development or calibration rest on development data). The last column lists the functions where the route fires in at least half of the runs in its range on the final test's grid (``none'' with the largest shares where none does; A2: where a replication on fresh ids 810--829 differs; first call of each run).}
  \Description[Table of gates and probes: thresholds, routes, costs, defaults, ranges, origins and where each fires]{A seven-column table with one row per gate or probe of AMM, in the order in which a run meets them: the COBYQA gate, the pair test and the chain reading, the basin reading and the basin descent, the prefix, the curvature probe, the ATM stage and STEP check, the plateau probe, the rugged and fractal routes, the smooth-route descent and SLSQP stage, the exit-status and hill-valley readings, the kink probe, the rugged BFGS try and the finisher. For each it gives the thresholds with their symbols, the reading and the engine it selects, the cost in evaluations, the default path, the dimension and budget range, the function and composite that motivated it, and the functions on which it fires in the final test's route-firing logs.}
  \label{tab:params}
  \scriptsize
  \begin{tabular}{\paramcols}
    \toprule
    Thresholds & Reading $\to$ route & Cost & Default & Acts at & Motivated by & Fires on \\
    \midrule
    $D_s{=}3$, $k_s{=}10$ & COBYQA gate: the whole budget to restarted COBYQA & -- & next gate & $D\le3$, $B\le10D$ & small budgets, $D=2$--3 (c43) & f1--f24 \\
    $D_a{=}5$, $k_{\mathrm{sep}}{=}100$, $\tau_{\mathrm{add}}{=}10^{-6}$ & pair test: additive over pairs $\to$ STEP restarts & $3D$ & next gate & $D\ge5$, $B\ge100D$ & f3 (c38) & f1--f5 \\
    $k_{\mathrm{ch}}{=}300$, $\tau_{\mathrm{band}}{=}0.02$ & only adjacent pairs interact, band check $\to$ coordinate search & $6D$ & next gate & $D\ge5$, $B\ge300D$ & f20 (c38) & f20 \\
    $D_b{=}3$, $k_b{=}100$, $\tau_{r2}{=}0.5$, $\tau_{\mathrm{lc}}{=}0.1$ & basin reading: several basins $\to$ MLSL & design $+4D$ & design's best point & $D=2$--3, $B\ge100D$ & f21, f22 (c39) & none (f21 .40, f22 .30) \\
    $k_d{=}300$ & one smooth basin $\to$ one SLSQP descent first & -- & -- & $D=2$--3, $B\le300D$ & f6, f8, f9, f12, f13 (c42) & f24; A2: none \\
    $k_{\mathrm{pre}}{=}100$, $k_{\mathrm{des}}{=}31$ & prefix spends the whole budget; below $\min(31D,200)$ one descent, no probe & -- & -- & $B\le100D$ & \xD{30}--\xD{100} (c36); \xD{10} (c40) & f1--f24 \\
    $\beta^\ast{=}1.45$ & curvature: $\beta\ge1.45$ $\to$ SLSQP (prefix) or the smooth route & 24 (12 below \xD{30}) & (1+1)-CMA-ES or populations & prefix; $D\ge5$ & f1, f2, f6, f8--f14, f20--f22 vs f7, f15--f19, f23, f24 (calibration) & f6, f8--f14, f21--f22 \\
    $\alpha{=}0.2$, $c_A{=}60$, $s_{\mathrm{sep}}{=}0.5$, $k_{\mathrm{st}}{=}1000$ & ATM stage, then STEP with a separability check (two-axis from $1000D$) & re-evaluations & local probes & $B>100D$ & development; $D=5$, \xD{3000} (c39) & f6--f19, f21--f24 \\
    $D_l{=}5$, $h_{\mathrm{tie}}{=}3{\cdot}10^{-4}$ & plateau: ties $\to$ IPOP-aCMA-ES, $4\lambda$ & 4 & curvature & $D\ge5$, non-separable & f7 (c28) & f7 \\
    $c_r{=}640$ / $2000$, $c_8{=}2500$, $\tau_{\mathrm{fr}}{=}1.5$ & rugged with $b\ge640D$ ($D\ge20$) or $2000D$ ($D=5$--19) $\to$ IPOP ($8\lambda$ from $D=20$ with $b\ge2500D$); fractal reading $\to$ restarted SLSQP & 32 (fractal) & finisher & $D\ge5$; fractal $D\ge20$ & f17, f18 (c33); f15 (c37); f23 (c36) & f15--f19, f23--f24 \\
    $k_{\mathrm{bf}}{=}300$ & smooth reading $\to$ one scaled BFGS descent & -- & SLSQP stage & $D\ge5$, $B\le300D$ & f6, f12--f14 (c41) & f6, f8--f14, f21--f22 \\
    $c_q{=}100$, $h_{\mathrm{fd}}{=}10^{-9}$, $c_2{=}400$ & SLSQP stage of $\min(b/3,100D)$; stall monitor past the cap; second descent if $400D$ left & 0 & finisher & smooth route & f10, f11, f14 (development); f8, f9 (c34) & f6, f8--f12, f21--f22; A2: f6, f8--f14, f21--f22 \\
    -- & exit status: failed majority $\to$ L-BFGS-B; hill-valley: separate basins $\to$ sampled SLSQP restarts & 0; $\le9$ (3 points, up to 3 pairs) & finisher & smooth route (exit status also in the prefix's SLSQP restarts) & f12 (c32); f21, f22 (c35) & f21--f22 \\
    $\tau_k{=}5{\cdot}10^{7}$, $c_k{=}100$ & kink: second difference at $10^{-7}$ of the width $\ge5{\cdot}10^{7}$ $\to$ scaled BFGS & 6 & finisher & stalled stage, $b\ge100D$ & f13 (c37) & none (f13 .36) \\
    $k_{\mathrm{try}}{=}100$, $D_{\mathrm{try}}{=}9$, $k_{\mathrm{rt}}{=}500$ & rugged reading $\to$ one BFGS try of $100D$ & -- & finisher & $D\le9$, $B\le500D$, $b\ge200D$ & f13, $D=5$ (c37) & f13, f15, f18, f24; A2: f13, f15, f17, f19, f24 \\
    $g{=}100$, $c_{\mathrm{run}}{=}150$, $k_{\mathrm{pr}}{=}30$, $\beta_{\mathrm{in}}{=}1.3$, $D_\beta{=}5$, $\epsilon_f{=}10^{-11}$ & finisher: (1+1)-CMA-ES; aCMA populations on a stall with $100D$ left, after a run of $150D$ ($D\ge10$) or if $\beta<1.3$ at $30D$ ($D\ge5$) & $\beta$ probe & -- & all & f17, f18 (development) & not logged \\
    \bottomrule
  \end{tabular}
\end{table*}

\subsection{Probes}\label{sec:tests}
The \emph{curvature} probe is the slope $\beta$ of the logarithm of the mirrored loss $\tfrac12(f(x+hu)+f(x-hu))-f(x)$ against $\log h$ at three scales (a quadratic gives $\beta=2$). The plateau probe reads ties at tiny steps, the pair test an additivity residual over coordinate pairs, the basin reading a rank-quadratic fit with second differences and the hill-valley test a point between two converged ends; the kink probe reads mirrored second differences at $10^{-7}$ of the box width, and the fractal probe their slopes up to $10^{-4}$ (Appendix~\ref{app:probes}). The curvature, pair and fractal statistics and the basin reading's second differences are invariant to $f\mapsto af+c$ ($a>0$); the plateau and hill-valley tests and the basin reading's rank fit are invariant to any strictly increasing transformation of $f$. The kink threshold, the stall monitor, the exit-status test and the absolute tolerances of Section~\ref{sec:limits} are not.

\subsection{Engines}\label{sec:engines}
The engines are SciPy's SLSQP \cite{kraft1988software}, L-BFGS-B \cite{byrd1995limited} and COBYQA \cite{virtanen2020scipy,ragonneau2022model,ragonneau2026cobyqa}; a scaled finite-difference BFGS after the nonsmooth BFGS of \citet{lewis2012nonsmooth}; STEP, a round-robin coordinate search in the style of BSrr \cite{swarzberg1994s,posik2015dimension}; MLSL from Sobol points \cite{rinnooykan1987stochastic}; and Nelder--Mead restarts at $D\le2$ \cite{nelder1965simplex}. The pool also holds a (1+1)-Cholesky-CMA-ES \cite{igel2006computational} with active update \cite{arnold2010active} and quadratic-model steps, IPOP-aCMA-ES and the ATM sampler \cite{bodner2019benchmarking}, a population-based random search with success-rate step control. Only the chain route uses a decoupled coordinate search, the one engine not run alone in Section~\ref{sec:res-ladder}. Appendix~\ref{app:engines} lists every setting.

\section{Development process and experimental design}\label{sec:design}

\subsection{The development loop}\label{sec:loop}
The development log of AMM lists 71 cycles. From composite c32 on, a cycle finds a losing cell (a function, dimension and budget where AMM trails the field of Section~\ref{sec:baselines}) on development instances and traces its mechanism. It then pre-registers a fix (a probe and an engine, the slice and numeric range of its gain, the accepted losses and the decision rule), screens it on development instances and confirms it once against the method without it on never-used bbob instances, which are then spent. A fix is kept, or \emph{promoted}, only if its prediction held; promoted fixes enter the next composite (c23--c46 in the code; c47 was the last candidate). AI coding-agent sessions (Claude Code) proposed the fixes, wrote the pre-registrations, and ran and analysed the batches; the author decided on and approved every pre-registration and promotion.

\paragraph{The record.}\label{sec:built} In the 16 confirmations of c32--c47, 32 of 36 written predictions held; the last candidate, c47, failed and was not promoted (full record and a worked pass of the loop in Appendix~\ref{app:build}). Single probe-and-engine pairs raised the target share on their slices by up to $+.773$ (the chain route on f20) and $+.711$ (the separable route on f3; Table~\ref{tab:confirm}). These add-one results show robustness to new instances of the same 24 functions, not a component's marginal value in the final configuration (Section~\ref{sec:res-parts}) or that AMM works on other functions (Section~\ref{sec:res-general}).

\subsection{Held-out evaluation}\label{sec:heldout}
\paragraph{Data split.} Ids are COCO's instance numbers (split in Appendix~\ref{app:tables}). The field comparisons on ids 61--70 steered about twenty composites, so counts computed on them are in-sample. The final test used 101--110 once (10 instances, where COCO's archive uses 15, to keep 111--115 in reserve). MA-BBOB rows 900--999 were never run before the test; an earlier composite (c44) ran once on rows 3--302 for information, and that run set the MA-BBOB test's predicted ranges, not a design decision.

\paragraph{Measure and the tested regime.} For each function, dimension, instance and budget $B=kD$, a method runs once and scores its target share: the share of 51 equally weighted targets, $10^{2}$ to $10^{-8}$, that its best $f$-value reaches, which is the value at $B$ of COCO's runtime empirical cumulative distribution (ECDF) \cite{hansen2016cocob}; Appendix~\ref{app:perd} adds the final target and log-precision. The final test's $k\in\{10,30,100,300,1000\}$ and $D\in\{2,5,10,20,40\}$ are development grid points (development also used \xD{3000} and $D=3$, outside the pre-registered test grid), so the test measures robustness to new instances only. We call this setting, bbob f1--f24 at these $D$ and $k$ with $B$ known before the run, the \emph{tested regime} (limits in Section~\ref{sec:limits}). AMM uses $B$ to choose its route, so each $k$ is a separate COCO dataset; results are compared per $k$ over (function, dimension, budget) \emph{cells}, and pooled differences average the 24 functions and five dimensions (per-dimension intervals in Appendix~\ref{app:perd}).

\paragraph{Statistics.} Differences are paired by function, dimension, instance and seed. Final-test and ladder intervals are 95\% percentile bootstraps over the 120 (function, dimension) cell means (2,000 resamples; Appendix~\ref{app:perd} repeats them over the 24 functions); Holm's correction over the five budgets of a comparison applies to its tests, not its intervals, and per-(dimension, budget) intervals are uncorrected. AMM has the highest target share in a cell if its mean is within $10^{-12}$ of the best competitor's, as fixed before the results were read; outright and tied cells are reported apart. Two methods are \emph{level} when the 95\% interval of their difference contains zero. A cell is credibly behind if, under a two-level moment-shrinkage model over its 10 paired runs (\emph{eb2}; Appendix~\ref{app:build}), the posterior probability of a deficit larger than .02 exceeds .9 against any competitor, and credibly ahead if that holds for a lead against every competitor that ran it. The rule has no multiplicity control: on a 480-cell development A/A slice (one configuration run twice) a similar count produced 15 false behind cells, so these counts describe the data and are not significance statements.

\paragraph{Design budget and pre-registration.} Appendix~\ref{app:build} counts the development's batches, confirmation runs and composites; the variants screened and the total compute were not logged in full. The final test, the MA-BBOB test, both parts of the ladder, the ablation, the generalization runs, the route-firing predictions, the anytime mode's final-test runs, the \xD{10^4} and \xD{10^5} runs, the probes-at-default control, the NGOpt rerun and the runs off the development grid were pre-registered before their results existed.

\subsection{Baselines and the ladder of controls}\label{sec:baselines}
\paragraph{Same harness.} Same-harness methods run on AMM's instances, dimensions and budgets, once per budget: pycma 4.5.0 IPOP-aCMA-ES and BIPOP-aCMA-ES \cite{hansen2019cma}, restarted SciPy SLSQP, SciPy differential evolution (DE) \cite{virtanen2020scipy}, optuna 5.0 TPE (tree-structured Parzen estimator) \cite{akiba2019optuna} on a capped slice and nevergrad 1.0.12 NGOpt \cite{rapin2018nevergrad}. With HCMA's \emph{archive levels}, the per-budget target shares read from its archived anytime runs (its own instances, which differ by submission year, $D=2$--20), they form \emph{the field}, up to seven competitors per cell. Every \emph{arm}, a method with its settings, starts from a uniform point in the box, runs at its library defaults except SLSQP's and DE's stopping tolerances, and restarts until $B$ is spent or the final target is reached; nevergrad's MetaModel step needed an in-memory fix for NumPy 2.5 that touches nothing else. Appendix~\ref{app:arms} gives the seeds, settings, TPE slice and methods not run, among them lq-CMA-ES, TuRBO-1 and an ELA selector over AMM's engines. Our harness's IPOP-CMA-ES reads .020 to .057 below the archive's pycma IPOP entry per budget, so archive curves are not handicapped.

\paragraph{Ladder of controls.} To separate the value of the routing from that of the engine pool, AMM is compared on instances 101--110 with a ladder of controls, whose rungs are (1) each of its engines alone except the chain route's coordinate search; (2) fixed hand-overs between its stages: the earliest composite, and SLSQP then IPOP-CMA-ES with 20\% and 80\% of the budget \cite{hansen2019global}; (3) a selector that picks the best of the ten engines and a greedy schedule of up to three fixed-share stages of them, both fitted per dimension and budget on development ids 61--70, the schedule in the spirit of, not as a reproduction of, greedy restart schedules \cite{schapermeier2025greedy}; (4) nevergrad's NgIohTuned, an a-priori wizard; and (5) a post-hoc oracle, the per-cell best of all other methods (the field included, HCMA's archive levels excepted). The ladder was pre-registered in two parts, the first before the final test's results were read; Appendix~\ref{app:arms} gives both parts, the rungs' settings and their exceptions.

\paragraph{Parity of effort.} No competitor was tuned, while AMM's 67 non-default fields (Appendix~\ref{app:codemap}) were set in development. Two comparisons address part of this asymmetry. First, the selector and the greedy schedule draw on AMM's development data (ids 61--70, in-sample: Section~\ref{sec:heldout}) to choose and split among its engines, though they do not tune the engines. Second, the COCO archive's pycma IPOP entry, run outside our harness, trails AMM by $+.100$ to $+.274$ (Appendix~\ref{app:arms}), against $+.121$ to $+.312$ for the harness's arm. A competitor tuned on the development ids was not run, so the part of AMM's lead that tuning would close is not measured.

\paragraph{MA-BBOB\@.} The same configuration also runs MA-BBOB \cite{vermetten2023mab} in the setting of its competitions \cite{vermetten2025gecco}, with the same generator, 2-D and 5-D, \xD{2000} and the normalized anytime area, on the held-out public rows 900--999 of the 2024 and 2025 problem sets (the competitions' own test instances are hidden), five seeds, against the organisers' archived baseline runs, the 2025 winner NADE \cite{stein2025neighborhood} and our harness's pycma IPOP-CMA-ES; Appendix~\ref{app:arms} describes the archived runs, a check of our NADE transcription and the target-share views reported beside the anytime area. The 10-D and 20-D problems and a bbob-tuned modular CMA-ES were not run.

\section{Results}\label{sec:results}

\subsection{Final test on the official instances}\label{sec:res-final}

\begin{table*}[t]
  \caption{AMM on held-out instances 101--110 (never used in design), per budget $kD$. Top: function--dimension cells of 120, against the same-harness field and HCMA's archive levels (NgIohTuned is a rung of the ladder below, not a competitor), with the tie and posterior rules of Section~\ref{sec:heldout}. Bottom: AMM minus each control, pooled over functions and dimensions, with 95\% bootstrap intervals over cells (intervals over functions in Appendix~\ref{app:perd}); the engines-alone row reuses the field's SLSQP and IPOP-CMA-ES runs; the oracle leaves out HCMA's levels.}
  \Description[Final test: cell counts and the ladder of controls per budget]{A table with five budget columns, from 10 to 1000 times the dimension, and two blocks of rows. The first block counts the function-dimension cells where AMM has the highest target share (and how many of these are ties), is credibly ahead of every competitor, is credibly behind at least one, or neither. The second block gives the difference in target share between AMM and each rung of the ladder of controls, with intervals: a fixed hand-over of its own stages without probes, SLSQP then IPOP-CMA-ES, a selector that picks the best of ten engines per dimension and budget from development data, a greedy schedule of the engines, the best of the ten engines alone, nevergrad's NgIohTuned, and the per-cell oracle of every other method, which AMM trails from 10 to 100 times the dimension, matches at 300 and leads at 1000 times the dimension.}
  \label{tab:final}
  \scriptsize
  \begin{tabular}{@{}p{0.40\textwidth} lllll@{}}
    \toprule
    & \multicolumn{5}{c}{budget $kD$, $k=$} \\
    \cmidrule(l){2-6}
    & 10 & 30 & 100 & 300 & 1000 \\
    \midrule
    \multicolumn{6}{@{}l}{\emph{Final test, instances 101--110: cells of 120}} \\
    highest target share & 57 & 61 & 79 & 90 & 95 \\
    \quad of which tied with the best competitor & 30 & 16 & 15 & 35 & 50 \\
    credibly ahead of every competitor that ran the cell & 2 & 12 & 33 & 32 & 31 \\
    credibly behind at least one competitor & 11 & 10 & 4 & 6 & 4 \\
    neither credibly ahead nor credibly behind & 107 & 98 & 83 & 82 & 85 \\
    \midrule
    \multicolumn{6}{@{}l}{\emph{Ladder of controls: $\Delta$ = AMM minus the rung [95\% CI]}} \\
    fixed hand-over of AMM's stages without probes (the earliest composite) & $+.049$ & $+.094$ & $+.166$ & $+.211$ & $+.143$ \\[-1pt]
    & {\scriptsize[$+.031$, $+.071$]} & {\scriptsize[$+.062$, $+.132$]} & {\scriptsize[$+.118$, $+.218$]} & {\scriptsize[$+.163$, $+.269$]} & {\scriptsize[$+.102$, $+.189$]} \\
    SLSQP then IPOP-CMA-ES (20\% / 80\% of the budget) & $+.120$ & $+.144$ & $+.218$ & $+.262$ & $+.200$ \\[-1pt]
    & {\scriptsize[$+.080$, $+.166$]} & {\scriptsize[$+.109$, $+.182$]} & {\scriptsize[$+.171$, $+.269$]} & {\scriptsize[$+.211$, $+.318$]} & {\scriptsize[$+.151$, $+.255$]} \\
    selector: best of ten engines per dimension and budget, from development ids & $+.016$ & $+.030$ & $+.067$ & $+.153$ & $+.187$ \\[-1pt]
    & {\scriptsize[$+.011$, $+.020$]} & {\scriptsize[$+.000$, $+.060$]} & {\scriptsize[$+.036$, $+.103$]} & {\scriptsize[$+.109$, $+.201$]} & {\scriptsize[$+.138$, $+.244$]} \\
    greedy schedule of the engines, built on development ids & $+.016$ & $+.028$ & $+.061$ & $+.112$ & $+.154$ \\[-1pt]
    & {\scriptsize[$+.012$, $+.020$]} & {\scriptsize[$-.002$, $+.058$]} & {\scriptsize[$+.029$, $+.098$]} & {\scriptsize[$+.075$, $+.151$]} & {\scriptsize[$+.111$, $+.203$]} \\
    best of its ten engines alone, per budget (Table~\ref{tab:engines}) & $+.015$ & $+.031$ & $+.073$ & $+.154$ & $+.222$ \\[-1pt]
    & {\scriptsize[$+.007$, $+.022$]} & {\scriptsize[$+.016$, $+.047$]} & {\scriptsize[$+.040$, $+.111$]} & {\scriptsize[$+.109$, $+.204$]} & {\scriptsize[$+.170$, $+.280$]} \\
    NgIohTuned (nevergrad 1.0.12), an a-priori wizard & $+.083$ & $+.132$ & $+.214$ & $+.359$ & $+.235$ \\[-1pt]
    & {\scriptsize[$+.048$, $+.126$]} & {\scriptsize[$+.095$, $+.172$]} & {\scriptsize[$+.170$, $+.263$]} & {\scriptsize[$+.299$, $+.419$]} & {\scriptsize[$+.186$, $+.289$]} \\
    per-cell oracle of every row above and the field (post hoc) & $-.016$ & $-.050$ & $-.045$ & $-.003$ & $+.038$ \\[-1pt]
    & {\scriptsize[$-.022$, $-.011$]} & {\scriptsize[$-.072$, $-.031$]} & {\scriptsize[$-.060$, $-.030$]} & {\scriptsize[$-.030$, $+.029$]} & {\scriptsize[$+.013$, $+.069$]} \\
    \bottomrule
  \end{tabular}
\end{table*}

\begin{figure*}[t]
  \centering
  \includegraphics[width=\textwidth,alt={Three line charts, for dimensions 2, 10 and 40, of the share of targets reached against the budget for AMM and the field on the official instances 101 to 110.}]{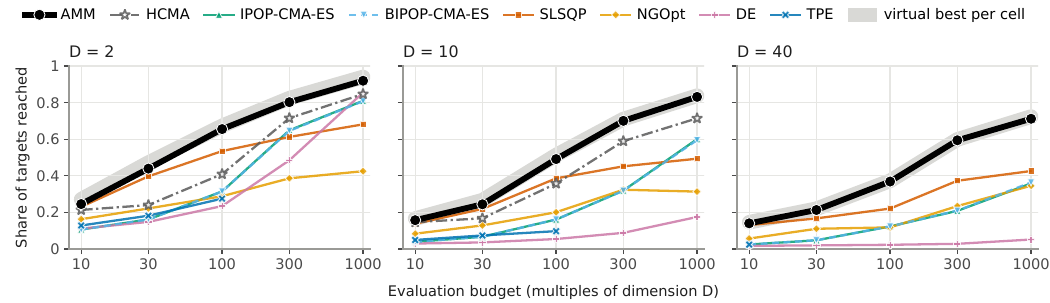}
  \caption{Final test per dimension (instances 101--110; same harness, one seed, each budget a separate run): share of the 51 targets reached (functions weighted equally) against the budget at $D=2$, 10 and 40 ($D=5$ and 20 in Figure~\ref{fig:ecdf-all}, Appendix~\ref{app:perd}); SLSQP restarted, TPE on its slice, HCMA from its archive levels ($D=2$--20). Grey band: the per-cell virtual best of the methods drawn, AMM included. NADE and ERADS, run after the final test, are not drawn (Figure~\ref{fig:teaser} and Table~\ref{tab:winners}).}
  \Description[Target share against budget, one chart per dimension (final test)]{Three line charts in a row, for dimensions 2, 10 and 40, under a legend that names the lines. The x-axis is the evaluation budget in multiples of the dimension, on a logarithmic scale from 10 to 1000; the y-axis is the share of targets reached, from 0 to 1. Each algorithm is one line with its own marker: AMM thick and black, HCMA grey and dash-dotted with hollow stars at dimensions up to 20 only, TPE blue with crosses where it ran. A light grey band marks the per-cell virtual best of the methods drawn, AMM included. The line of AMM is the highest or level at almost every budget; at dimensions 10 and 40 its gap to the other methods widens with the budget, and at dimension 2 it is largest at 100 times the dimension; NGOpt's line is flat or falls between 300 and 1000 times the dimension at dimension 10.}
  \label{fig:ecdf}
\end{figure*}
Every arm ran once on instances 101--110, and every run returned a finite value within its budget (6,000 runs per arm; 2,880 for TPE's slice). AMM has the highest target share in 382 of the 600 cells (by budget in Table~\ref{tab:final} and Figure~\ref{fig:teaser}, per dimension in Figure~\ref{fig:ecdf}), 236 outright and 146 tied (115 ties at a share of 1, all targets reached, and 25 at 0). No other method leads in more than 23 cells at any budget. It is credibly ahead of every competitor that ran the cell in 110 cells, credibly behind in 35 and neither in 455; without HCMA's archive levels it is highest in 415 (288 outright). Most of its mean margin over the per-cell best of all seven ($+.0435$) comes from f20, f3 and f4 (59.6\%), where the chain and separable routes act. Without them it is highest in 319 of 525 cells (60.8\%, against 63.7\%). Against pycma's IPOP-CMA-ES it leads by $+.121$ [$+.080$, $+.168$] at \xD{10}, $+.203$ [$+.153$, $+.254$] at \xD{30}, $+.303$ [$+.244$, $+.361$] at \xD{100}, $+.312$ [$+.251$, $+.375$] at \xD{300} and $+.222$ [$+.170$, $+.280$] at \xD{1000}, above zero under Holm's correction at every budget and in 25 of 25 dimension--budget cells. Against restarted SLSQP it leads by $+.015$, $+.037$, $+.118$, $+.225$ and $+.302$, with per-cell intervals that include zero in 4 of 25 ($D=2$ at \xD{10} and \xD{30}; $D=5$ and 10 at \xD{30}; Appendix~\ref{app:perd}). Against HCMA's archive levels it is ahead from \xD{30} upward (by $+.10$ to $+.17$) and level at \xD{10} ($+.016$), where HCMA has the highest mean in 13 cells. The pre-registered predictions held for the IPOP and SLSQP differences at every budget, and the predicted ranges for the credibly-behind and credibly-ahead counts (30--75 and 70--120) were met. After the final test, the 2025 MA-BBOB winner NADE \cite{stein2025neighborhood} and van Stein's 2024 algorithm ERADS \cite{stein2024llamea} were also run from their authors' code on the same grid and seeds; the decision to report this run was made after its results were seen. Both trail AMM at every budget and IPOP-CMA-ES from \xD{100} up (at \xD{1000}, .478 and .438 against .820 and .598), AMM is credibly behind NADE in one cell and ERADS in none, and with both in the field it would be highest in 380 cells instead of 382 (Appendix~\ref{app:winners}). Appendix~\ref{app:cocopp} post-processes COCO-observer recordings of the same runs against archived algorithms, with two budgets run after the final test under their own pre-registration, \xD{10000} and \xD{100000}; they lie beyond the developed budgets and enter no headline claim (Section~\ref{sec:res-anytime} uses them as the anytime mode's reference).

Figures~\ref{fig:ecdf} and~\ref{fig:ecdf-all} draw{} the final-test runs per dimension (intervals in Table~\ref{tab:perd}). At each of the five dimensions restarted SLSQP stays within $.010$ to $.019$ of AMM at \xD{10} and falls behind as the budget grows, by $+.238$ to $+.351$ at \xD{1000}. AMM's own share at \xD{1000} falls with the dimension, from .920 at $D=2$ to .831 at $D=10$ and .712 at $D=40$ (Table~\ref{tab:largeb}). NGOpt's curve is flat or falls from \xD{300} to \xD{1000} at $D=10$ ($.324$ to $.314$) and drops at $D=5$ ($.501$ to $.177$), one of the falls Section~\ref{sec:res-ladder} counts.

\subsection{The value of the routing}\label{sec:res-ladder}
Against every control we ran except the post-hoc oracle, AMM leads at every budget under the pre-registered one-sided tests (Holm over the five budgets; Table~\ref{tab:final}). From \xD{100} up the pooled margins show that the routing, not the pool alone, carries the gain, though not in every cell: the greedy schedule of the same engines is credibly ahead in 18 cells there (Section~\ref{sec:res-losses}). With every probe forced to its default on the same engines and schedule (pre-registered; fresh ids 880--899), AMM's share falls by $.022$, $.043$, $.160$, $.289$ and $.221$ at \xD{10} to \xD{1000}, each fall above zero under Holm, and its engines on that schedule are then no better than the best engine alone (Appendix~\ref{app:p0}). The best of its ten engines alone (Table~\ref{tab:engines}), the selector and the greedy schedule trail by only $+.015$ to $+.031$ at \xD{10} and \xD{30}. At \xD{30} the intervals of the selector and the schedule reach zero (Table~\ref{tab:final}), so at these budgets AMM is mostly the right engine run well. These rungs run the engines other than STEP, MLSL and the scaled BFGS at the harness's settings, not AMM's (Appendix~\ref{app:arms}). No arm with AMM's settings for them was run, so part of the small margins at \xD{10} and \xD{30} may come from those settings. Against the per-cell oracle of all these methods and the field (HCMA's levels not), AMM is lower at \xD{10}--\xD{100}, level at \xD{300} and higher only at \xD{1000} ($+.038$ [$+.013$, $+.069$], though resampled over the 24 functions [$-.001$, $+.095$]). It closes .49, .38, .60, .98 [.82, 1.21] and 1.26 [1.09, 1.51] of the gap between the ten-engine selector and that oracle at \xD{10} to \xD{1000}. Of the 23 pre-registered ranges of the ladder's first part, 15 held and 8 were exceeded in AMM's favour; of the 42 ranges of its second part, 37 held. Appendix~\ref{app:perd} lists the five others and the two category predictions that missed.

NGOpt, the a-priori wizard of the field, trails by $+.078$ at \xD{10} to $+.463$ at \xD{1000}, above zero under Holm's correction at every budget (intervals in Appendix~\ref{app:perd}). Nevergrad's NgIohTuned, a rung of the ladder, trails at every budget too (Table~\ref{tab:final}); it is about as good as NGOpt except at \xD{100} and \xD{1000}, where it is better by .076 and .228. As an eighth competitor NgIohTuned would leave AMM highest in 373 of 600 cells (225 outright), credibly behind in 39 and ahead of every competitor in 107. As pre-registered, it stays a rung, and the counts of Section~\ref{sec:res-final} exclude it. NGOpt's own share fails a budget check: it falls by more than .02 from one budget to the next larger one in 49 of its 120 function--dimension series (AMM's series fall in 8, Section~\ref{sec:limits}), and a pre-registered rerun traces the fall at $D=5$ to NGOpt itself: above \xD{500} at $5\le D<10$ it switches to a portfolio of nine short CMA-ES runs, and a rerun without the harness's NumPy patch (NumPy 2.3.5, 1,440 runs) agrees within $.014$ at all six settings checked (Appendix~\ref{app:arms}). Credited with its best smaller-budget level in each cell, NGOpt trails by $+.375$ at \xD{1000}, and AMM is highest in 381 cells instead of 382.

\subsection{Beyond bbob's function identities}\label{sec:res-general}
On MA-BBOB (setting in Section~\ref{sec:baselines}), AMM has a higher mean anytime area than every archived 2024 baseline (modcma by $+.047$ [$+.023$, $+.071$] at $D=2$ and $+.046$ [$+.014$, $+.078$] at $D=5$). Among the archived 2025 runs, NADE's and three baselines, it is level with the 2025 winner NADE at $D=2$ ($-.007$ [$-.019$, $+.009$]), second to it at $D=5$ ($-.049$ [$-.068$, $-.029$]) and level there with van Stein's 2024 algorithm ERADS \cite{stein2024llamea} ($+.001$ [$-.034$, $+.038$]); the 2025 archives hold one run per problem, so these comparisons are unpaired. Its final target share at \xD{2000} is level with IPOP-CMA-ES ($-.002$ [$-.012$, $+.008$] to $+.016$ [$-.001$, $+.035$]), while its anytime area trails IPOP-CMA-ES by $.026$ to $.064$. On the 2024 set's budget grid (point values over two seeds) it leads IPOP-CMA-ES in final target share by $+.031$ to $+.161$ at \xD{10}--\xD{100}, less than on bbob, trails it at \xD{300} for $D=5$ ($.610$ against $.733$; at $D=2$ it leads by $+.007$), and leads by $+.013$ and $+.044$ at \xD{1000}. On MA-BBOB its strength is thus the fixed-budget share at small and medium budgets. Every pre-registered prediction held, except one budget-grid cell that came out above its range in AMM's favour.

\paragraph{Transformed, rotated and box-constrained bbob.} In a batch pre-registered before any of its runs, AMM and same-harness competitors ran on fresh instances of three bbob variants. All 17 of its predictions (58 numeric ranges) held, with one fix to how runs were paired before its second analysis (Appendix~\ref{app:general}). COCO's bbob-boxed suite, formerly sbox-cost \cite{vermetten2023analysis}, adds bound constraints and moves the optimum within the box for 19 of the 24 functions. On it (ids 840--849, $D=2$--40; the field of Section~\ref{sec:baselines} without HCMA, NGOpt at $D=2$ only), AMM leads IPOP-CMA-ES by $+.117$ to $+.308$ and SLSQP by $+.011$ to $+.291$, above zero under Holm's correction at every budget and within .012 of its final-test lead over IPOP-CMA-ES. It has the highest share in 451 of 600 cells and is credibly ahead of every competitor in 155 and behind in 26. Under two affine maps of $f$, a box shrunk tenfold and the first map with that box (ids 810--819, $D=2$--20; definitions in Appendix~\ref{app:general}), AMM's lead over IPOP-CMA-ES stays above zero in all 16 tests of the four variants at \xD{30}--\xD{1000} (Holm over 20 with the identity). AMM is not invariant to the affine maps, though (Section~\ref{sec:limits}): its share falls by .032, .053 and .030 under the first, the second and the combined change, while IPOP-CMA-ES's does not change under the affine maps; the box change alone leaves AMM's unchanged ($+.001$). Rotating f1--f4 about their optima (ids 820--839, $D=2$--40; f5 is not rotated, Appendix~\ref{app:general}) leaves the sphere unchanged for every solver. It cuts AMM's lead over IPOP-CMA-ES on f3 from $+.505$ to $+.029$ and on f4 from $+.301$ to a level $+.011$, as predicted: there the lead comes from the coordinate-wise routes. On the rotated ellipsoid it keeps $+.390$, and over rotated f2--f4 it stays ahead of IPOP-CMA-ES at every budget ($+.015$ to $+.265$).

\paragraph{Dimensions and budgets off the development grid.} Every dimension and budget multiplier of the final test is a point of the grid that Table~\ref{tab:params}'s thresholds were set on. At $D=4$, 7, 15 and 30 and at \xD{20}, \xD{200}, \xD{350} and \xD{600} (pre-registered; fresh ids 870--879, all 24 functions), AMM leads IPOP-CMA-ES at each of 13 pooled settings under Holm (by $+.118$ to $+.295$ at the new dimensions and $+.160$ to $+.272$ at the new budgets of the old dimensions) and restarted SLSQP at each ($+.018$ to $+.293$), and on these fresh instances the final test's margins replicate within $.014$. The margin over IPOP-CMA-ES is about $.02$ below the grid's interpolation ($-.018$ [$-.031$, $-.005$]) and about $.06$ below at $D=4$, where above \xD{100} no gate and no landscape probe is active and the lead shrinks to $+.076$--$+.090$ at \xD{300}--\xD{600}, with intervals reaching zero (Appendix~\ref{app:unseen}).

\subsection{Parts of the method: ablation and route firing}\label{sec:res-parts}

\begin{table*}[t]
  \caption{Leave-one-route-out analysis of AMM on fresh bbob ids 770--809 (one seed per id; 206,280 runs): each arm is AMM with one route family switched off. $\Delta$ = target share without the family minus with it, on the family's slice (where its confirmation measured a gain; negative means the family helps), with its 95\% cluster-bootstrap interval over the 40 ids. Further columns: the range predicted before the run, and $\Delta$ on the cells outside the slice (``--'' where the slice is the whole block). Holm over the 13 rows rejects every row. Nothing was pruned.}
  \Description[Ablation: thirteen route families, each with the fall in target share when it is removed]{Thirteen rows, one per route family (the last, the two COBYQA hand-overs of c44 and c45), with the composite that introduced it, the slice of functions, dimensions and budgets on which it is measured, the change in target share with its interval when the family is removed, the range predicted before the run, and the change outside the slice. Every change on the slice is negative, from about minus seven tenths of a percentage point for the COBYQA hand-overs to about minus seventy-seven for the coordinate-search route on f20.}
  \label{tab:ablation}
  \scriptsize
  \begin{tabular}{@{}p{0.19\textwidth} l p{0.21\textwidth} l l l@{}}
  \toprule
    Route family (switch removed) & From & Slice & $\Delta$ on the slice [95\% CI] & Predicted & $\Delta$ off the slice \\
  \midrule
    COBYQA gate at $D\le3$, $B\le10D$ & c43 & all, D 2--3, 10 $\cdot D$ & $-.0158$ [$-.0196$, $-.0121$] & $-.035$ to $-.008$ & -- \\
    pair test $\to$ STEP restarts (separable) & c38 & f3, D 5--40, 100--1000 $\cdot D$ & $-.2239$ [$-.2352$, $-.2118$] & $-.70$ to $-.15$ & $+.0005$ [$-.0040$, $+.0051$] \\
    pair test $\to$ coordinate search (chain) & c38 & f20, D 5--40, 300--3000 $\cdot D$ & $-.7714$ [$-.7818$, $-.7600$] & $-.90$ to $-.50$ & $+.0013$ [$-.0020$, $+.0046$] \\
    basin reading $\to$ MLSL; basin descent & c39, c42 & f21, f22, D 2--3, 100--3000 $\cdot D$ & $-.1000$ [$-.1293$, $-.0716$] & $-.20$ to $-.05$ & $-.0061$ [$-.0143$, $+.0020$] \\
    prefix curvature routing; descent first & c25--c40 & 13 smooth functions, D 3--40, 10--100 $\cdot D$ & $-.1327$ [$-.1346$, $-.1308$] & $-.35$ to $-.08$ & $-.0035$ [$-.0043$, $-.0027$] \\
    ATM stage (its share goes to the next stage) & c23 & f15--f24, D 2--40, 300--1000 $\cdot D$ & $-.0202$ [$-.0291$, $-.0116$] & $-.07$ to 0 & $-.0212$ [$-.0280$, $-.0146$] \\
    STEP stall check & c39 & all, D 5--10, 1000--3000 $\cdot D$ & $-.0108$ [$-.0145$, $-.0073$] & $-.06$ to $-.003$ & $-.0008$ [$-.0038$, $+.0021$] \\
    smooth reading $\to$ scaled BFGS descent & c41 & f6, f12, f13, f14, D 5--40, 300 $\cdot D$ & $-.1676$ [$-.1773$, $-.1580$] & $-.30$ to $-.08$ & $-.0085$ [$-.0120$, $-.0051$] \\
    SLSQP stage gates: L-BFGS-B, cap gate, second descent & c32, c34 & f8, f9, f12, D 10--40, 100--1000 $\cdot D$ & $-.0985$ [$-.1111$, $-.0865$] & $-.30$ to $-.05$ & $-.0028$ [$-.0031$, $-.0025$] \\
    hill-valley $\to$ sampled SLSQP restarts & c35 & f21, f22, D 10--40, 1000--3000 $\cdot D$ & $-.2327$ [$-.2687$, $-.1965$] & $-.30$ to $-.08$ & $-.0120$ [$-.0143$, $-.0097$] \\
    kink route and rugged BFGS try & c37 & f13, D 5--40, 300--1000 $\cdot D$ & $-.1645$ [$-.1791$, $-.1496$] & $-.45$ to $-.10$ & $+.0003$ [$+.0000$, $+.0006$] \\
    plateau and rugged IPOP routes, fractal exemption & c28--c37 & f7, f15, f17, f18, f23, D 5--40, 100--3000 $\cdot D$ & $-.0368$ [$-.0413$, $-.0317$] & $-.25$ to $-.04$ & $+.0003$ [$+.0002$, $+.0004$] \\
    COBYQA hand-overs (Table~\ref{tab:rules}, rules (ii), (iii)) & c44, c45 & all, D 5--20, 10--30 $\cdot D$ & $-.0069$ [$-.0075$, $-.0063$] & $-.025$ to $-.003$ & -- \\
  \bottomrule
  \end{tabular}
\end{table*}

A leave-one-route-out ablation of AMM on fresh development ids (770--809, 206,280 runs) shows that every route family pays on its own slice. Removing any one of the twelve route families lowers its slice's target share by .0108 or more in point estimate, up to .7714 for the chain route on f20 (Table~\ref{tab:ablation}), and the thirteenth arm, without the two COBYQA hand-overs (rules (ii) and (iii) of Table~\ref{tab:rules}), lowers it by .0069. Outside its own slice no removal gained more than $+.0013$, and on its own slice the ATM stage costs less to remove than nine of the other eleven route families. The plateau and fractal family is mixed: its fractal exemption (the route that sends fractal readings to restarted SLSQP instead of IPOP populations) alone does nothing measurable on f23 at $D=5$--40 ($+.001$ [$-.010$, $+.012$]), where it acts only from $D=20$. The family's fall (Table~\ref{tab:ablation}) was narrowly outside its predicted range, the one slice-effect prediction missed (per-route values, A/A draws and code-identity checks in Appendix~\ref{app:ablation}). A leave-one-out measures a family's marginal contribution given the others, and six rows (the COBYQA hand-overs row included) bundle{} two to four mechanisms. 

Route-firing logs (last column of Table~\ref{tab:params}; heat-maps in Appendix~\ref{app:firing}) show that the separable and chain routes fire exactly on the functions that have their property: STEP on f1--f5 and the coordinate search on f20 only. Elsewhere most routes fire on the group of functions that share the property their probe reads (not the one-descent-first route of $D=2$--3), and six of the twelve function-motivated rows of Table~\ref{tab:params} fire on a function outside those that motivated them. Of 17 pre-registered predictions about firing, each with a range and a separate falsifying threshold, 12 held, four missed their ranges without being falsified and one was falsified. On the fresh ids the basin route reads f21 and f22 as multi-basin in only 41\% and 19\% of the runs at $D=2$--3 (at least 50\% predicted).

\subsection{The anytime mode}\label{sec:res-anytime}
AMM plans its phases from the budget it is given, so a run cut short is not the run that budget would have made. An optional anytime mode, added after the final test and not used by the configuration benchmarked above, reruns AMM on doubling plans (\xD{10}, \xD{20}, \dots) from the same start point and seed, answers every point evaluated before from memory, and hands over to an ordinary run once the next plan would exceed $10^4$ evaluations (Appendix~\ref{app:anytime}). On the final-test problems, which its development never used, all seven ranges of its own pre-registration held: it is level with AMM told its budget at \xD{10^4} and \xD{10^5}, and before the hand-over it pays for not knowing the budget, by $.05$ to $.12$ at \xD{30}--\xD{1000} (Table~\ref{tab:anytime}). Against the field it stays above every method at every budget except SLSQP at \xD{30} ($-.018$ $[-.027, -.008]$). Appendix~\ref{app:anytime} gives its mechanism, its cost in CPU time and when to use it.

\subsection{Where AMM does not win}\label{sec:res-losses}
In the final test AMM is credibly behind at least one competitor in 35 of 600 cells (by budget in Table~\ref{tab:final}), by dimension ($D=2$, 5, 10, 20, 40) in 11, 7, 8, 4 and 5. The per-cell map, with the full lists of the cells named below, is Figure~\ref{fig:lossmap} (Appendix~\ref{app:lossmap}), and the counts are descriptive (Section~\ref{sec:heldout}).

\paragraph{The smallest budgets, smooth functions.} Restarted SLSQP is credibly ahead in 11 cells, 10 of them at \xD{10}--\xD{30} on smooth functions at $D=2$--10. The largest gaps are f10 at $D=5$ and \xD{30} ($-.31$) and f12 at $D=2$ and \xD{30} ($-.26$). There a probe costs a large share of the budget, so AMM probes least and inherits the limits of COBYQA, SLSQP and the (1+1)-CMA-ES. Its COBYQA gate was accepted in development with named trades, losses accepted in advance, of $.05$ on f13 ($D=2$) and f11 ($D=3$) and $.04$ on f10 ($D=3$). A deficit against SLSQP seen in development on f9 ($D=10$, \xD{300}) is not among the 35 cells.

\paragraph{HCMA's archive levels.} They lead alone in nine cells, the largest on f7, the step ellipsoid, at $D=20$ and \xD{300} ($-.27$), and share two cells on f14 with IPOP- and BIPOP-CMA-ES (below).

\paragraph{NGOpt.} NGOpt leads in six cells: f9 at $D=5$ and 10 at \xD{10}, f17 at $D=5$ at \xD{100} and \xD{300} (by up to $-.090$), and f20 and f23 at $D=40$ at \xD{10}.

\paragraph{Restarted CMA-ES and TPE\@.} IPOP- and BIPOP-CMA-ES lead in seven cells on f12, f14, f17 and f18, the three largest on f18 at $D=2$ and \xD{1000} (BIPOP, $-.15$) and on f12 and f14 at $D=2$ and \xD{300} ($-.14$, $-.12$). TPE leads on f17 and f18 at $D=5$ and \xD{30} ($-.06$, $-.04$).

\paragraph{Against the controls.} By the same rule, AMM is credibly behind the greedy schedule of its engines in 25 cells, 7 at \xD{30} and 18 from \xD{100} up, the largest on f2 at $D=40$ and \xD{30} ($-.70$), and behind NgIohTuned in 7 cells, five at \xD{300} and two at \xD{10}, four of them not among the 35 (the largest on f16 at $D=2$ and \xD{300}, $-.33$).

\paragraph{Rotated, anytime and cheap problems.} Coordinate-wise routes cannot fire on rotated separable functions and fall back at the price of their probes. On rotated f3 and f4 (Section~\ref{sec:res-general}) the best of IPOP-, BIPOP-CMA-ES, SLSQP and DE leads AMM by more than .02 in 4 of the 25 cells of each (in none unrotated). An anytime measure such as MA-BBOB's area ranks AMM lower than its final value does: on the 2025 set at $D=5$ it has the highest anytime area among itself and the four archived runs on that set (NADE's and three baselines) on only 10\% of the problems (ERADS on 57\%). COBYQA alone takes about 4.4~ms per evaluation at $D=2$--3 (outside COCO's procedure), which matters when $f$ is cheap.

\section{Discussion and limitations}\label{sec:limits}
\paragraph{What probes do that other routers cannot.} A probe separates problems that metadata cannot: f3 and f15 are the same Rastrigin function without and with a rotation. A router that reads only $D$ and $B$ must send both to the same engine, while the pair test's $3D$ evaluations send f3 to STEP restarts (the separable route's gain in Section~\ref{sec:built}) and leave f15 on the general path. The probes' thresholds were not designed leave-one-function-out (an in-sample comparison with fitted routers is in Appendix~\ref{app:router}); the pooled value of reading $f$ (Section~\ref{sec:res-ladder}) is measured against one definition of the probes' defaults, and at \xD{10} that definition also removes a COBYQA stage that no reading chooses.

\paragraph{The budget must be known.} AMM told its budget chooses its route from $B$, so it is not an anytime algorithm (Section~\ref{sec:res-anytime} evaluates an optional anytime mode), and its share need not grow with $B$. In 8 (function, dimension) pairs its mean share drops by more than .02 from one budget to the next larger one, the two largest on f14 at $D=40$ (1.00 at \xD{300}, .865 at \xD{1000}) and f12 at $D=10$ (.986 to .880).

\paragraph{Absolute constants and coordinates.} The kink threshold $\tau_k=5\cdot10^{7}$ is tied to the scale of bbob's $f$-values. The restart tolerance $10^{-11}$, SLSQP's $\mathrm{ftol}=10^{-14}$ and the finite-difference step $10^{-9}$ are absolute, and the probe scales are fixed shares of the box width. The pair test, the chain route and STEP read structure in the given coordinates. The chain route detects a coupling that, among the bbob functions used in calibration, only f20 showed, and may be read as an f20 device. Section~\ref{sec:res-general} measures what both cost: under the affine maps of $f$ AMM's share falls by .03 to .05, so its target shares hold only at bbob's scale of $f$, and rotating f3 and f4 removes most of the lead their coordinate-wise routes give.

\paragraph{Probes outside their calibrated range.} In development, a separability route calibrated at \xD{300}--\xD{1000} misrouted rugged functions at \xD{3000} ($-.18$ to $-.25$ per cell). A fractal exemption lost $.16$ on f16 because Weierstrass's function read as rough as Katsuura's on other instances. Validity ranges limit such failures but cannot exclude a misreading on a problem class unlike bbob's.

\paragraph{Scope.} The final test, and with it the tested regime of Section~\ref{sec:heldout}, covers noiseless, box-bounded problems with $D=2$--40 and budgets of \xD{10}--\xD{1000}. No function family outside bbob's was tested under pre-registration; an exploratory check on 20 other functions at $D=2$--4 (Appendix~\ref{app:fresh}) found AMM level with restarted SLSQP at \xD{30}, ahead at \xD{100}, level with IPOP-CMA-ES at \xD{300}--\xD{1000} and behind NADE at \xD{1000}. The $D$-keyed thresholds (Appendix~\ref{app:routing}) come from the development grid; off it the lead held (Section~\ref{sec:res-general}), but at $D=4$, where above \xD{100} no gate and no landscape probe is active (only STEP's separability check and the finisher's stall rule) and a run takes the general path of the ATM stage, STEP and the finisher, the lead over IPOP-CMA-ES shrinks to about $+.08$ at \xD{300}--\xD{600}. Dimensions above 40 were not tested. The box sets probe scales, trust radii and restart points (it is mapped onto $[-5,5]^D$ inside AMM), so it should bound the region where the optimum plausibly lies; no box wider than bbob's or unbounded problem was tested. Finite differences assume exact $f$-values, and the quasi-Newton engines cost $O(D^2)$ or more per step. COCO's timing experiment was not run, for AMM or for its anytime mode.

\section{Conclusion and future work}\label{sec:conclusion}
On bbob, a router that tests each engine's assumption before running it spent a known budget better, pooled over cells, than any engine it routes among. Run once on held-out instances, one configuration of AMM has the highest target share among up to seven competitors in 382 of 600 cells (236 outright), leads IPOP-CMA-ES by $+.12$ to $+.31$ and beats every ladder control except the post-hoc oracle at every budget.

Within the tested regime (Section~\ref{sec:limits}), the routing, not the pool alone, carries the pooled gain from \xD{100} up; at \xD{10} and \xD{30}, where most of the 35 credibly-behind cells lie, AMM is mostly the right engine run well and restarted SLSQP comes close. In pre-registered runs the lead over IPOP-CMA-ES holds under affine and box transformations of bbob, on its box-constrained suite and, pooled, on rotated f2--f4, but the coordinate-wise routes pay only in the given coordinates (rotating f3 and f4 removes most of their lead); on MA-BBOB AMM's final share at \xD{2000} is only level with IPOP-CMA-ES's and its anytime area lower; and without a known budget the anytime mode loses $.05$ to $.12$ at \xD{30}--\xD{1000} (Section~\ref{sec:res-anytime}).

\paragraph{Future work.} MA-BBOB's generator could serve development and a held-out final test. Fitted and scale-free thresholds, rotation-invariant separability probes, shared batch evaluations and a compiled core would address Section~\ref{sec:limits}'s limitations and the overhead on cheap $f$; and resumable engines would narrow the anytime mode's gap.

\section*{Software availability}

AMM is an open-source Python package (Apache-2.0) that depends only on NumPy and SciPy, installable with \texttt{pip install amm-optimizer} (import \texttt{amm\_optimizer}). Its entry point \texttt{minimize} takes the inputs listed in Section~\ref{sec:method} (signature in Appendix~\ref{app:codemap}), and runs are deterministic given the seed. The benchmark harnesses stop each run at its budget; \texttt{minimize} itself does not cap the budget and can overrun it when the caller does not stop it. The source, at tag \texttt{v0.3.0}, is at {\def\UrlBigBreaks{\do\:}\url{https://github.com/BjBodner/amm-optimizer}}. Its preset \texttt{c46} is the configuration benchmarked here (the anytime mode of Section~\ref{sec:res-anytime} is in release 0.5.0, not in this tag; releases 0.3.0 and 0.5.0 reproduce every final-test run of \texttt{c46} bit for bit), and the repository holds the benchmark runner and the development's screening records (\texttt{benchmarks/}). The pre-registrations, the result files and the scripts behind the tables of this paper are not yet in the repository; they will be archived with a DOI and cited in the next version.

\begin{acks}
The author used Claude models (Anthropic), through Claude Code, for code development, for running and analysing the experiments, for literature search, and for initial drafting and editing the text. However, the text, tables and figures have been heavily edited and mostly rewritten/recreated by the author. The author is responsible for the content of this paper. The author received no external funding for this work.
\end{acks}

\bibliographystyle{agsm-doi}
\bibliography{references}

\appendix

\section{Code map and input audit}\label{app:codemap}\label{app:alg}
Algorithms~\ref{alg:main} and~\ref{alg:prefix} in Section~\ref{sec:routing} give the benchmarked configuration's control loop and its prefix, and Table~\ref{tab:codemap} where each of their lines lives in the code.

\begin{table*}[tp]
  \caption{Where each line of Algorithms~\ref{alg:main} and~\ref{alg:prefix} lives in the code (package \texttt{amm\_optimizer}; line numbers at tag \texttt{v0.3.0} of the public repository, whose preset \texttt{c46} is the benchmarked configuration; the tag renames the package and fixes two crashes outside the benchmarked range, and its recorded runs of \texttt{c46} are unchanged).  The configuration is the preset \texttt{c46} in \texttt{presets.py}; its local-phase route is \texttt{QN\_ROUTES["qn44"]} in \texttt{qn\_phase.py}.}
  \Description[Code map: algorithm lines and source functions]{A three-column table: the algorithm line, the step it performs, and the module and function that implement it, with line numbers.}
  \label{tab:codemap}
  \footnotesize
  \begin{tabular}{@{}l p{0.30\textwidth} p{0.55\textwidth}@{}}
    \toprule
    Line & Step & Module and function \\
    \midrule
    -- & rule (viii): the box mapped onto $[-5,5]^D$ & \texttt{api.py:minimize} (L102--129) \\
    \ref{ln:cobyqa} & COBYQA gate & \texttt{api.py:minimize} (L220--232); \texttt{cobyqa\_phase.py:run\_cobyqa\_restarts} (L28) \\
    \ref{ln:pair} & pair test, separable and chain routes & \texttt{api.py:\_additive\_route} (L508); \texttt{chain\_phase.py:pair\_test} (L66), \texttt{band\_consistency} (L182), \texttt{run\_coord\_hess} (L566); \texttt{step\_phase.py:run\_step\_phase} (L40) \\
    \ref{ln:basin}--\ref{ln:basin2} & basin reading, MLSL, basin descent & \texttt{api.py:minimize} (L279--315); \texttt{basin\_phase.py:multi\_basin\_route} (L209), \texttt{rank\_quadratic\_r2} (L85), \texttt{local\_reading} (L103) \\
    \ref{ln:prefix}; Alg.~\ref{alg:prefix} & prefix, with rules (i)--(v) and (ix) & \texttt{api.py:\_prefix\_runs} (L685), \texttt{\_prefix\_kind} (L695), \texttt{\_small\_switch\_kw} (L629), \texttt{\_prefix\_ridge\_kw} (L648), \texttt{\_parked2\_kw} (L663); \texttt{es\_phase.py:run\_es\_phase} (L54), \texttt{\_guarded\_descents} (L615); \texttt{qn\_phase.py:plateau\_check} (L1458), \texttt{curvature\_beta} (L1332), \texttt{run\_slsqp\_restarts} (L426); \texttt{cobyqa\_phase.py:run\_cobyqa\_from} (L90), \texttt{run\_cobyqa\_restarts} (L28); \texttt{pop\_cma.py:run\_population\_restarts} (L34) \\
    \ref{ln:atm} & ATM stage & \texttt{api.py:minimize} (L383--431); \texttt{search.py:Search} (L37) \\
    \ref{ln:step} & STEP hand-over and separability check & \texttt{api.py:minimize} (L439--468); \texttt{step\_phase.py:run\_step\_phase} (L40) \\
    \ref{ln:local} & local probes and routes & \texttt{api.py:\_local\_phase} (L838), \texttt{\_qn\_local\_phase} (L896); \texttt{qn\_phase.py:plateau\_check} (L1458), \texttt{curvature\_beta} (L1332), \texttt{micro\_slopes} (L720), \texttt{run\_scaled\_bfgs} (L902); \texttt{pop\_cma.py:run\_population\_restarts} (L34) \\
    \ref{ln:early} & early ends of the smooth route & \texttt{qn\_phase.py:failure\_majority} (L1232), \texttt{run\_lbfgsb\_restarts} (L1238), \texttt{multi\_basin} (L1486), \texttt{run\_sampled\_restarts} (L1057), \texttt{kink\_check} (L687) \\
    \ref{ln:direct}--\ref{ln:try} & rules (vii) and (vi): direct populations, capped BFGS try & \texttt{api.py:\_qn\_local\_phase} (L1265--1316), \texttt{\_kink\_reads} (L1332); \texttt{qn\_phase.py:run\_scaled\_bfgs} (L902) \\
    \ref{ln:finish} & finisher & \texttt{api.py:\_finish\_gated\_then\_ipop} (L1339, direct start L1354); \texttt{cma\_phase.py:run\_cma} (L409); \texttt{pop\_cma.py:run\_population\_restarts} (L34) \\
    \bottomrule
  \end{tabular}
\end{table*}

\paragraph{Code map.} At tag \texttt{v0.3.0} of the public repository the control loop is \texttt{src/amm\_optimizer/api.py} (\texttt{minimize} and the helpers it calls), and its phases are the modules named after them in the same folder (\texttt{cobyqa\_phase.py}, \texttt{chain\_phase.py}, \texttt{basin\_phase.py}, \texttt{es\_phase.py}, \texttt{step\_phase.py}, \texttt{qn\_phase.py}, \texttt{pop\_cma.py}, \texttt{cma\_phase.py}; \texttt{search.py} for the ATM stage). The benchmarked configuration is the preset \texttt{c46} in \texttt{presets.py}; with the presets it builds on, it sets 67 of the 160 fields of the dataclass \texttt{Variant} in \texttt{config.py} (which holds every default) away from their defaults, and its local-phase route is \texttt{QN\_ROUTES["qn44"]} in \texttt{qn\_phase.py}. The tag renames the package and fixes two crashes outside the benchmarked range; its recorded runs of \texttt{c46} are unchanged.

\paragraph{Input audit.}\label{app:audit} Every source file of the benchmarked package was searched for reads of a function or instance identifier, an optimal value or a target, and the suite position. The search returned 4 candidate lines, none of which reaches a decision: two chain-route docstring lines describing its calibration toy, a dataclass field-name lookup in the configuration module, and a quasi-Newton-phase docstring mentioning instances with a large $|f_{\mathrm{opt}}|$. The entry point is \texttt{minimize(fun, x0, budget, *, seed, rng, config, profile, bounds)}.

\section{Routing, probes and engines in full}\label{app:routing}
This appendix gives every rule of the benchmarked configuration{} (thresholds and symbols in Table~\ref{tab:params}).

\paragraph{Whole-budget gates.} Section~\ref{sec:routing} walks through the three gates (Algorithm~\ref{alg:main}, lines~\ref{ln:cobyqa}--\ref{ln:basin2}). COBYQA is a trust-region method on quadratic interpolation models. The basin gate's Sobol design has $\min(16D,\max(D+1,\lfloor B/4\rfloor))$ points, and the gate reads the basins at the design's best point; after a basin reading that does not fire, the run continues from that point.

\paragraph{The prefix ($B\le100D$).} The prefix spends the whole budget (Algorithm~\ref{alg:prefix}): it evaluates a uniform design of $\max(4,D)$ points and, from \xD{30}, runs STEP on half of the remaining evaluations. The curvature probe then sends a smooth reading to restarted SLSQP (L-BFGS-B after a majority of failed exits) and a rugged one to the (1+1)-CMA-ES or, at \xD{100} on a rugged non-separable reading, to IPOP populations ranked by a quadratic surrogate. Below \xD{31} and 200 evaluations, where the probe would take a large share, the prefix skips it: one SLSQP descent from the design's best point, then the (1+1)-CMA-ES on the rest. At $D\le2$, Nelder--Mead restarts replace the evolution strategy, preceded up to \xD{30} by one SLSQP descent.

\paragraph{The ATM stage and the hand-over ($B>100D$).} The samples of the ATM stage (line~\ref{ln:atm}) later warm-start the covariance of the (1+1)-CMA-ES; a separable verdict of STEP's check (Appendix~\ref{app:probes}, line~\ref{ln:step}) goes straight to the finisher.

\paragraph{Local probes and routes ($D\ge5$, non-separable).} The plateau and curvature probes classify the incumbent as on a plateau, smooth or rugged (line~\ref{ln:local}).
On a plateau, IPOP-aCMA-ES runs from a uniform point with four times the default population, doubling per restart, to the end of the budget.
A rugged reading with at least $c_rD$ left (Table~\ref{tab:params}) gets the same IPOP schedule. From $D=20$ a fractal probe first diverts fractal readings to restarted SLSQP, and with at least $c_8D$ left the schedule starts at eight times the default population.
Other rugged readings get one capped BFGS try at $D\le9$ up to \xD{500}, with at least \xD{200} left, then the finisher with populations sized to the budget left (lines~\ref{ln:direct}--\ref{ln:try}); at $0.5\le\beta<1$ rule (vii) of Table~\ref{tab:rules} skips the try and, with at least \xD{100} left, starts those populations at once, and rule (vi) lets a try that reads a kink at its cap run to the end of the budget.
A smooth reading up to \xD{300} gets one descent of a scaled finite-difference BFGS. An SLSQP stage follows on a third of the remaining budget, capped at \xD{100}. If a stall monitor, read once at the cap, says the first descent is still progressing, the descent may continue past the cap; otherwise it continues, unless finished, for at most another \xD{100}. Three readings of the stage end the local phase early (line~\ref{ln:early}), each handing the rest to one engine. A majority of failed SLSQP exits sends it to L-BFGS-B, converged descents in separate basins to sampled SLSQP restarts, and, when the stage stalled, a kink with at least \xD{100} left to the scaled BFGS. If a single converged descent passed the monitor with at least \xD{400} left, one more descent follows from a uniform point; otherwise the finisher continues from the stage's best point.

\paragraph{The finisher.} A restarted (1+1)-Cholesky-CMA-ES (Appendix~\ref{app:engines}) runs from the incumbent (line~\ref{ln:finish}). It hands over when it stalls with at least \xD{100} left, when an in-run curvature probe at \xD{30} reads $\beta<1.3$ ($D\ge5$), or when a run reaches \xD{150} ($D\ge10$). The protected restarted aCMA-ES populations that take over do not grow: twice the default size $4+\lfloor3\ln D\rfloor$ or, after a rugged local reading, four, two or one times it by the budget left (start points and step sizes in \texttt{api.py} and \texttt{pop\_cma.py}). Protected means that the schedule stops after a run that fails to improve on the incumbent. From $D=10$ the (1+1) then resumes at the best point, and a final uniform restart uses any evaluations left.

\paragraph{Rules added by the last three composites.} Composites c44--c46 added the rules of Table~\ref{tab:rules}, which Table~\ref{tab:params} does not list as rows and Algorithms~\ref{alg:main} and~\ref{alg:prefix} mark by their numerals (their fields in the preset \texttt{c46}, Appendix~\ref{app:codemap}; their confirmations in Table~\ref{tab:confirm}). The COBYQA hand-overs (ii) and (iii) are the ablation's thirteenth arm (Table~\ref{tab:ablation}).
\begin{table*}[tp]
  \caption{Rules added by composites c44--c46 (Appendix~\ref{app:routing}), with where each acts, the reading that triggers it, the action and the evaluations it adds (``--'': none beyond the action). Rules (ii) and (iii) are the two COBYQA hand-overs.}
  \Description[Rules added by the last three composites]{A nine-row table, one row per rule (i) to (ix) added by composites c44 to c46: where it acts in dimension and budget, the reading that triggers it, the action, the extra evaluations and the composite, with its motivating function where one is named. Rules (ii) and (iii) hand the rest of the prefix or budget to restarted COBYQA.}
  \label{tab:rules}
  \scriptsize
  \begin{tabular}{@{}l p{0.20\textwidth} p{0.27\textwidth} p{0.30\textwidth} l l@{}}
    \toprule
    Rule & Acts at & Reading & Action & Evaluations & From \\
    \midrule
    (i) & $D\ge5$ and $B>30D$: the prefix, after its design and STEP & the plateau probe reads a plateau & the rest of the prefix to IPOP-aCMA-ES populations & 4 & c44; f7 \\
    (ii) & $B\le10D$ and $4\le D\le20$ (below, the COBYQA gate takes the budget): the prefix's first SLSQP descent, read after its first and second line searches & the first line search took at least six trials, or the first two gradients are not parallel ($|\cos|<0.99$), unless the second line search accepted its first trial & the rest of the budget to restarted COBYQA (trust radius 0.25 of the box width); COBYQA also replaces the prefix's (1+1)-CMA-ES after a descent that ended early or a rugged curvature reading & -- & c44 \\
    (iii) & $B\le30D$: the prefix's SLSQP descent on a smooth reading, read once after seven iterations & a ridge: every full step overshoots by a similar factor and one backtrack recovers it, as across f13's sharp ridge & the rest of the prefix to restarted COBYQA (trust radius 0.1 of the box width) & -- & c45 \\
    (iv) & every prefix budget & an undefined prefix curvature reading (fewer than two positive mirrored losses) & counts as smooth unless the plateau probe then reads a plateau & 4 & c45 \\
    (v) & the one-descent prefixes & a descent that did not move, unless its first gradient was exactly zero & the descent is retried from uniform points, up to five times & -- & c45 \\
    (vi) & a capped rugged BFGS try ($D\le9$) & the kink probe reads a kink at the try's best point at its cap & the try continues to the end of the budget & 6 & c45; f13 \\
    (vii) & a rugged local curvature reading & $0.5\le\beta<1$ & the capped BFGS try and the finisher's first (1+1)-CMA-ES are skipped: the finisher's aCMA-ES populations start at once from a uniform point (with at least \xD{100} left) & -- & c45 \\
    (viii) & every box & -- & mapped onto $[-5,5]^D$ inside AMM (no change on bbob) & -- & c45 \\
    (ix) & $D\ge10$, $B=100D$ & a finite rugged prefix curvature reading below 0.15 & the (1+1)-CMA-ES is kept instead of the surrogate-ranked IPOP populations & -- & c46; f23 \\
    \bottomrule
  \end{tabular}
\end{table*}

\subsection{Probes}\label{app:probes}
\emph{Curvature} $\beta$: mirrored pairs $x\pm hu$ at $h\in\{0.03,0.1,0.3\}$ of the box width along four random unit directions (two below \xD{30}); $\beta$ is the least-squares slope of the logarithm of the mean mirrored loss of Section~\ref{sec:tests} against $\log h$; $\beta\ge1.45$ reads smooth (24 or 12 evaluations).
\emph{Plateau}: four mirrored probes at $3\cdot10^{-4}$ of the box width; a plateau if at least half of them repeat $f(x)$ exactly (4 evaluations).
\emph{Pair test}: the additivity residual $|f_{ij}-f_i-f_j+f_0|/(|f_i-f_0|+|f_j-f_0|)$ over the $D-1$ adjacent and $D$ random distant coordinate pairs. A residual of at least $10^{-6}$ on a distant pair reads non-separable; on adjacent pairs only, it reads a chain ($3D$ evaluations: $f$ at the point, $D$ single moves and the $2D-1$ pairs). The band check compares banded factors of finite-difference Hessians at two points ($6D$ evaluations).
\emph{Basin reading}: the $R^2$ of a full quadratic fit to the ranks of $f$ on the design (at most $0.5$ reads several basins), and the median over axes of $|\log_{10}|c_i(10^{-5})/c_i(10^{-3})||$, where $c_i(h)=(f(x+he_i)+f(x-he_i)-2f(x))/h^2$ and $h$ is a share of the mean box width (at most $0.1$ reads smooth). Both readings also need fewer than half of the axes to be flat, with all four differences zero ($4D$ evaluations).
\emph{STEP check}: it fails if a stored point, re-evaluated after the incumbent moved along another axis, differs from its separable prediction by more than $10^{-7}$ relative. From \xD{1000}, once STEP stalls, a two-axis additivity check replaces it: a stalled STEP never moves the incumbent, so the one-axis check would always pass.
\emph{Exit status}: after at least two SLSQP descents, more failed than converged exits.
\emph{Stall monitor}: SLSQP's iterations rebuilt from its evaluation stream; progressing while, over the last ten iterations, the median ratio of actual to predicted decrease is positive and at most 60\% of the line searches backtrack (no extra evaluations).
\emph{Hill-valley}: two converged end points lie in separate basins if a point at $\tfrac14$, $\tfrac12$ or $\tfrac34$ of the segment between them is worse than both \citep[up to three pairs;][]{maree2018real}.
\emph{Kink}: the median over three fixed unit directions $u$ of $|f(x+hu)+f(x-hu)-2f(x)|/h^2$, at $h=10^{-7}$ of the mean box width, is at least $5\cdot10^{7}$ (6 evaluations).
\emph{Fractal}: raw mirrored second differences $|f(x+hu)+f(x-hu)-2f(x)|$, not divided by $h^2$, at $h=10^{-7},\dots,10^{-4}$ of the mean box width along four fixed directions. Their $\log_{10}$ ratio between neighbouring scales is 2 where $f$ is smooth; the reading is fractal if the medians of the two finest ratios are both below $1.5$, unless the probe line shows a kink or a plateau (32 evaluations).
Section~\ref{sec:tests} states their invariances.

\subsection{Engines}\label{app:engines}
Restarted engines restart from uniform points until their share is spent.
SLSQP runs with the box as bounds, a forward-difference step of $10^{-9}$ (SciPy's default left a truncation floor on ill-conditioned functions in calibration) and $\mathrm{ftol}=10^{-14}$. L-BFGS-B uses ten correction pairs and no tolerance stop. COBYQA starts with a trust radius of $0.25$ of the box width ($0.1$ after rule (iii) of Table~\ref{tab:rules}).
The scaled finite-difference BFGS (``scaled BFGS'' in the tables; one engine for the smooth and the kink routes) follows \citet{lewis2012nonsmooth}. It takes forward differences along the eigenvectors of its inverse Hessian with steps clipped to $[10^{-12},10^{-6}]$, uses an Armijo and weak-Wolfe line search and, where the quasi-Newton step leaves the box, takes the quadratic model's minimizer over the box, as in SLSQP.
STEP takes parabolic steps in the separable route. The decoupled coordinate search runs global line searches along the columns of $M^{-1}$, where $H=M^\top\Lambda M$ is the unit-triangular factorization of a finite-difference Hessian; each direction moves one decoupled coordinate.
MLSL \cite{rinnooykan1987stochastic} evaluates a scrambled Sobol design of $16D$ points. In each round it then starts up to three SLSQP descents (step $10^{-9}$, $\mathrm{ftol}=10^{-14}$), in increasing $f$, from design points with no better design point or descent end within MLSL's critical radius $r_N$ ($\sigma=2$, $N$ the design points so far), and adds $16D$ Sobol points. The sampled SLSQP restarts of the smooth route add $10D$ uniform points before each descent and start it from the best sample point farther than $0.1\sqrt{D}$ mean box widths from the incumbent, every earlier start and every earlier descent's end (from a uniform point if none is). Their descents use SciPy's difference step and $\mathrm{ftol}=10^{-14}$; a descent still above the incumbent stops once its last $3(D+1)$ evaluations gained less than 1\% of its gap to the incumbent.
The (1+1)-CMA-ES's active update is triggered by a candidate worse than the fifth-last accepted point, and model steps in its own coordinates $z=A^{-1}(x-m)/\sigma$. Up to $D=10$, every other step goes to the minimizer of a full quadratic model fitted by least squares to the run's last $(D+1)(D+2)$ points (once $(D+1)(D+2)/2+D$ exist since the last restart), capped at $\|z\|=3\sqrt{D}$ and skipped unless the model's Hessian is positive definite. Above $D=10$, a diagonal quadratic model ($2D+1$ terms, fitted with rank weights to recent points) is tried at a period that adapts to its ratio of actual to predicted decrease (\texttt{cma\_phase.py}), with steps capped at $\|z\|=\sqrt{D}+2$. An improving model step is accepted but does not adapt $\sigma$ or the covariance. IPOP-aCMA-ES is a standard $(\mu/\mu_w,\lambda)$-CMA-ES with active update and box clipping. Its surrogate variant (the prefix's populations at \xD{100}) keeps its run's last evaluated points and fits the largest linear, diagonal or full quadratic model ($p$ terms; full only up to $D=20$) whose $1.1p$ is at most three quarters of the points kept, by weighted least squares on values over the best 75\% of the latest $2p$ of them. The model sorts each generation, and true evaluations proceed down that order in growing batches until the model's Kendall $\tau$ with the true values of the recent points reaches 0.85. Evaluated candidates rank by their true values and the others by the model; after a generation evaluated in full, the model pauses (\texttt{surrogate.py}). Both CMA-ES engines are reimplementations in NumPy.
In the ATM sampler, four sub-samplers draw perturbations around the incumbent and steps along the last improving direction, and a resource allocator moves samples between them. Their widths follow success-rate control with target rates from $0.05$ to $0.3$; a rank-$\mu$ covariance is learned up to $D=10$, the initial design is uniform, and samples are folded into the box.

\section{How the method was built: rules and full record}\label{app:build}
All results in this appendix are validation or development data, not the held-out evaluation.

\paragraph{Decision rules.} Differences between configurations are paired by function, dimension, instance and seed, with 95\% intervals from a cluster bootstrap over (function, dimension, instance). Every confirmation stated its predictions, slices and numeric ranges before the first run, named the cells where a loss was accepted (named trades), and applied Holm's correction across its predictions. If some predictions failed but the guards held, only the components whose predictions held were promoted. Screens on development instances used a fixed acceptance rule: no Holm-significant harm in any function group and budget, no dimension-budget confidence interval below zero, no flag from a two-level shrinkage gate (\emph{eb2}) outside the named trades, and the predicted sign on the primary slice.

\paragraph{The eb2 rule.} For a pair of arms, the paired differences of each (instance, seed) are averaged over the budgets; the mean and standard error of these means give each (function, dimension) group's mean. Group means are shrunk toward their grand mean with a between-group variance estimated by moments. Each (function, dimension, budget) cell's mean paired difference over its instances is then shrunk toward its group's shrunk mean with a between-cell variance estimated by moments. Its normal posterior carries the variance of the shrunk group mean. A cell is flagged when the posterior probability of a difference below $-.02$ exceeds .9; the same rule on negated differences finds credible gains. The 25\% narrowness below concerns per-cell bootstrap intervals, not eb2.

\paragraph{Calibration with A/A runs.} A/A runs, the same configuration twice with shifted seeds, calibrated the rules and changed the protocol. The guardrail ``no group loses more than .02'' failed about 59\% of the time with no true effect. A count of ``clearly behind'' cells produced 15 false clearly-behind cells on a 480-cell A/A slice, against a limit of 4 that had rejected several arms. Per-cell bootstrap intervals were about 25\% too narrow (mean $z^2=1.60$ over 14 A/A cells), so single-cell intervals are widened by a factor of 1.25. The \emph{eb2} gate raised no false flag in its A/A replay and detected 79\% of injected harms of $-.10$. That calibration was itself in-sample, and the gate was later found uncalibrated on slices without smooth functions, where it is reported but does not decide.

\begin{table*}[t]
  \caption{Pre-registered confirmations of the components added from c32 to c46, and of the rejected c47 (validation data). Each composite was compared with its predecessor on bbob instance ids never used before, one seed per instance (two for c40, c42 and c43); estimates are differences in target share with 95\% cluster-bootstrap intervals. Rows: the 22 component predictions of c32--c43, the 8 of c44--c46 and c47's prediction; the five other predictions of the c32--c43 confirmations (each composite against its predecessor over all cells, or against IPOP-CMA-ES) also held. These results show robustness to new instances of the same functions only.}
  \Description[Confirmation table: 31 predictions, 27 held and 4 did not]{A six-column table with one row per pre-registered component prediction from composite c32 to c47: the composite, the component (test and engine), the target slice of functions, dimensions and budgets, the fresh instance ids, the estimate with its interval, and whether the prediction held. Four rows did not hold: two tests of the same kink route for f13, a plateau switch for f7 and c47's extended SLSQP-stage restart; their intervals include zero.}
  \label{tab:confirm}
  \footnotesize
  \begin{tabular}{@{}l p{0.30\textwidth} p{0.25\textwidth} l l l@{}}
    \toprule
    Composite & Component (test $\to$ engine) & Target slice & Fresh ids & Estimate [95\% CI] & Held \\
    \midrule
    c32 & SLSQP exit-status majority $\to$ L-BFGS-B & f12, $D=10$--40, \xD{100}--\xD{1000} & 176--185 & $+.356$ [$+.281$, $+.427$]$^a$ & yes \\
    c33 & rugged reading, $D\ge20$ $\to$ IPOP from a uniform point & f17, f18, $D=20$--40, \xD{1000}--\xD{3000} & 186--195 & $+.171$ [$+.100$, $+.241$] & yes \\
    c34 & stall monitor at the stage cap; second descent & f8, f9, $D=20$--40, \xD{300}--\xD{1000} & 196--210 & $+.172$ [$+.122$, $+.224$] & yes \\
    c35 & hill-valley basins $\to$ sampled SLSQP restarts & f21, f22, $D=10$--40, \xD{1000}--\xD{3000} & 211--245 & $+.191$ [$+.155$, $+.228$] & yes \\
    c35 & kink route (smaller SLSQP step) & f13, $D=20$--40, \xD{1000}--\xD{3000} & 211--225 & $+.028$ [$-.011$, $+.069$] & no \\
    c36 & prefix: separable smooth reading $\to$ SLSQP; no ATM tail up to \xD{100} & all, $D=2$--40, \xD{30}--\xD{100} & 246--285 & $+.016$ [$+.015$, $+.018$] & yes \\
    c36 & fractal reading $\to$ restarted SLSQP & f23, $D=20$--40, \xD{1000}--\xD{3000} & 246--285 & $+.044$ [$+.036$, $+.053$] & yes \\
    c36 & same kink route, retested & f13, $D=20$--40, \xD{1000}--\xD{3000} & 246--285 & $+.017$ [$-.006$, $+.039$] & no \\
    c37 & kink $\to$ scaled BFGS & f13, $D=10$--40, \xD{1000} & 286--325 & $+.304$ [$+.280$, $+.328$] & yes \\
    c37 & rugged reading, $D\le9$ $\to$ one BFGS try & f13, $D=5$, \xD{300} & 286--325 & $+.377$ [$+.289$, $+.461$] & yes \\
    c37 & rugged reading, $D=5$--19 $\to$ IPOP & f15, $D=5$--10, \xD{3000} & 286--325 & $+.106$ [$+.016$, $+.196$] & yes \\
    c37 & plateau check in the prefix & f7, $D=5$--40, \xD{100} & 286--325 & $+.008$ [$-.004$, $+.018$] & no \\
    c38 & pair test: chain $\to$ coordinate search & f20, $D=5$--40, \xD{300}--\xD{3000} & 326--365 & $+.773$ [$+.758$, $+.786$] & yes \\
    c38 & pair test: separable $\to$ STEP restarts & f3, $D=5$--40, \xD{100} & 326--365 & $+.711$ [$+.675$, $+.745$] & yes \\
    c39 & basin reading $\to$ MLSL & f21, f22, $D=2$--3, \xD{100}--\xD{3000} & 490--529 & $+.107$ [$+.076$, $+.138$] & yes \\
    c39 & STEP stall $\to$ two-axis check & all, $D=5$, \xD{3000} & 490--529 & $+.040$ [$+.029$, $+.052$] & yes \\
    c40 & whole budget to the prefix; descent first & all, $D=5$--10, \xD{10} & 530--569 & $+.024$ [$+.021$, $+.028$] & yes \\
    c40 & descent first below \xD{31} & all, $D=3$--5, \xD{30} & 530--569 & $+.050$ [$+.044$, $+.056$] & yes \\
    c41 & smooth reading $\to$ one scaled BFGS descent & f6, f12, f13, f14, $D=5$--40, \xD{300} & 570--609 & $+.163$ [$+.151$, $+.175$] & yes \\
    c42 & smooth single-basin reading $\to$ one SLSQP descent & f6, f8, f9, f12, f13, $D=2$--3, \xD{100}--\xD{300} & 400--449 & $+.103$ [$+.092$, $+.114$] & yes \\
    c43 & $D\le3$, $B\le10D$ $\to$ COBYQA & all, $D=2$, \xD{10} & 366--399 & $+.026$ [$+.019$, $+.033$] & yes \\
    c43 & same & all, $D=3$, \xD{10} & 366--399 & $+.017$ [$+.013$, $+.021$] & yes \\
    c44 & plateau check in the prefix $\to$ IPOP populations & f7, $D=5$--40, \xD{100} & 610--649 & $+.079$ [$+.043$, $+.116$] & yes \\
    c44 & small-budget COBYQA hand-over of the prefix & all, $D=5$--20, \xD{10} & 610--649 & $+.012$ [$+.010$, $+.014$] & yes \\
    c45 & rugged BFGS try continued on a kink reading & f13, $D=5$, \xD{300} & 650--689 & $+.075$ [$+.055$, $+.095$] & yes \\
    c45 & prefix ridge reading $\to$ COBYQA & f13, $D=20$--40, \xD{30} & 650--689 & $+.095$ [$+.083$, $+.108$] & yes \\
    c45 & undefined prefix reading $\to$ smooth unless a plateau & f12--f14, $D=20$--40, \xD{100} & 650--689 & $+.045$ [$+.030$, $+.061$] & yes \\
    c45 & descent retries in the one-descent prefix & f10--f12, $D=2$--5, \xD{30} & 650--689 & $+.047$ [$+.030$, $+.066$] & yes \\
    c45 & rugged reading, $0.5\le\beta<1$ $\to$ populations at once & all, $D=5$--10, \xD{300} & 650--689 & $+.006$ [$+.005$, $+.008$] & yes \\
    c46 & low prefix curvature, $D\ge10$ $\to$ (1+1)-CMA-ES & f23, $D=10$--40, \xD{100} & 690--729 & $+.046$ [$+.040$, $+.051$] & yes \\
    c47 & extended SLSQP-stage restart (not promoted) & f8, f9, $D=5$--40, \xD{300} & 730--769 & $+.007$ [$-.002$, $+.017$] & no \\
    \bottomrule
  \end{tabular}

  \smallskip
  {\footnotesize $^a$Rescored with exact f12 optima (Appendix~\ref{app:build}, ``A benchmarking pitfall''); first reported as $+.296$ [$+.209$, $+.386$].}
\end{table*}

\paragraph{Every component, on new instances.} The 13 holdout or confirmation runs before c32's (fresh ids within 16--100 and 116--175, for the composites up to c31) were pre-registered too, their predictions written in the development log before each run. This paper does not score those predictions; its counts cover c32--c47. Candidates reached a confirmation only after passing a pre-registered development screen: in cycles 63--71, 7 of the 15 screens with a recorded outcome rejected their candidate (the development batches' results notes record them). In the twelve confirmations of c32--c43 (Table~\ref{tab:confirm}), 24 of 27 pre-registered predictions held. The three that failed were not promoted: a kink route for f13, in its first confirmation and in a retest on more instances, and a plateau check in the prefix for f7. The three composites confirmed after c43 (c44--c46, fresh ids 610--729) held all 8 of their predictions; the last candidate, c47 (ids 730--769), failed its single prediction and was not promoted. Over these 16 confirmations, 32 of 36 predictions held. These are add-one results (Section~\ref{sec:built}).

\paragraph{A pass of the loop.} On f13, bbob's sharp-ridge function, SLSQP stalls at a kink. A first fix (a smaller SLSQP step, predicted to gain $+.04$ to $+.15$ in target share) gained $+.028$ with an interval that included zero and failed again on a retest (predicted $+.02$ to $+.08$). The redesign (a scaled finite-difference BFGS, predicted $+.10$ to $+.35$) gained $+.304$ and was kept (Table~\ref{tab:confirm}).

\paragraph{Design budget.} The development log lists 92 numbered batches (104 entries with sub-batches and repeats) and 29 holdout or confirmation runs on fresh ids; the code records 24 composites (c23--c46), the last of which is benchmarked here. The 230 arm files in \texttt{benchmarks/arms} are the only record of the arms screened, which, like the total compute, were not logged in full.

\paragraph{Against restarted CMA-ES, on new instances.} Three confirmations compared an intermediate composite with pycma's IPOP-CMA-ES, run in the same harness on never-used instances. c23 led by $+.116$ [$+.106$, $+.127$] in the primary score (mean over \xD{30}--\xD{300}, $D=2$--20; ids 46--60), and c31 by $+.199$ [$+.179$, $+.219$] over the cells of its confirmation (ids 166--175). c32 led by $+.208$ [$+.195$, $+.222$] over $D=5$--40 and \xD{30}--\xD{1000} (ids 176--185), each of its 16 dimension--budget intervals above zero, from $+.141$ to $+.300$ (rescored with exact f12 optima, like the pooled lead). These comparisons use one seed per instance; the final test is the first comparison of the benchmarked configuration with IPOP-CMA-ES on unused instances.

\paragraph{Routers fitted in development.}\label{app:router} In development data (in-sample, without intervals), a router given the function's identity closed 53--70\% of the gap to the per-run oracle under leave-one-instance-out validation and none under leave-one-function-out; the unfitted rule ``separable verdict, then SLSQP'' closed 34\%.

\paragraph{The 2019 entry of the ATM sampler, reproduced.} \citet{bodner2019benchmarking} benchmarked the ATM sampler on the BBOB-2009 noiseless testbed; its headline, 23 of 24 functions solved in 2-D, is at $2\cdot10^{5}\cdot D$ evaluations. AMM's reimplementation reproduces the archived data: at $D=2$ and \xD{1000} its mean target share is .459, against .460 and .466 for the two archived trials.

\paragraph{The first quasi-Newton comparison changed the design.} When restarted SciPy SLSQP was added to the development field (ids 61--70), c23 had been predicted to lead it by $+.10$ to $+.25$ in the primary score. Instead SLSQP led c23 by $.032$ [$.018$, $.047$], on the smooth functions f6 and f8--f14 and on Gallagher's f21 and f22, at budgets up to \xD{300}. Every quasi-Newton route of AMM was added after this result, and SLSQP is in the field and in the SLSQP-then-IPOP-CMA-ES rung of the ladder (Section~\ref{sec:baselines}).

\paragraph{A benchmarking pitfall.} Optimal values reconstructed from COCO data lines that carry ten significant digits were wrong on 291 f12 instances, by .01 to .36, because $f$ at the origin is of order $10^{8}$--$10^{9}$ there. This affected two confirmations, c32's and c33's: their runs that had stopped on a falsely reached final target were re-run, and all were rescored with exact values, after which only c32's estimates changed (Table~\ref{tab:confirm}, footnote a). Reference values in benchmarking code should be checked by a second route, for example by evaluating $f$ at COCO's $x_{\mathrm{opt}}$.

\section{Related work, continued}\label{app:related}
\paragraph{One classical solver per budget band.} NEWUOA beats BIPOP-CMA-ES on sphere- and Rosenbrock-like functions and, restarted, on the Gallagher functions, but not elsewhere at tight targets \cite{hansen2010black}. Under restarts, Nelder--Mead wins in low dimension and quadratic models in higher \cite{posik2012restarted}; Powell's BOBYQA, NEWUOA and LINCOA trail SciPy's SLSQP and BFGS \cite{brockhoff2025benchmarking}.

\paragraph{Restart and surrogate CMA-ES\@.} IPOP- and BIPOP-CMA-ES restart CMA-ES with growing or interleaved populations \cite{hansen2009benchmarking,auger2005restart}, and NBIPOP-aCMA-ES adapts the split \cite{loshchilov2012alternative}. lq-CMA-ES, saACM-ES and DTS-CMA-ES add surrogates \cite{hansen2019global,loshchilov2012self,bajer2019gaussian}. Each is one mechanism inside one evolution strategy; AMM runs (1+1)- and IPOP-CMA-ES \cite{hansen2016cma} as engines.

\paragraph{Automated and LLM design.} LLaMEA generates optimizers with a language model and evaluates them on bbob or MA-BBOB \cite{stein2024llamea}, and BLADE benchmarks such generators \cite{stein2025blade}. Per-function tuning of a modular CMA-ES, validated on unused instances, matches or improves the BBOB-2009 per-function virtual best on half of the functions \cite{nobel2026modular}.

\section{Settings of the same-harness arms}\label{app:arms}
All arms run in the same harness on COCO's bbob problems, with the start points and restarts of Section~\ref{sec:baselines} and the settings of Table~\ref{tab:arms}. All runs use instances 101--110 and seed 9000. Random streams depend on (seed, function, $D$, instance), not on the budget, so the five budgets of a problem are dependent, as are HCMA's levels, read from one anytime run. Each competitor receives the box $[-5,5]^D$ through its own bounds argument. AMM evaluates only points in the box; its runs used a wrapper that would clip a point outside it and charge a quadratic penalty (weight 1). Every final-test result file records Python 3.12.10, NumPy 2.5.4, SciPy 1.18.1, cocoex 2.8.2, cma 4.5.0, one BLAS thread and Windows 11; the processor model was not recorded.
\begin{table*}[tp]
  \caption{Settings of the same-harness arms and the ladder's engines alone and NgIohTuned (Section~\ref{sec:baselines}), and of NADE and ERADS, run after the final test (Appendix~\ref{app:winners}).}
  \Description[Settings of the same-harness arms]{A three-column table with one row per arm: IPOP-aCMA-ES, BIPOP-aCMA-ES, SLSQP, differential evolution, TPE, NGOpt, the engines alone of the ladder's first rung, NgIohTuned, NADE and ERADS. For each it gives the package and version, and the settings used.}
  \label{tab:arms}
  \scriptsize
  \begin{tabular}{@{}p{0.13\textwidth} p{0.17\textwidth} p{0.64\textwidth}@{}}
    \toprule
    Arm & Package & Settings \\
    \midrule
    IPOP-aCMA-ES & pycma 4.5.0, \texttt{fmin2} & initial step size 2; population $4+\lfloor3\ln D\rfloor$, doubled at every harness restart; active update (pycma's default); pycma's default stopping tolerances (the harness passes only the budget, a seed, the bounds and the population size) \\
    BIPOP-aCMA-ES & pycma 4.5.0 & pycma's own BIPOP restarts (up to 9), the first from the harness's start point and later ones from uniform points in $[-4,4]^D$ \\
    SLSQP & SciPy 1.18.1 & the box as bounds; $\mathrm{ftol}=10^{-14}$; SciPy's default finite-difference step (about $1.5\cdot10^{-8}$; AMM's SLSQP descents use $10^{-9}$ except its sampled restarts, Appendix~\ref{app:engines}) \\
    DE & SciPy \texttt{differential\_evolution} & its defaults (best1bin, population $15D$, Latin-hypercube start that includes the harness's start point, L-BFGS-B polish); both tolerances 0; at most $\lfloor B/(15D)\rfloor+1$ generations, about 67 at \xD{1000} \\
    TPE & optuna 5.0 \texttt{TPESampler} & its defaults; one float per coordinate in $[-5,5]$; the start point enqueued as the first trial; a capped slice: $D=2$--20 at \xD{10} and \xD{30}, $D=2$--10 at \xD{100}, $D=40$ at \xD{10} \\
    NGOpt & nevergrad 1.0.12 & an array parameter bounded to the box and initialised at the start point; the remaining budget; one worker; the MetaModel fix (Appendix~\ref{app:arms}, \emph{NGOpt}) \\
    Engines alone (rung 1) & as each engine & STEP, MLSL and the scaled BFGS with AMM's settings (Appendix~\ref{app:engines}); SLSQP with SciPy's difference step; L-BFGS-B with tolerance stops and SciPy's difference step; COBYQA with an initial trust radius of 0.1 of the box width (AMM's gate: 0.25); Nelder--Mead and the ATM sampler at their defaults; the (1+1)-CMA-ES without active update or model steps; IPOP-CMA-ES as pycma \\
    NgIohTuned (rung 4) & nevergrad 1.0.12 & through NGOpt's wrapper, with the same MetaModel fix \\
    NADE & authors' code, Zenodo 15074959 & published defaults (population 50; neighbourhood 5; F 0.5 and CR 0.9, adapted at rate 0.1; a restart after 1,000 evaluations without improvement); budget factor $B/D$ \\
    ERADS & authors' code, Zenodo 11573574 & published defaults (population 50; F rising from 0.55 to 0.85 over the budget; CR 0.95; memory 0.3; no restarts); budget factor $B/D$ \\
    \bottomrule
  \end{tabular}
\end{table*}
\emph{IPOP-aCMA-ES against the archive.} Against the COCO archive's pycma IPOP entry (its own instances, so unpaired), the harness's arm reads .021, .057, .050, .038 and .020 lower at \xD{10} to \xD{1000} (mean .037; lower by more than .05 in 119 of the 600 cells, higher by more than .05 in 15). AMM leads the archived entry by $+.100$, $+.145$, $+.253$, $+.274$ and $+.202$. The harness restarts the arm from a new uniform point with a doubled population, and the instances differ; which of these causes the gap was not tested.
\emph{BIPOP-aCMA-ES} returned the same result as IPOP-CMA-ES (final precision and evaluation count) in 5,486 of the 6,000 final-test runs, including every run up to \xD{30}.
\emph{NGOpt.} nevergrad 1.0.12's MetaModel calls \texttt{float()} on a one-element array, which NumPy 2.5 rejects, so every NGOpt run that reached its surrogate step crashed. The harness recompiles that one function in memory from its own source, taking the element explicitly, and changes nothing else. The final test's records do not log which sub-optimizer NGOpt chose. A pre-registered rerun (D=5 and 10, \xD{100}--\xD{1000}, instances 101--110, 1,440 runs) logged its choices in a separate environment with NumPy 2.3.5, where nevergrad's own code runs unmodified, and the patch off: at $5\le D<10$ above \xD{500} NGOpt replaces its metamodel-assisted CMA-ES by a bet-and-run portfolio of nine short CMA-ES runs, of which the best continues, and at $D=5$ its share falls from $.499$ to $.177$ there as in the final test ($.501$ to $.177$; the 240 runs of the \xD{1000} cell are bit-identical to the final test's). Its share is within $.014$ of the final test's at all six settings. nevergrad 1.0.12's CmaFmin2 calls pycma without a seed, so the final test's NGOpt cells on that path ($D=5$--20 at \xD{100}--\xD{300}, $D=20$--40 at \xD{1000}) are single clock-seeded draws; rerunning the same configuration moved the share at $D=5$, \xD{100} by $-.004$ [$-.026$, $+.013$]. Section~\ref{sec:res-ladder} counts the series with a fall (49 of 120); of their 50 falls, 40 end at \xD{1000}, while the series of IPOP-, BIPOP-CMA-ES, SLSQP and DE never fall, and NGOpt fell on development ids too ($D=5$: $.476$ to $.191$ from \xD{300} to \xD{1000}). At $D=5$ and \xD{1000} it reaches only .698, .341 and .355 of the targets on the sphere, the ellipsoid and the linear slope, which IPOP-CMA-ES solves; at $D=2$ and \xD{1000} its mean share is .425 against IPOP-CMA-ES's .810. Its falls from \xD{300} to \xD{1000} lie at $D=2$, 5 and 10 (5, 22 and 12 series) and once at $D=40$. The rerun covers $D=5$ and 10 only: the falls at $D=2$ (5 series) and $D=40$ (1) were not rerun, and the rerun changed the NumPy release and the MetaModel fix together, so it does not separate them; the logged optimizer choice attributes the $D=5$ fall to NGOpt's own switch above \xD{500} (in nevergrad 1.0.12's source: NGOpt14 portfolios at $5\le D<20$, nine copies at $D=5$ and \xD{1000}; NGOpt21 below $D=5$; CMA-ES from $D=20$).
\emph{Ladder of controls} (Section~\ref{sec:baselines}): rung (1) ran COBYQA alone at $D=20$--40 above \xD{100} on the test ids only, for cost, so there the selector of rung (3) chooses among nine engines, COBYQA alone not having been run on the development ids. The ladder's second part (STEP, MLSL and the scaled BFGS alone, COBYQA at $D=20$--40 above \xD{100}, the ten-engine selector, the greedy schedule and NgIohTuned) was pre-registered before its own runs, and the schedule's stages were fixed on development ids before any test run of it.
\emph{Methods not run.} Wiz, restarted COBYQA and BOBYQA, TuRBO-1, lq-CMA-ES, IPOP-lq-CMA-ES and an ELA selector over AMM's engines are not in the field (NgIohTuned and restarted COBYQA alone are rungs of the ladder). The SLSQP-then-lq-CMA-ES portfolio appears only as an unpaired archive curve (Appendix~\ref{app:cocopp}); the same-harness SLSQP-then-IPOP-CMA-ES rung stands in for it.
\emph{NADE and ERADS on bbob} (Appendix~\ref{app:winners}) ran unmodified from their authors' files (CC BY 4.0), with shims outside them: \texttt{np.Inf} aliased to \texttt{np.inf} for NumPy 2, a problem object with the dimension and the bounds $[-5,5]$, the budget through their own budget factor, a hard cap that scores the best $f$ within $B$ evaluations, and as seed the first random integer of the final-test seed's optimizer stream, as for the other library arms.
\emph{MA-BBOB\@.} The MA-BBOB test compares AMM with the organisers' baseline runs of the 2024 problem set (modcma, modde, DiagonalCMA, differential evolution and RCobyla, from the data of \citet{vermetten2023mab}, 50 runs per public problem), with the 2025 winner NADE \cite{stein2025neighborhood}, and with the three baselines that the 2025 organisers shipped with their training set \cite{vermetten2025gecco}, namely van Stein's 2024 algorithm ERADS \cite{stein2024llamea}, Almonacid's PSS, second in 2024, and ACA, which we could not identify; the 2025 data hold one run per public problem, and neither competition published a leaderboard. A transcription of NADE in our harness reproduces its archived mean anytime area within $.003$. Because AMM spends its first evaluations on probes, Section~\ref{sec:res-general} also reports the target share at \xD{2000} and, on the 2024 set (two seeds), at \xD{10}--\xD{1000}.

\section{Final test per dimension}\label{app:perd}
\begin{figure*}[tp]
  \centering
  \includegraphics[width=\textwidth,alt={Two line charts, for dimensions 5 and 20, of the share of targets reached against the budget for AMM and the field on the official instances 101 to 110.}]{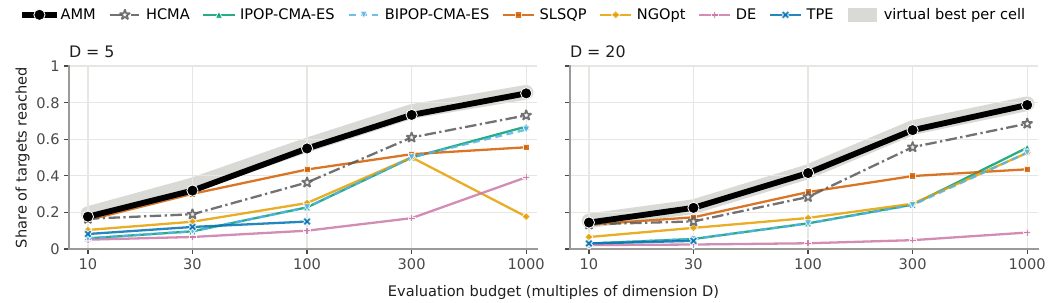}
  \caption{As Figure~\ref{fig:ecdf}, at $D=5$ and 20.}
  \Description[Target share against budget at dimensions 5 and 20 (final test)]{Two line charts, for dimensions 5 and 20, side by side under a legend that names the lines. The x-axis is the evaluation budget in multiples of the dimension, on a logarithmic scale from 10 to 1000; the y-axis is the share of targets reached. Each algorithm is one line with its own marker: AMM thick and black, HCMA grey and dash-dotted with hollow stars, TPE blue with crosses where it ran. A light grey band marks the per-cell virtual best of the methods drawn, AMM included. NGOpt's line falls between 300 and 1000 times the dimension at dimension 5.}
  \label{fig:ecdf-all}
\end{figure*}

Figure~\ref{fig:ecdf} (Section~\ref{sec:res-final}) and Figure~\ref{fig:ecdf-all} give the target-share curves at every dimension, and Tables~\ref{tab:perd}, \ref{tab:perd-fun} and~\ref{tab:engines} the final test per dimension, over functions and per engine. Over functions (Table~\ref{tab:perd-fun}; each function's five dimensions resampled together) every difference's interval stays above zero except those of the selector and the greedy schedule at \xD{30}, whose lower ends reach zero, and those of the per-cell oracle, which lie below zero at \xD{10}--\xD{100} and include zero at \xD{300} and \xD{1000}. The gap share's interval at \xD{1000} includes 1. Under two other measures, the order against the closest competitors holds except at the smallest budgets against SLSQP. On the final target alone (the share of runs that reach $10^{-8}$), AMM leads IPOP-CMA-ES at every budget ($+.080$ to $+.321$) and SLSQP from \xD{100} up ($+.080$ to $+.363$), and is level with SLSQP at \xD{10} and \xD{30} ($-.006$ [$-.016$, $+.002$] and $-.006$ [$-.018$, $+.004$]). Its mean $\log_{10}$ precision, clipped to $[-8,2]$, is lower than IPOP-CMA-ES's by 1.21 to 3.15 and than SLSQP's by 0.14 to 3.03 at every budget, every interval excluding zero. The normalized area under the ECDF up to $B$ could not be computed for the competitors, because the final test recorded only their final values (AMM's runs also went through COCO's observer, Appendix~\ref{app:cocopp}). Whether the order holds under this anytime measure on bbob is not known; on MA-BBOB, the measure ranks AMM below IPOP-CMA-ES (Section~\ref{sec:res-general}).

\paragraph{The ladder's pre-registered ranges.} The 8 ranges of the ladder's first part exceeded in AMM's favour (Section~\ref{sec:res-ladder}) were SLSQP then IPOP-CMA-ES at \xD{10}--\xD{300}, the earliest composite at \xD{300} ($+.211$ against at most $+.20$), and the gap share at \xD{100}--\xD{1000} (against .2 to .8). With the smaller oracle of that part (seven engines alone, the two hand-overs and the field), the gap shares from \xD{100} up were 1.34 to 1.40. In the second part, five of the 42 ranges missed: NgIohTuned at \xD{10} came out above its range, STEP alone at \xD{30} and MLSL alone at \xD{100} and \xD{300} below theirs, and the schedule's gain over the selector at \xD{300} above its range ($+.041$). Two category predictions also missed: that SLSQP would stay the best engine alone at \xD{30}--\xD{300} (Table~\ref{tab:engines}), and that the gap share would not fall below .8 at \xD{100}.

\begin{table*}[tp]
  \caption{Final test (instances 101--110): AMM minus IPOP-CMA-ES and restarted SLSQP per dimension and budget, and minus NGOpt pooled. Entries: mean target-share difference over the 24 functions with a 95\% bootstrap interval over functions (2,000 resamples; not corrected for multiplicity); ``pooled'': over the 120 (function, dimension) cells, interval over cells, as in Section~\ref{sec:res-final} (which gives IPOP-CMA-ES's) and Table~\ref{tab:final} (NGOpt's falls: Section~\ref{sec:res-ladder}).}
  \Description[Per-dimension differences against IPOP-CMA-ES, SLSQP and NGOpt]{Twelve rows: five dimensions for IPOP-CMA-ES, five dimensions and a pooled row for restarted SLSQP, and a pooled row for NGOpt; five budget columns from 10 to 1000 times the dimension; each cell gives AMM's mean lead in target share with its interval.}
  \label{tab:perd}
  \scriptsize
  \begin{tabular}{@{}ll ccccc@{}}
    \toprule
    & & \multicolumn{5}{c}{budget $kD$, $k=$} \\
    \cmidrule(l){3-7}
    Competitor & $D$ & 10 & 30 & 100 & 300 & 1000 \\
    \midrule
    IPOP-CMA-ES & 2 & $+.139$ [$+.068$, $+.232$] & $+.278$ [$+.176$, $+.391$] & $+.341$ [$+.225$, $+.464$] & $+.154$ [$+.054$, $+.259$] & $+.110$ [$+.038$, $+.197$] \\
     & 5 & $+.119$ [$+.038$, $+.227$] & $+.222$ [$+.117$, $+.338$] & $+.322$ [$+.189$, $+.460$] & $+.233$ [$+.126$, $+.350$] & $+.182$ [$+.077$, $+.313$] \\
     & 10 & $+.118$ [$+.035$, $+.234$] & $+.177$ [$+.076$, $+.296$] & $+.330$ [$+.192$, $+.471$] & $+.381$ [$+.250$, $+.513$] & $+.231$ [$+.112$, $+.367$] \\
     & 20 & $+.115$ [$+.028$, $+.231$] & $+.170$ [$+.069$, $+.292$] & $+.275$ [$+.150$, $+.409$] & $+.407$ [$+.253$, $+.564$] & $+.233$ [$+.124$, $+.359$] \\
     & 40 & $+.116$ [$+.022$, $+.238$] & $+.166$ [$+.063$, $+.288$] & $+.245$ [$+.125$, $+.387$] & $+.385$ [$+.238$, $+.534$] & $+.353$ [$+.215$, $+.499$] \\
    \midrule
    SLSQP (restarted) & 2 & $+.015$ [$-.017$, $+.047$] & $+.042$ [$-.007$, $+.095$] & $+.120$ [$+.078$, $+.171$] & $+.190$ [$+.099$, $+.293$] & $+.238$ [$+.131$, $+.357$] \\
     & 5 & $+.018$ [$+.007$, $+.029$] & $+.018$ [$-.023$, $+.054$] & $+.115$ [$+.048$, $+.205$] & $+.215$ [$+.116$, $+.328$] & $+.295$ [$+.165$, $+.438$] \\
     & 10 & $+.019$ [$+.011$, $+.029$] & $+.025$ [$-.007$, $+.058$] & $+.105$ [$+.035$, $+.205$] & $+.249$ [$+.145$, $+.366$] & $+.338$ [$+.200$, $+.495$] \\
     & 20 & $+.011$ [$+.0005$, $+.021$] & $+.052$ [$+.025$, $+.085$] & $+.104$ [$+.045$, $+.187$] & $+.251$ [$+.141$, $+.376$] & $+.351$ [$+.213$, $+.507$] \\
     & 40 & $+.010$ [$+.001$, $+.019$] & $+.046$ [$+.022$, $+.074$] & $+.148$ [$+.064$, $+.260$] & $+.221$ [$+.115$, $+.348$] & $+.286$ [$+.162$, $+.428$] \\
     & pooled & $+.015$ [$+.007$, $+.022$] & $+.037$ [$+.020$, $+.053$] & $+.118$ [$+.085$, $+.156$] & $+.225$ [$+.183$, $+.276$] & $+.302$ [$+.246$, $+.362$] \\
    \midrule
    NGOpt & pooled & $+.078$ [$+.044$, $+.121$] & $+.143$ [$+.101$, $+.188$] & $+.290$ [$+.231$, $+.349$] & $+.357$ [$+.297$, $+.422$] & $+.463$ [$+.400$, $+.524$] \\
    \bottomrule
  \end{tabular}
\end{table*}

\begin{table*}[tp]
  \caption{Pooled comparisons of the final test (instances 101--110) with 95\% bootstrap intervals over the 24 functions instead of the 120 (function, dimension) cells. Rows: AMM minus each method per budget $kD$ (means as in Section~\ref{sec:res-final} and Tables~\ref{tab:final} and~\ref{tab:perd}), and the share of the gap between the ten-engine selector and the per-cell oracle that AMM closes.}
  \Description[Pooled comparisons with intervals over functions]{Eleven rows: IPOP-CMA-ES, restarted SLSQP and NGOpt from the field; the earliest composite, SLSQP then IPOP-CMA-ES, the ten-engine selector, the greedy schedule, the best engine alone and NgIohTuned from the ladder; the per-cell oracle; and the gap share. Five budget columns from 10 to 1000 times the dimension; each cell gives the mean and an interval over functions. Every interval lies above zero except the selector and the greedy schedule at 30 times the dimension, which reach zero, and the oracle row, which is below zero from 10 to 100 times the dimension and contains zero at 300 and 1000 times the dimension.}
  \label{tab:perd-fun}
  \scriptsize
  \begin{tabular}{@{}l ccccc@{}}
    \toprule
    & \multicolumn{5}{c}{budget $kD$, $k=$} \\
    \cmidrule(l){2-6}
    AMM minus & 10 & 30 & 100 & 300 & 1000 \\
    \midrule
    IPOP-CMA-ES & $+.121$ [$+.040$, $+.232$] & $+.203$ [$+.104$, $+.318$] & $+.303$ [$+.189$, $+.422$] & $+.312$ [$+.202$, $+.420$] & $+.222$ [$+.131$, $+.328$] \\
    SLSQP (restarted) & $+.015$ [$+.005$, $+.025$] & $+.037$ [$+.014$, $+.058$] & $+.118$ [$+.066$, $+.194$] & $+.225$ [$+.136$, $+.330$] & $+.302$ [$+.188$, $+.432$] \\
    NGOpt & $+.078$ [$+.012$, $+.175$] & $+.143$ [$+.062$, $+.243$] & $+.290$ [$+.177$, $+.416$] & $+.357$ [$+.245$, $+.476$] & $+.463$ [$+.358$, $+.567$] \\
    \midrule
    earliest composite & $+.049$ [$+.021$, $+.088$] & $+.094$ [$+.036$, $+.170$] & $+.166$ [$+.090$, $+.260$] & $+.211$ [$+.124$, $+.303$] & $+.143$ [$+.079$, $+.215$] \\
    SLSQP then IPOP-CMA-ES & $+.120$ [$+.041$, $+.228$] & $+.144$ [$+.081$, $+.222$] & $+.218$ [$+.131$, $+.314$] & $+.262$ [$+.165$, $+.366$] & $+.200$ [$+.110$, $+.308$] \\
    ten-engine selector & $+.016$ [$+.011$, $+.020$] & $+.030$ [$-.000$, $+.068$] & $+.067$ [$+.015$, $+.144$] & $+.153$ [$+.062$, $+.261$] & $+.187$ [$+.092$, $+.303$] \\
    greedy schedule & $+.016$ [$+.012$, $+.020$] & $+.028$ [$-.000$, $+.062$] & $+.061$ [$+.010$, $+.135$] & $+.112$ [$+.053$, $+.187$] & $+.154$ [$+.083$, $+.239$] \\
    best engine alone & $+.015$ [$+.005$, $+.025$] & $+.031$ [$+.011$, $+.057$] & $+.073$ [$+.015$, $+.156$] & $+.154$ [$+.060$, $+.264$] & $+.222$ [$+.131$, $+.328$] \\
    NgIohTuned & $+.083$ [$+.016$, $+.182$] & $+.132$ [$+.062$, $+.222$] & $+.214$ [$+.137$, $+.303$] & $+.359$ [$+.249$, $+.475$] & $+.235$ [$+.144$, $+.341$] \\
    per-cell oracle & $-.016$ [$-.025$, $-.009$] & $-.050$ [$-.089$, $-.024$] & $-.045$ [$-.061$, $-.029$] & $-.003$ [$-.047$, $+.059$] & $+.038$ [$-.001$, $+.095$] \\
    \midrule
    gap share closed & .49 [.34, .67] & .38 [$-.00$, .65] & .60 [.24, .76] & .98 [.67, 1.48] & 1.26 [.99, 1.80] \\
    \bottomrule
  \end{tabular}
\end{table*}

\begin{table*}[tp]
  \caption{Mean target share over the 120 (function, dimension) cells of the final test (instances 101--110) for AMM and each of its ten engines alone (all but the chain route's coordinate search), restarted until the budget is spent (exceptions and settings in Appendix~\ref{app:arms}).}
  \Description[Mean target share of AMM and of each engine alone]{Eleven rows, AMM and its ten engines (SLSQP, L-BFGS-B, COBYQA, Nelder-Mead, the (1+1)-CMA-ES with restarts, IPOP-CMA-ES, the ATM sampler, STEP, MLSL, the scaled BFGS), and five budget columns from 10 to 1000 times the dimension. AMM has the highest value in every column; SLSQP, MLSL, the scaled BFGS and IPOP-CMA-ES are the best engines at different budgets.}
  \label{tab:engines}
  \scriptsize
  \begin{tabular}{@{}l rrrrr@{}}
    \toprule
    & \multicolumn{5}{c}{budget $kD$, $k=$} \\
    \cmidrule(l){2-6}
    & 10 & 30 & 100 & 300 & 1000 \\
    \midrule
    AMM & .173 & .288 & .496 & .696 & .820 \\
    \midrule
    SLSQP & .158 & .252 & .378 & .471 & .518 \\
    L-BFGS-B & .139 & .183 & .297 & .406 & .483 \\
    COBYQA & .153 & .187 & .271 & .376 & .488 \\
    Nelder--Mead & .036 & .078 & .175 & .239 & .312 \\
    (1+1)-CMA-ES, restarted & .087 & .141 & .223 & .340 & .488 \\
    IPOP-CMA-ES & .052 & .086 & .193 & .384 & .598 \\
    ATM sampler (defaults) & .024 & .035 & .065 & .122 & .227 \\
    STEP & .146 & .211 & .270 & .300 & .329 \\
    MLSL & .157 & .257 & .407 & .510 & .552 \\
    scaled BFGS & .143 & .242 & .423 & .541 & .595 \\
    \bottomrule
  \end{tabular}
\end{table*}

\subsection{NADE and ERADS on the final-test grid}\label{app:winners}
After the final test, two MA-BBOB algorithms by van Stein were run from their authors' published code on the final-test grid (f1--f24, instances 101--110, $D=2$--40, \xD{10}--\xD{1000}, one run per budget, paired with the stored final-test runs by instance and seed): NADE (code: Zenodo 15074959), the 2025 MA-BBOB winner \cite{stein2025neighborhood}, and ERADS (code: Zenodo 11573574), van Stein's 2024 algorithm; settings in Table~\ref{tab:arms}, shims in Appendix~\ref{app:arms}. A short pre-registration was written before the full run, but the decision to report the run here was made after its results were seen. Re-run through the same runner, 468 stored final-test runs of three field arms came out bit for bit identical. Both trail AMM at every budget. They are close to IPOP-CMA-ES at \xD{10} and \xD{30}, .004 to .014 lower in mean (the run's pre-registration expected them within .02 of it at \xD{10}), and below it from \xD{100} up (Table~\ref{tab:winners}). By the posterior rule of Section~\ref{sec:heldout}, AMM is credibly ahead of NADE in 408 of the 600 cells and behind in one (f7 at $D=5$ and \xD{300}: .778 against .941), and ahead of ERADS in 423 and behind in none. With both in the field, AMM has the highest target share in 380 of the 600 cells (234 outright, 146 tied) instead of the pre-registered 382 (236 and 146), is credibly behind some competitor in 36 instead of 35 and credibly ahead of every competitor in 109 instead of 110; NADE is highest in four cells (f4, f15 and f20 at $D=2$ and \xD{1000}; f7 at $D=5$ and \xD{300}), ERADS in none. The run's predicted ranges for both algorithms' means, the two credible counts and the highest count with both added (357--380) held; its predictions on their order, on ERADS's cells and on where NADE's cells come from missed, and the one credibly-behind cell lies on f7, not on the multimodal functions predicted. Both algorithms start from a uniform random population of 50 points, so at \xD{10} for $D\le5$ their runs are pure random sampling (at \xD{30} for $D=2$, random sampling plus at most one generation), and neither was designed for $D=10$--40 or for budgets other than \xD{2000} (ERADS was generated by LLaMEA on 5-D bbob \cite{stein2024llamea}, NADE tuned on MA-BBOB at $D=2$ and 5): the run shows the published entries off their design point, not what a differential evolution tuned for bbob could do.
\begin{table}[tp]
  \caption{NADE and ERADS on the final-test grid (instances 101--110; run after the final test, Appendix~\ref{app:winners}): mean target share over the 120 (function, dimension) cells per budget $kD$, beside AMM and IPOP-CMA-ES, and AMM minus each of the two.}
  \Description[NADE and ERADS on the final-test grid]{Six rows and five budget columns from 10 to 1000 times the dimension: the mean target share of AMM, IPOP-CMA-ES, NADE and ERADS, then AMM minus NADE and AMM minus ERADS. Both are close to IPOP-CMA-ES at the two smallest budgets and below it from 100 times the dimension, and below AMM at every budget; AMM's lead over them grows from about .13 at the smallest budget to .34 to .45 at the largest two.}
  \label{tab:winners}
  \footnotesize
  \begin{tabular}{@{}l rrrrr@{}}
    \toprule
    Budget $kD$, $k=$ & 10 & 30 & 100 & 300 & 1000 \\
    \midrule
    AMM & .173 & .288 & .496 & .696 & .820 \\
    IPOP-CMA-ES & .052 & .086 & .193 & .384 & .598 \\
    NADE & .048 & .072 & .138 & .280 & .478 \\
    ERADS & .048 & .077 & .132 & .243 & .438 \\
    \midrule
    AMM minus NADE & $+.125$ & $+.216$ & $+.358$ & $+.416$ & $+.342$ \\
    AMM minus ERADS & $+.125$ & $+.211$ & $+.364$ & $+.453$ & $+.382$ \\
    \bottomrule
  \end{tabular}
\end{table}

\section{Final-test loss map}\label{app:lossmap}
\paragraph{The cells behind.} Section~\ref{sec:res-losses} counts the cells where AMM is credibly behind; these are the cells. Restarted SLSQP leads in 10 cells at \xD{10}--\xD{30} on f2, f8 and f10--f14 at $D=2$--10, and on f13 at $D=40$ and \xD{100} ($-.025$). HCMA's archive levels lead alone on f7, the step ellipsoid, at $D=10$ and \xD{100} ($-.08$) and at $D=20$ and \xD{300} ($-.27$), on f19 at $D=10$ at \xD{10}--\xD{300} ($-.03$ to $-.05$), and in three cells on f13 and f24. IPOP- and BIPOP-CMA-ES lead together in six cells: f12 and f14 at $D=2$ and \xD{300} ($-.14$ and $-.12$), f14 at $D=20$ and 40 and \xD{1000} ($-.037$ and $-.041$), and f17 at $D=40$ and f18 at $D=20$, both at \xD{30} ($-.025$ and $-.033$); BIPOP-CMA-ES alone leads on f18 at $D=2$ and \xD{1000} ($-.15$). The f14 cells at $D=2$ and 20 are also HCMA's. The greedy schedule of AMM's engines is credibly ahead in 7 cells at \xD{30} (f2, f3 and f4 at $D=20$--40, f12 at $D=2$) and in 18 from \xD{100} up (on f4, f6, f10--f14, f19 and f23 at $D=10$--40). The largest gaps are f2 at \xD{30} at $D=40$ and 20 ($-.70$ and $-.63$), f12 at $D=2$ and \xD{30} ($-.46$), f4 at $D=20$ and \xD{300} ($-.44$) and f10 at $D=40$ and \xD{100} ($-.43$). Of the 7 cells where NgIohTuned is credibly ahead, four are not among the 35: f16 at $D=2$ ($-.33$), f23 at $D=40$ and 20 ($-.09$ and $-.07$) and f19 at $D=20$ ($-.05$), all at \xD{300}.
\begin{figure*}[tp]
  \centering
  \includegraphics[width=\textwidth,alt={Heatmap of AMM's target share minus the best competitor's, for 24 functions by 5 dimensions and 5 budgets, on the official instances 101 to 110, with the leader named where AMM trails.}]{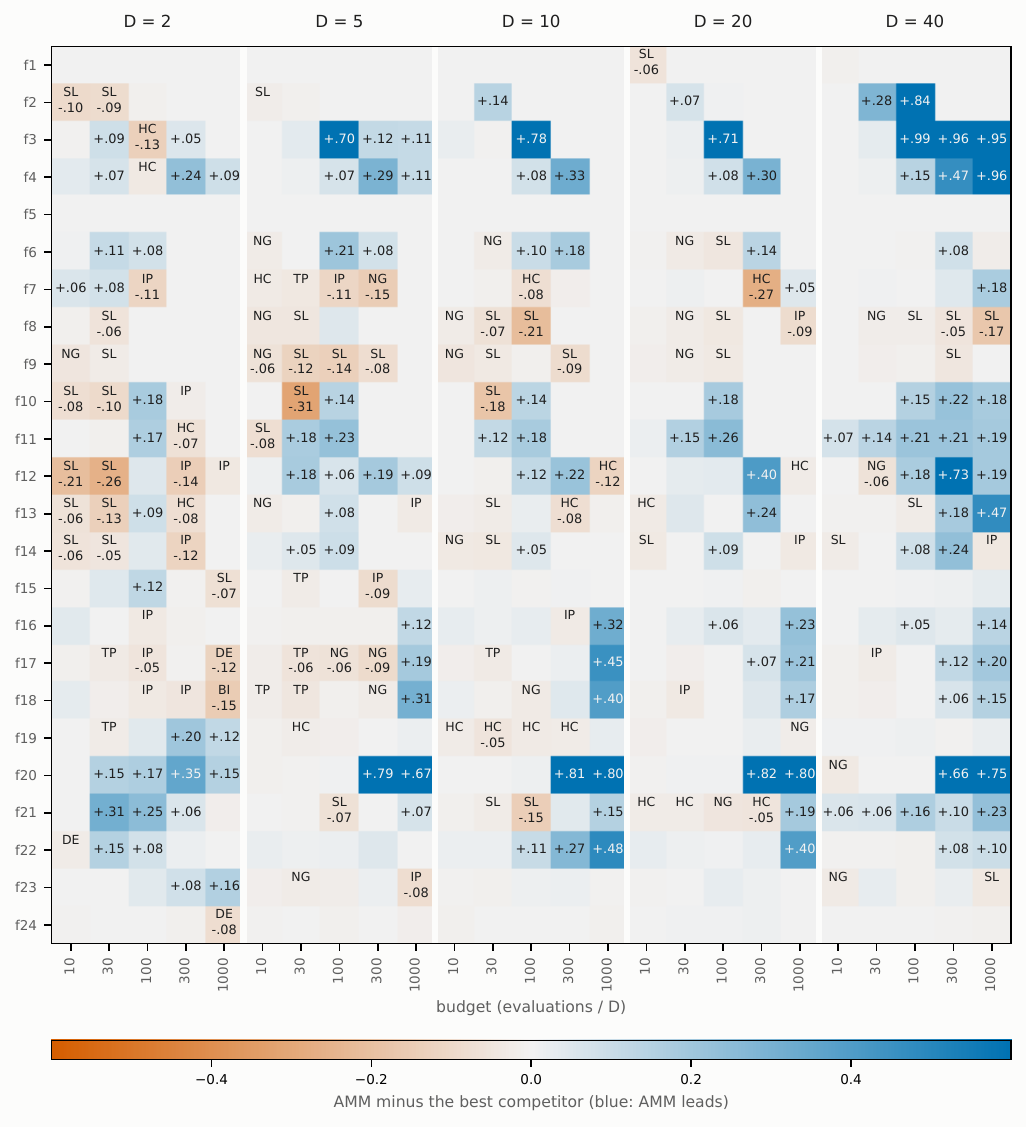}
  \caption{Final test (bbob instances 101--110, one run per instance, 10 per cell): AMM's target share minus the best competitor's per function, dimension and budget (\xD{10} to \xD{1000}); competitors as in Figure~\ref{fig:ecdf}, HCMA from its archive levels. Values of at least $.05$ are printed. Where AMM trails by more than $.02$ in mean (a descriptive threshold; the 35 credibly-behind cells of Section~\ref{sec:res-losses} use the posterior rule of Section~\ref{sec:heldout}), the leader is named: SL restarted SLSQP, HC HCMA's archive levels, NG NGOpt, IP and BI IPOP- and BIPOP-CMA-ES, TP TPE, DE differential evolution.}
  \Description[Per-cell lead or deficit against the best competitor (final test)]{A heatmap with the 24 bbob functions as rows and, side by side, the five dimensions 2, 5, 10, 20 and 40, each with the five budgets from 10 to 1000 times the dimension as columns. A diverging scale runs from a deficit to a lead. Differences of at least .05 are printed, and a two-letter code of the leading competitor is written where the method trails. Deficits concentrate at the smallest budgets and at a few functions.}
  \label{fig:lossmap}
\end{figure*}

\section{Data split and spent ids}\label{app:tables}

\begin{table}[t]
  \caption{Data used to design, validate and evaluate AMM (raw bbob instance ids). Each validation set served one confirmation and was then spent; Appendix~\ref{app:tables} (Spent ids) lists the ids of the ablation and the later analyses (770--869).}
  \Description[Data-split table: five rows from calibration to evaluation]{Four columns: role, instance ids, what the data were used for, and whether they were used during design. Calibration, development and the development field were used throughout design; validation sets once per composite; the evaluation data never.}
  \label{tab:split}
  \footnotesize
  \begin{tabular}{@{}p{0.17\columnwidth} p{0.22\columnwidth} p{0.28\columnwidth} p{0.13\columnwidth}@{}}
    \toprule
    Role & Data & Used for & In design? \\
    \midrule
    Calibration & ids 1--3 & probe scales, thresholds, traces & yes \\
    Development & ids 4, 5, 71, 72; 32 runs per cell & screening each candidate & yes \\
    Development field & ids 61--70; one seed & comparison with the field after each composite (in-sample) & yes \\
    Validation & fresh ids, e.g.\ 176--185, \dots, 730--769 & one pre-registered confirmation per composite & once per composite \\
    Evaluation & ids 101--110 (seed 9000); MA-BBOB rows 900--999 & the headline claims & no \\
    \bottomrule
  \end{tabular}
\end{table}

\paragraph{Spent ids.} Ids 1--100 and 116--869 are spent and 111--115 reserved (Table~\ref{tab:split}). Ids 610--729 confirmed c44--c46, 730--769 the rejected c47 and 770--809 the ablation; 810--869 went to later analyses, among them the route-firing replication on 810--829 (Appendix~\ref{app:firing}), the generalization runs on 810--849 (Appendix~\ref{app:general}) and the anytime mode's development on 850--869 (Appendix~\ref{app:anytime}). On 810--829 the firing and generalization runs are not independent, though the firing log scores no target share and neither analysis selected on the other.

\section{Official COCO ECDFs for all dimensions}\label{app:cocopp}
\begin{figure*}[tp]
  \centering
  \includegraphics[width=0.455\textwidth,alt={COCO runtime ECDF at dimension 2: fraction of function-target pairs reached against the logarithm of evaluations per dimension, for AMM at seven budgets and four archived algorithms.}]{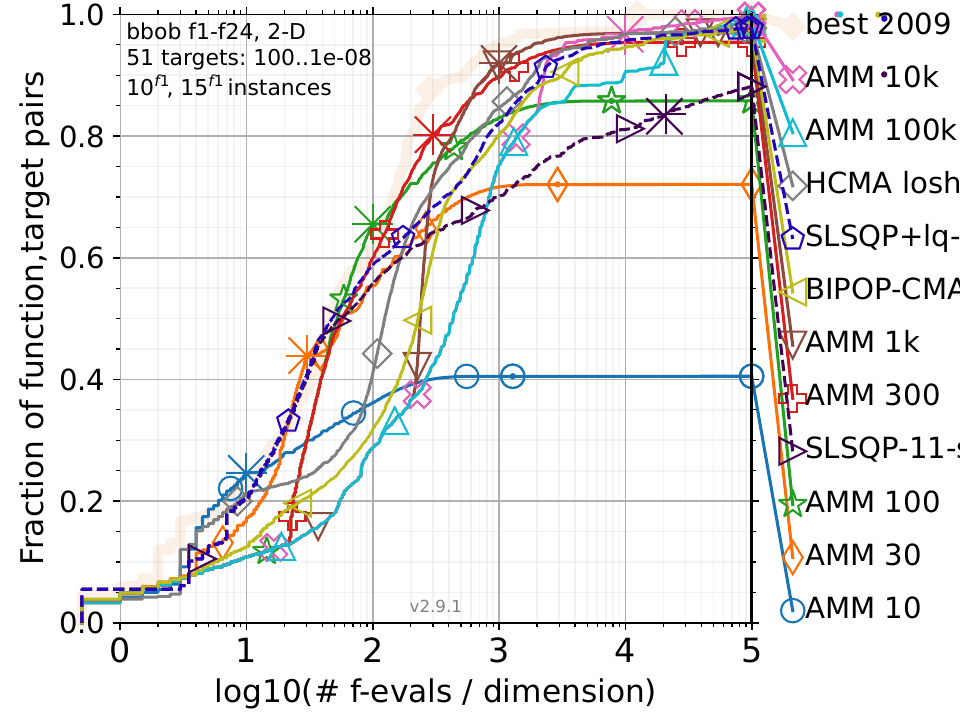}\hfill\includegraphics[width=0.455\textwidth,alt={COCO runtime ECDF at dimension 5: fraction of function-target pairs reached against the logarithm of evaluations per dimension, for AMM at seven budgets and four archived algorithms.}]{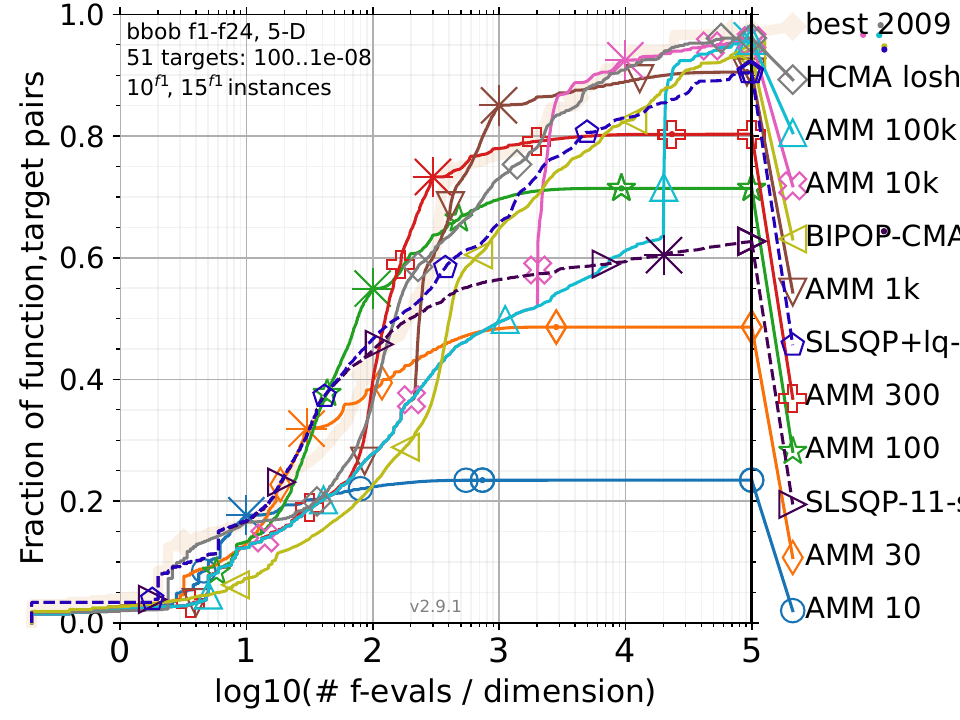}

  \includegraphics[width=0.455\textwidth,alt={COCO runtime ECDF at dimension 10: fraction of function-target pairs reached against the logarithm of evaluations per dimension, for AMM at seven budgets and four archived algorithms.}]{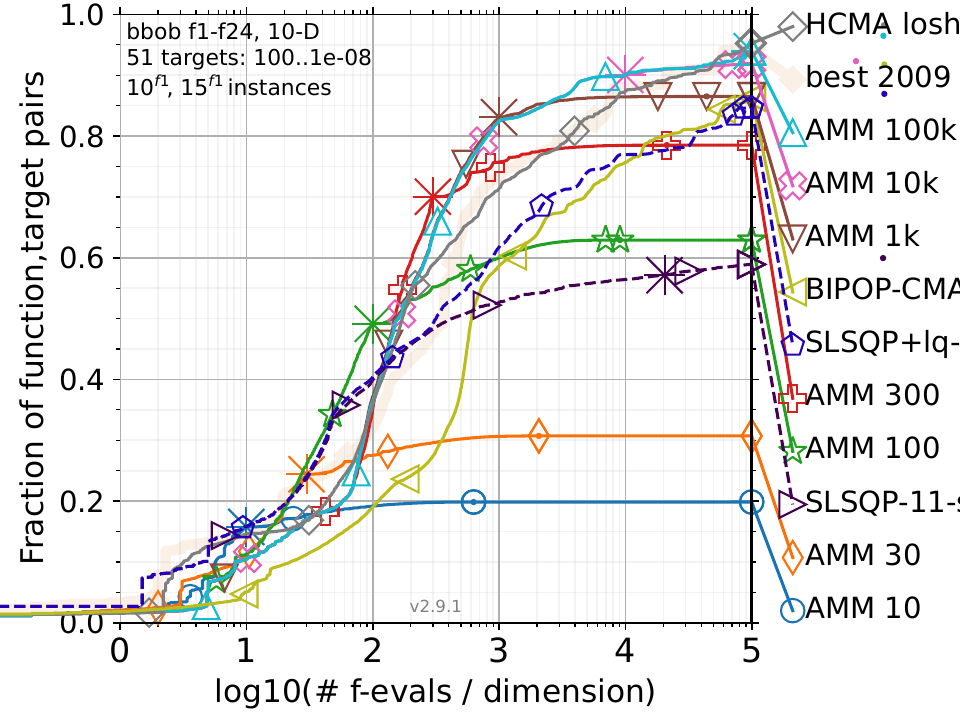}\hfill\includegraphics[width=0.455\textwidth,alt={COCO runtime ECDF at dimension 20: fraction of function-target pairs reached against the logarithm of evaluations per dimension, for AMM at seven budgets and four archived algorithms.}]{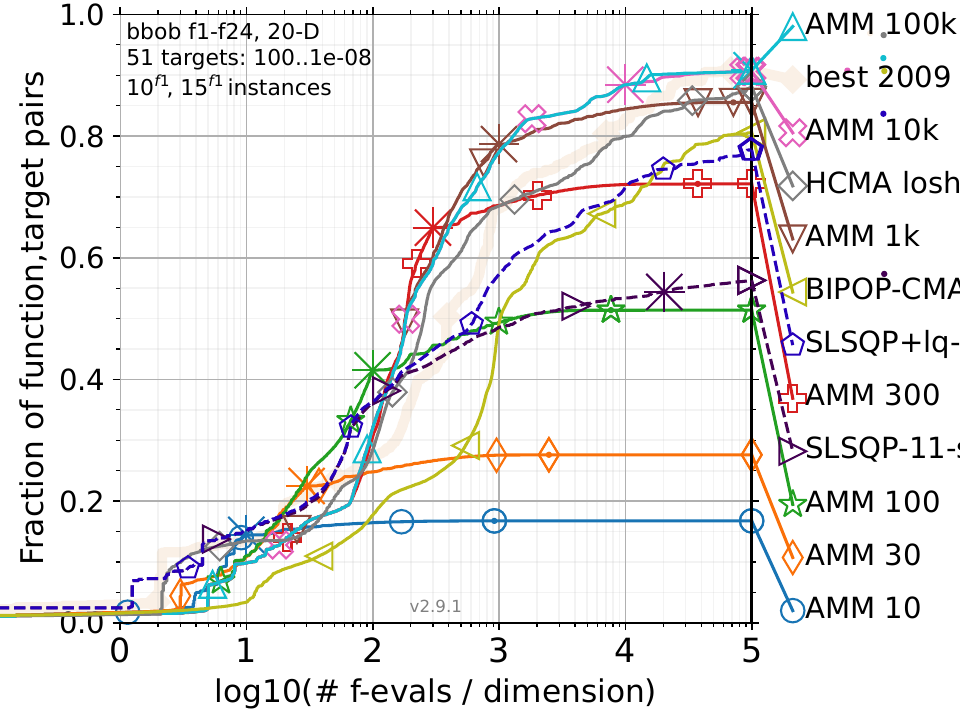}

  \includegraphics[width=0.455\textwidth,alt={COCO runtime ECDF at dimension 40: fraction of function-target pairs reached against the logarithm of evaluations per dimension, for AMM at seven budgets and four archived algorithms.}]{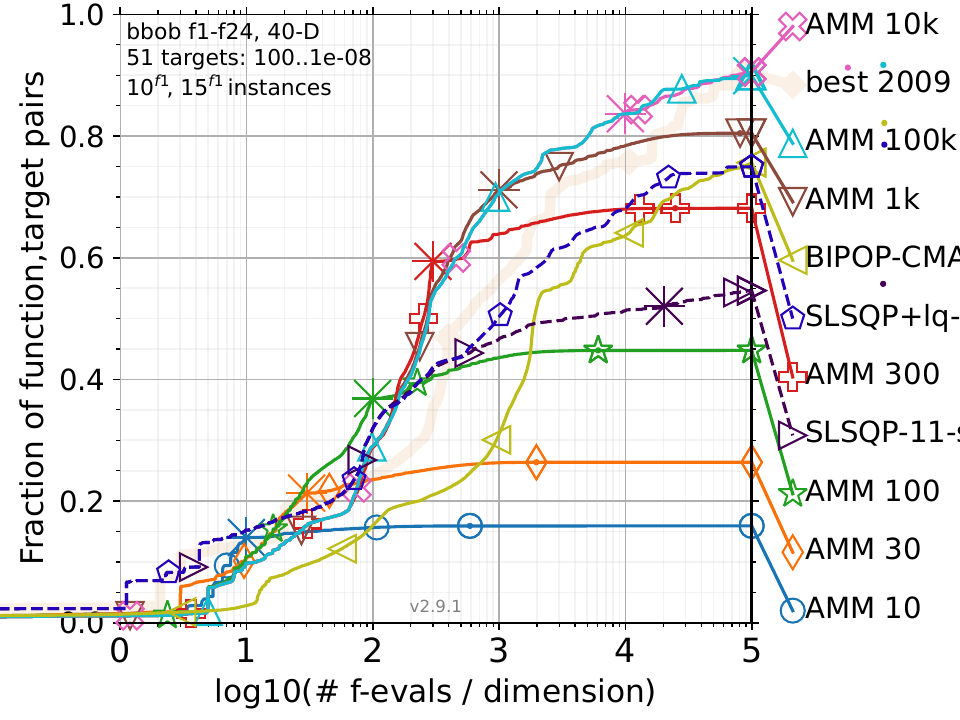}
  \caption{Official COCO runtime ECDFs (\texttt{cocopp} 2.9.1) of AMM on bbob f1--f24, 51 targets from $10^{2}$ to $10^{-8}$. Each AMM curve (``AMM $k$''; 1k, 10k and 100k are \xD{1000}, \xD{10000} and \xD{100000}) is a separate fixed-budget run with budget $kD$ through COCO's observer, on 10 instances (101--110). Beyond its budget, an AMM curve comes from cocopp's simulated restarts, so it is meaningful up to its budget. The \xD{10000} and \xD{100000} runs lie beyond the developed budgets and enter no headline claim; at $D=20$ and 40 they used an implementation speedup that changes no evaluation. The other curves are archive entries on their own 15 instances, so the comparison is not paired: HCMA (no archive data at $D=40$), BIPOP-CMA-ES, SLSQP+lq-CMA-ES and SLSQP; the pale curve is the best of the 2009 algorithms.}
  \Description[Official COCO runtime ECDFs at dimensions 2, 5, 10, 20 and 40]{Five line charts, two per row, of the fraction of function-target pairs reached, from 0 to 1, against the logarithm of the number of evaluations per dimension, from 0 to 5. Seven curves are AMM at budgets of 10, 30, 100, 300, 1000, 10000 and 100000 times the dimension; each rises up to its budget and continues, nearly flat, through cocopp's simulated restarts. The other curves are archived algorithms, which continue to 100000 times the dimension, and a pale curve for the best of the 2009 algorithms.}
  \label{fig:cocopp-all}
\end{figure*}

AMM's final-test runs were also recorded through COCO's observer, one dataset per budget, and every observer trace equals that of the harness run scored in Section~\ref{sec:res-final}; Figure~\ref{fig:cocopp-all} post-processes these recordings against archived algorithms. Figure~\ref{fig:cocopp-all} adds two budgets run after the final test, \xD{10000} and \xD{100000}, under their own pre-registration. Table~\ref{tab:largeb} gives AMM's mean shares of the 51 targets there.
\begin{table}[tp]
  \caption{AMM told its budget at \xD{1000} and at the two larger budgets run after the final test (instances 101--110): mean share of the 51 targets per dimension.}
  \Description[AMM's target share at 1000, 10000 and 100000 times the dimension]{Three rows, for budgets of 1000, 10000 and 100000 times the dimension, and five columns, for dimensions 2, 5, 10, 20 and 40; each cell is AMM's mean target share. The share grows with the budget in every dimension, from .712 to .899 at dimension 40.}
  \label{tab:largeb}
  \footnotesize
  \begin{tabular}{@{}l rrrrr@{}}
    \toprule
    Budget & $D=2$ & 5 & 10 & 20 & 40 \\
    \midrule
    \xD{1000} & .920 & .851 & .831 & .787 & .712 \\
    \xD{10000} & .967 & .925 & .900 & .884 & .837 \\
    \xD{100000} & .989 & .957 & .940 & .908 & .899 \\
    \bottomrule
  \end{tabular}
\end{table} Eight of the ten pre-registered ranges for these gains held (from \xD{1000} to \xD{10000}, $+.02$ to $+.10$; from \xD{10000} to \xD{100000}, $+.00$ to $+.06$); at $D=40$ both gains came out above them ($+.125$ and $+.062$). The predicted comparisons with the archive at these budgets were not computed.

\section{Ablation details}\label{app:ablation}
{} The pair test's STEP route pays only at \xD{100} (f3: $-.670$ there, about zero at \xD{300} and \xD{1000}). In the plateau and fractal family, the plateau route pays on f7 ($-.123$) and the IPOP routes on f15, f17 and f18 (about $-.02$ each). The code-identity check against the released code matched on 480 of 480 traces. With every family active, the switched code (the ablation build with its family switches) differed from the unmodified configuration on 7 runs in two of the 13 arms (3 and 4 of 240), all f7 at $D=5$ and \xD{300}, and only in the last ten evaluations of the restarted run; no score changed. The SLSQP stage gates also idle the kink trigger on f13 at $D=10$--20, a documented overlap of two families. Three A/A draws gave no rejection among 39 tests, and each slice's A/A yardstick (.0012 to .0132 across slices) is below that slice's $|\Delta|$.

\section{Route firing}\label{app:firing}
\begin{figure*}[tp]
  \centering
  \includegraphics[width=0.95\textwidth,alt={Twelve small heat-maps, one per route family, each with the 24 bbob functions as rows and the final test's 25 cells (dimensions 2, 5, 10, 20, 40, each with budgets 10, 30, 100, 300, 1000 times D) as columns; colour from white to blue is the share of runs, from 0 to 1, in which the route fires in the first call. The COBYQA gate is one blue column (D 2, 10 times D, every function); the STEP route fills f1 to f5 and the coordinate-search route f20 from 100 and 300 times D at D 5 and above; the ATM stage fills 300 and 1000 times D for most functions; the scaled BFGS descent appears only at 300 times D on the smooth functions; the hill-valley route is dark only for f21 and f22 at 300 and 1000 times D from D 5; the kink route is partial for f13; the plateau, rugged IPOP and fractal family is dark for f7 and, at 1000 times D and D 20 and 40, for f15 to f19, f23 and f24.}]{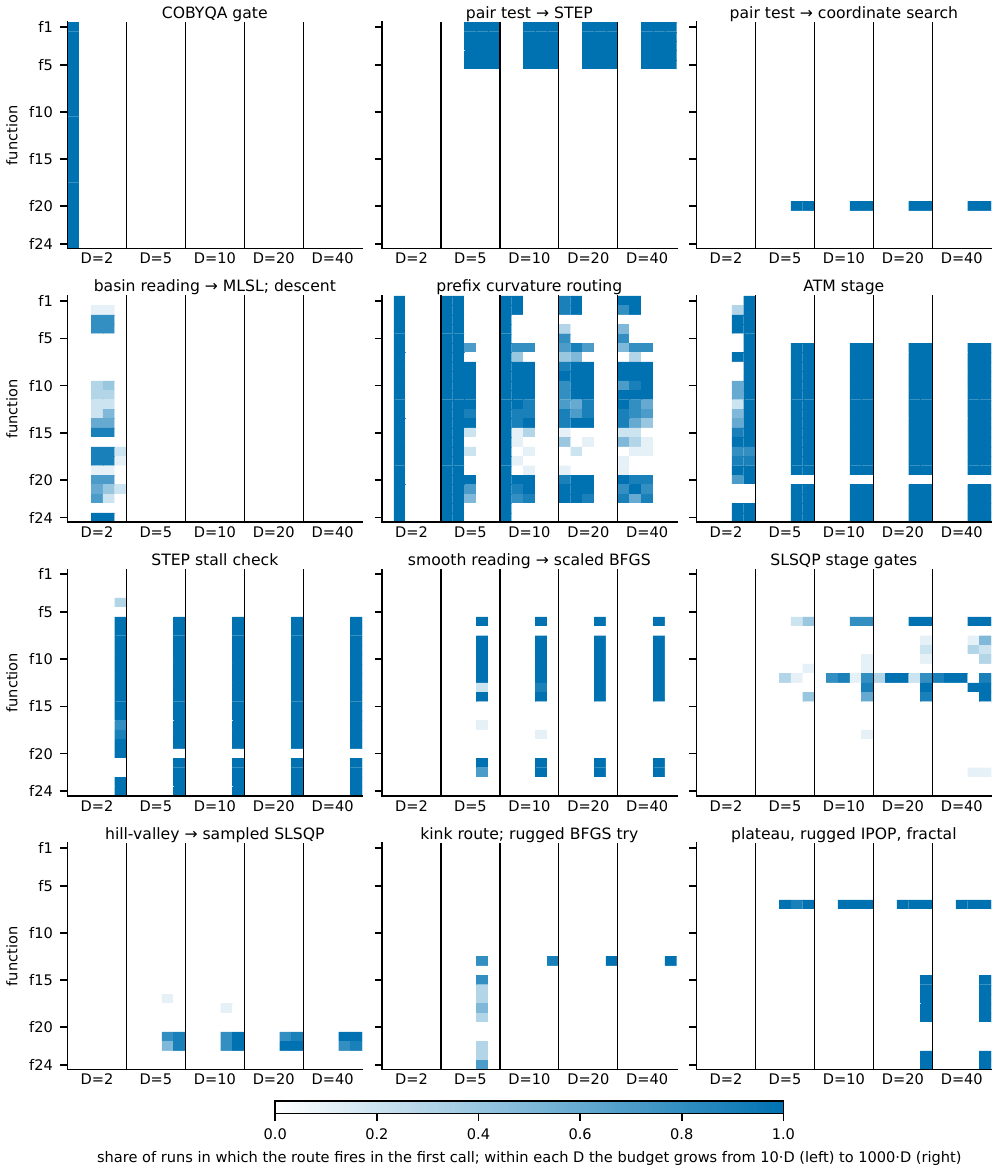}
  \caption{Where each route fires on the final test's grid (instances 101--110).}
  \Description[Route-firing heat-maps, final-test grid]{Twelve small heat-maps, one per route family, each with the 24 bbob functions as rows and the final test's 25 cells (dimensions 2, 5, 10, 20, 40, each with budgets 10, 30, 100, 300, 1000 times D) as columns; colour from white to blue is the share of runs, from 0 to 1, in which the route fires in the first call. The COBYQA gate is one blue column (D 2, 10 times D, every function); the STEP route fills f1 to f5 and the coordinate-search route f20 from 100 and 300 times D at D 5 and above; the ATM stage fills 300 and 1000 times D for most functions; the scaled BFGS descent appears only at 300 times D on the smooth functions; the hill-valley route is dark only for f21 and f22 at 300 and 1000 times D from D 5; the kink route is partial for f13; the plateau, rugged IPOP and fractal family is dark for f7 and, at 1000 times D and D 20 and 40, for f15 to f19, f23 and f24.}
  \label{fig:firing-a1}
\end{figure*}

\begin{figure*}[tp]
  \centering
  \includegraphics[width=0.95\textwidth,alt={The same twelve heat-maps for the fresh instances 810 to 829, with dimension 3 added and budgets up to 3000 times D, so 36 columns.}]{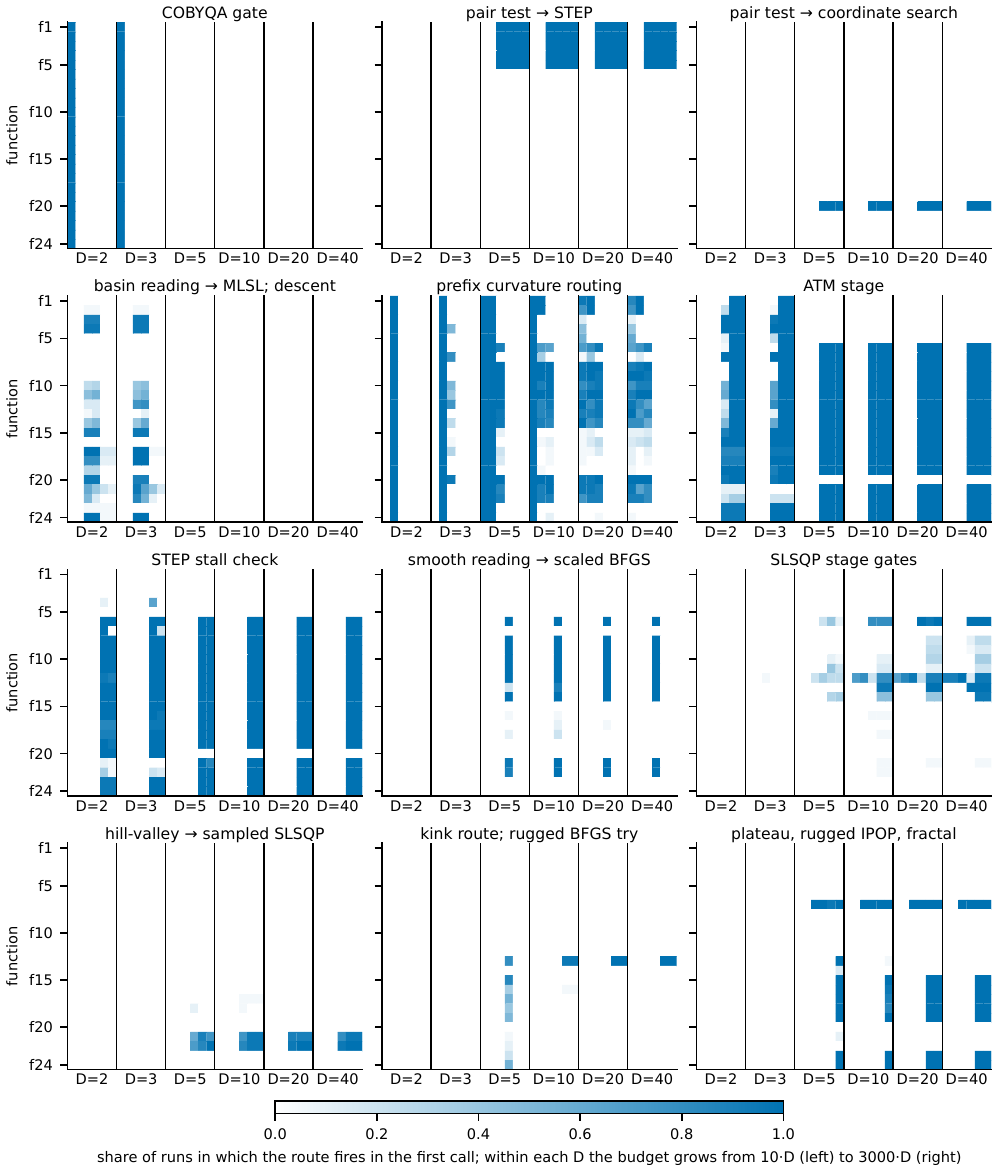}
  \caption{The same on fresh instances 810--829 (seed 9700), with $D=3$ and \xD{3000} added.}
  \Description[Route-firing heat-maps, fresh instances]{The same twelve heat-maps for the fresh instances 810 to 829, with dimension 3 added and budgets up to 3000 times D, so 36 columns.}
  \label{fig:firing-a2}
\end{figure*}

Route-firing logs record, for the first call of each run, whether each route fires; a route's firing share is the share of runs in which it fires, per function, dimension and budget. Figure~\ref{fig:firing-a1} covers the final test's grid (instances 101--110); its logged runs reproduce the recorded traces, so no target share is scored on them. Figure~\ref{fig:firing-a2} replicates it on fresh instances 810--829 with $D=3$ and \xD{3000} added, and 288 of 288 unlogged reruns on those fresh ids equal their logged records. Cell shares rest on 10 runs (final-test grid) or 20 (fresh grid), so a share has a standard error of up to $.16$ and $.11$. The released packages 0.3.0 and 0.5.0 reproduce all 6,000 final-test traces and all 6,000 route logs of Figure~\ref{fig:firing-a1} run for run, so the last column of Table~\ref{tab:params} and these figures describe the released code as well. The kink route reads f13 in 36\% of the runs in its range and the plateau route reads f7 in 59\%, so a route often misses the function that motivated it. Of the six rows that fire beyond their motivating functions (Section~\ref{sec:res-parts}), five do so on the fresh ids.

\section{Generalization runs in full}\label{app:general}
The generalization batch was pre-registered before any of its runs, with its claims, three blocks of arms, its analysis script and 17 predictions with numeric ranges, and approved by the author. It ran on fresh bbob instance ids 810--849 (seed 9000; one run per function, dimension, instance and budget; budgets of \xD{10}--\xD{1000}): 153 arms and 141,280 runs, none of which met a non-finite value. A run scores the share of the 51 targets its precision reaches; on a transformed problem, the precision is divided back by the factor $a$ of $f\mapsto af+c$. The analysis was run after every arm had finished. One deviation: the A/A arm of the transformed block was run, as pre-registered, with the seed shifted by 1,000, but the committed script paired runs by (instance, budget, seed). No A/A run therefore found its partner, and the script stopped before printing any statistic of that block. The A/A runs were re-keyed to their partner's seed and the script was run again, which reproduced the other two blocks exactly.

\paragraph{bbob-boxed (sbox-cost), ids 840--849, $D=2$--40.} HCMA's archive has no levels for these instances; TPE runs on the final test's capped slice. AMM has the highest share in 84, 78, 90, 99 and 100 of the 120 cells at \xD{10} to \xD{1000}. Of its 26 credibly-behind cells, SLSQP leads in 13 (at \xD{10}--\xD{100}, eight of them at $D=2$), IPOP- and BIPOP-CMA-ES in 9 (f7, f14, f17, f18), TPE in 3 and DE in 1. Table~\ref{tab:general} gives the pooled differences. AMM's lead over IPOP-CMA-ES is .142 smaller on the 19 functions whose optimum moved than on the 5 whose problems equal bbob's at \xD{10}, and .113 larger at \xD{100} (reported, as the pre-registration asks, because the difference exceeds .10).

\paragraph{Transformed bbob, ids 810--819, $D=2$--20.} The solver sees $y$ in the box $[-5s+b,5s+b]^D$ and $h(y)=af((y-b)/s)+c$, with $(a,c,s,b)$ equal to $(10^{3},-10^{5},1,0)$, $(10^{-3},100,1,0)$, $(1,0,0.1,3)$ and $(10^{3},-10^{5},0.1,3)$, beside the identity. AMM, IPOP-CMA-ES and SLSQP run every variant, and IPOP-CMA-ES and SLSQP receive the box through their bounds. The box variants test that AMM's internal map of any box onto $[-5,5]^D$ holds in the whole composite. Table~\ref{tab:general-t} gives each solver's change against the identity, and Table~\ref{tab:general} AMM's lead over IPOP-CMA-ES per variant. Under the affine maps AMM loses most on f10 and f11 ($-.240$ and $-.196$ under $10^{3}f-10^{5}$) and on f11 and f14 ($-.210$ and $-.205$ under $10^{-3}f+100$); under the box change alone no function loses more than .003.

\paragraph{Rotated f1--f4, ids 820--839, $D=2$--40.} Each problem is $f(x_{\mathrm{opt}}+R(x-x_{\mathrm{opt}}))$ with a seeded random orthogonal $R$ per function, dimension and instance; the optimum, its value and the box stay put. The same functions unrotated, on the same ids and seeds, are the comparison. f5 is not rotated, as the pre-registration states: its optimum is a corner of the box, where a rotation about the optimum has no canonical meaning. AMM minus IPOP-CMA-ES over the 25 (dimension, budget) cells, with intervals over the 20 ids, unrotated and rotated: f2 $+.437$ [$+.429$, $+.444$] and $+.390$ [$+.377$, $+.404$], f3 $+.505$ [$+.490$, $+.521$] and $+.029$ [$+.011$, $+.048$], f4 $+.301$ [$+.295$, $+.307$] and $+.011$ [$-.006$, $+.026$]. The rotation changes these leads by $-.047$, $-.476$ and $-.290$. Over rotated f2--f4, AMM leads IPOP-CMA-ES by $+.015$, $+.097$, $+.265$, $+.235$ and $+.105$ at \xD{10} to \xD{1000}, every interval above zero.

\begin{table*}[tp]
  \caption{Generalization runs: pooled differences in target share per budget $kD$, with 95\% bootstrap intervals over cells. Holm's correction runs over the five budgets for bbob-boxed and over the 20 tests of the five transformed rows (the identity included) at \xD{30}--\xD{1000}; every one is above zero. bbob-boxed: 120 (function, dimension) cells, ids 840--849, $D=2$--40. Transformed bbob: 96 cells, ids 810--819, $D=2$--20.}
  \Description[Generalization runs: AMM minus IPOP-CMA-ES and SLSQP per budget]{Seven rows and five budget columns from 10 to 1000 times the dimension. Two rows for bbob-boxed (AMM minus IPOP-CMA-ES and minus SLSQP) and five rows of AMM minus IPOP-CMA-ES on transformed bbob: the identity, two affine maps of the function values, a shrunk and shifted box, and the first map with the box. Every value is positive; the smallest lead over IPOP-CMA-ES is .063 at 10 times the dimension under the downward affine map.}
  \label{tab:general}
  \scriptsize
  \begin{tabular}{@{}l ccccc@{}}
    \toprule
    & \multicolumn{5}{c}{budget $kD$, $k=$} \\
    \cmidrule(l){2-6}
    & 10 & 30 & 100 & 300 & 1000 \\
    \midrule
    bbob-boxed, AMM minus IPOP-CMA-ES & $+.117$ [$+.078$, $+.162$] & $+.194$ [$+.145$, $+.245$] & $+.302$ [$+.244$, $+.360$] & $+.308$ [$+.247$, $+.369$] & $+.233$ [$+.183$, $+.292$] \\
    bbob-boxed, AMM minus SLSQP & $+.011$ [$+.002$, $+.019$] & $+.028$ [$+.012$, $+.043$] & $+.123$ [$+.087$, $+.164$] & $+.222$ [$+.176$, $+.275$] & $+.291$ [$+.236$, $+.354$] \\
    \midrule
    \multicolumn{6}{@{}l}{\emph{Transformed bbob, AMM minus IPOP-CMA-ES}} \\
    identity & $+.119$ [$+.074$, $+.171$] & $+.209$ [$+.153$, $+.268$] & $+.320$ [$+.254$, $+.387$] & $+.292$ [$+.230$, $+.359$] & $+.197$ [$+.143$, $+.261$] \\
    $f\mapsto10^{3}f-10^{5}$ & $+.111$ [$+.067$, $+.163$] & $+.145$ [$+.101$, $+.195$] & $+.270$ [$+.213$, $+.328$] & $+.280$ [$+.217$, $+.347$] & $+.175$ [$+.117$, $+.241$] \\
    $f\mapsto10^{-3}f+100$ & $+.063$ [$+.037$, $+.097$] & $+.142$ [$+.098$, $+.190$] & $+.253$ [$+.196$, $+.311$] & $+.261$ [$+.195$, $+.331$] & $+.149$ [$+.100$, $+.208$] \\
    box $[2.5,3.5]^D$ & $+.121$ [$+.078$, $+.172$] & $+.211$ [$+.155$, $+.270$] & $+.324$ [$+.260$, $+.392$] & $+.328$ [$+.262$, $+.398$] & $+.248$ [$+.192$, $+.310$] \\
    $10^{3}f-10^{5}$ and the box & $+.114$ [$+.070$, $+.165$] & $+.147$ [$+.102$, $+.197$] & $+.274$ [$+.217$, $+.333$] & $+.318$ [$+.251$, $+.388$] & $+.224$ [$+.167$, $+.287$] \\
    \bottomrule
  \end{tabular}
\end{table*}

\begin{table*}[tp]
  \caption{Transformed bbob (ids 810--819, $D=2$--20): change of each solver's target share against the identity problem, paired by run, mean over all cells and budgets with a 95\% bootstrap interval over the 10 ids. The A/A row reruns the identity with another seed.}
  \Description[Change of each solver's target share under each transformation]{Five rows (A/A, two affine maps, the box change, the combination) and three columns (AMM, IPOP-CMA-ES, SLSQP) of mean changes with intervals. AMM loses .03 to .05 under the affine maps and nothing under the box change; IPOP-CMA-ES is unchanged under the affine maps and loses .018 under the box changes; SLSQP loses under every change.}
  \label{tab:general-t}
  \scriptsize
  \begin{tabular}{@{}l ccc@{}}
    \toprule
    Change & AMM & IPOP-CMA-ES & SLSQP \\
    \midrule
    A/A (seed $+1000$) & $-.002$ [$-.012$, $+.006$] & $+.002$ [$-.004$, $+.009$] & $-.002$ [$-.008$, $+.005$] \\
    $f\mapsto10^{3}f-10^{5}$ & $-.032$ [$-.036$, $-.028$] & $-.000$ [$-.001$, $-.000$] & $-.067$ [$-.072$, $-.062$] \\
    $f\mapsto10^{-3}f+100$ & $-.053$ [$-.058$, $-.049$] & $+.001$ [$+.000$, $+.001$] & $-.043$ [$-.048$, $-.037$] \\
    box $[2.5,3.5]^D$ & $+.001$ [$-.000$, $+.002$] & $-.018$ [$-.022$, $-.013$] & $-.045$ [$-.051$, $-.040$] \\
    both & $-.030$ [$-.035$, $-.027$] & $-.018$ [$-.023$, $-.013$] & $-.124$ [$-.127$, $-.121$] \\
    \bottomrule
  \end{tabular}
\end{table*}

\section{The anytime mode in full}\label{app:anytime}
\paragraph{The mode.} The anytime mode, added after the final test, is not part of the benchmarked preset \texttt{c46} (Section~\ref{sec:res-anytime} evaluates it). In its anytime mode AMM first plans \xD{10} evaluations and, each time a plan is spent while the run goes on, plans twice as many in total (\xD{20}, \xD{40}, \dots). Each new plan reruns the configuration from the same start point and seed and answers every point evaluated before from memory, without a call of $f$. Once the next plan would exceed $10^4$ evaluations, one ordinary run planned for the remaining budget takes over, with the points evaluated so far as its archive. Until then the mode never reads the budget.

\paragraph{On the final-test instances.} The mode was developed on calibration ids 1--3 and fresh ids 850--869, never on instances 101--110. On the final-test problems all seven ranges of its own pre-registration held. It is level with AMM told its budget at \xD{10^4} and \xD{10^5} (Table~\ref{tab:anytime}; every dimension within $\pm .01$), where every run hands over to an ordinary run after its last replayed plan. Before the hand-over it pays for not knowing the budget, most at \xD{300} for $D \ge 5$ ($-.13$ to $-.16$), because AMM's plan changes at almost every doubling and a stage rebuilds little of the smaller plan's state; the \xD{1000} runs at $D\ge10$ and the \xD{300} runs at $D=40$ already hand over.
\begin{table*}[tp]
  \caption{Anytime mode against AMM told its budget, on the final-test instances 101--110 (seed 9000, $D=2$, 5, 10, 20, 40, f1--f24): mean target share per budget $kD$ and the paired difference (anytime minus budget known) with its 95\% cluster-bootstrap interval. The two largest budgets lie beyond the developed range.}
  \Description[Anytime mode against AMM told its budget, per budget]{A table with seven budgets from 10 to 100000 times the dimension; for each, the mean target share of the anytime mode, of AMM told its budget, and their paired difference with a confidence interval. The difference is near zero at 10, 10000 and 100000 times the dimension and negative, down to minus .124, between 30 and 1000 times the dimension.}
  \label{tab:anytime}
  \footnotesize
  \begin{tabular}{@{}l rrrrrrr@{}}
    \toprule
    Budget $kD$, $k=$ & 10 & 30 & 100 & 300 & 1000 & $10^4$ & $10^5$ \\
    \midrule
    anytime mode & .171 & .234 & .418 & .571 & .772 & .903 & .940 \\
    AMM, budget known & .173 & .288 & .496 & .696 & .820 & .903 & .939 \\
    difference & $-.002$ & $-.054$ & $-.078$ & $-.124$ & $-.048$ & $+.001$ & $+.001$ \\
    95\% interval & [$-.004$, $+.001$] & [$-.064$, $-.045$] & [$-.093$, $-.065$] & [$-.140$, $-.109$] & [$-.059$, $-.037$] & [$-.006$, $+.007$] & [$-.004$, $+.006$] \\
    \bottomrule
  \end{tabular}
\end{table*}

\paragraph{Replay and hand-over.} Each stage replays the smaller plan exactly up to the first decision the larger budget changes. The last replayed plan has 6,400 evaluations from $D=5$ and 5,120 at $D=2$. At the hand-over, the points evaluated so far replace the fresh points that the ordinary run's phases would draw: the prefix design, the sampler's first population, CMA-ES's first covariance and the restarted SLSQP's samples. A point the run asks for again is answered from memory.

\paragraph{Why state is recomputed.} With the current components, engine state cannot be kept and continued as the plan grows: SciPy's local searches expose neither a pause nor their internal state, every phase fixes its budget share and route when it starts, and the route itself changes with the budget. On a calibration problem (id 1), replay rebuilds 0--47\% of each stage.

\paragraph{Development tests of handing state over.} On fresh ids 850--869, handing that state to the larger plan did not close the gap to AMM told its budget. An archive warm start gained $+.016$ over replay at \xD{30} but lost .050 and .063 at \xD{100} and \xD{1000}; nesting the larger plans on a replayed head lost .026 and .041 at \xD{300} and \xD{1000}.

\paragraph{Route firing.} Route-firing logs of the anytime mode on the final-test grid (release 0.5.0, instances 101--110, descriptive; a run is read at its hand-over call if it has one, otherwise at its last stage) show the gates that need a large plan opening later. The pair test never opens at \xD{100}, where the last plan is \xD{80}, and at \xD{300} it opens on f3 and f4 only, because f1, f2 and f5 reach the final target in an earlier stage; the coordinate-search route opens only at \xD{1000}; the STEP stall check, the rugged IPOP and fractal routes and the rugged BFGS try reach a pooled share of .5 on no function. Three of the twelve function-motivated rows of Table~\ref{tab:params} fire beyond the functions that motivated them, against six for AMM. This fits the reason given above for the mode's loss at \xD{30}--\xD{1000} but does not prove it. Of the pre-registered firing predictions, 6 of 11 held, 2 fell outside their ranges and 3 were falsified: they ignored that the harness restarts a call that returns early and ends a run at the final target (28\% of the anytime runs).

\paragraph{Compute and use.} The mode keeps a run's state by recomputing it, not by storing it: no evaluation of $f$ is repeated, but the optimizer's own work is. This multiplies AMM's CPU time per evaluation at \xD{10}--\xD{1000}. It is for deterministic objectives whose budget is not known in advance or whose run may be stopped early. Until the hand-over a run stopped at any time holds the end of a complete smaller plan plus the best of the earlier ones: the result of AMM told a budget of between half and all of the evaluations spent. It should not be used when the budget is known and lies between about \xD{30} and \xD{1000}, where AMM told its budget is $.05$--$.12$ better, or when $f$ is cheap. Nor should it be used with noisy objectives, since a remembered value stands in for a new evaluation, or when the run may be stopped after the hand-over, since from then on it is an ordinary run planned for the remaining budget.

\section{Probes at their defaults}\label{app:p0}
A pre-registered control (fresh ids 880--899, seed 9000, $D=2$--40, f1--f24, \xD{10}--\xD{1000}, 12,000 runs per arm) runs AMM with every routing probe forced to the branch in Table~\ref{tab:params}'s ``Default'' column, without spending its evaluations: the same engine pool under the same budget schedule, minus every decision taken from $f$. Table~\ref{tab:p0} gives the mean target shares.

\begin{table*}[tp]
  \caption{Probes at their defaults: mean target share per arm and budget, and the difference with its 95\% interval (fresh ids 880--899, $D=2$--40, f1--f24).}
  \Description[Probes at their defaults]{A table of mean target share for six arms at five budgets from 10 to 1000 times the dimension, and the difference between AMM and AMM with every probe at its default, which is positive at every budget and largest at 300 times the dimension.}
  \label{tab:p0}
  \centering\footnotesize
\begin{tabular}{@{}l rrrrr@{}}
\toprule
Budget $kD$, $k=$ & 10 & 30 & 100 & 300 & 1000 \\
\midrule
AMM & .173 & .286 & .487 & .693 & .812 \\
every probe at its default & .151 & .243 & .328 & .404 & .591 \\
AMM, second seed (A/A) & .169 & .286 & .494 & .694 & .818 \\
restarted SLSQP alone & .157 & .251 & .381 & .468 & .518 \\
IPOP-CMA-ES alone & .052 & .087 & .192 & .384 & .593 \\
coordinate search alone & .036 & .168 & .202 & .328 & .379 \\
\midrule
AMM minus probes at default & $+.022$ & $+.043$ & $+.160$ & $+.289$ & $+.221$ \\
95\% interval & [$+.009$, $+.038$] & [$+.019$, $+.073$] & [$+.113$, $+.213$] & [$+.232$, $+.350$] & [$+.166$, $+.277$] \\
\bottomrule
\end{tabular}
\end{table*}

Every difference is above zero under Holm over the five budgets, and it accounts for at least all of AMM's lead over the better of SLSQP and IPOP-CMA-ES alone (ratio 1.0 to 1.5): with its probes at their defaults, the engines on AMM's schedule trail SLSQP alone by $.053$ and $.064$ at \xD{100} and \xD{300} and are level with IPOP-CMA-ES at \xD{1000}. The value sits on f3, f20, f21--f22 and f6--f14, and on f2, f4 and f5 through the STEP and coordinate routes; it is about zero on f15, f19, f23 and f24. Three caveats: the control is one definition of ``every probe at its default'' (a hand-built schedule without routing is the greedy-schedule rung of Section~\ref{sec:res-ladder}); at \xD{10}, $D=5$--20, switching the small-budget probe off also removes a COBYQA stage that no reading chooses, so part of the \xD{10} difference is not the probe's; and the best engine alone was chosen on these test ids, which favours the engine. One crash guard was added during the runs (a model fit of the (1+1)-CMA-ES returns no step on a singular matrix, a path AMM itself never reached); AMM's traces were unchanged by it. Of the control's 25 pre-registered ranges, 17 held. Six of the eight misses favour the probes: at \xD{300} and \xD{1000}, the difference to the default arm and its ratio to AMM's lead over the better engine alone came out above their ranges, and the default arm's gap to that engine below them (the pre-registration expected the engines on AMM's schedule to beat IPOP-CMA-ES at \xD{1000}; they are level, .591 against .593). The other two are A/A means ($+.007$ and $+.006$ at \xD{100} and \xD{1000}) outside the predicted $\pm.005$.

\section{Dimensions and budgets off the development grid}\label{app:unseen}
AMM as benchmarked, pycma's IPOP-CMA-ES and restarted SciPy SLSQP ran at $D=4$, 7, 15 and 30 (all nine budgets \xD{10}, 20, 30, 100, 200, 300, 350, 600 and 1000) and at the final test's dimensions with the new budgets \xD{20}, 200, 350 and 600, on fresh ids 870--879 (seed 9000, f1--f24; 58,320 runs; pre-registered). COCO's suite does not build $D=4$, 7, 15 or 30, so those dimensions use COCO's bare problems; at $D=5$ and 10 their $f$-values are bit-equal to the suite's. Table~\ref{tab:unseen} gives the margins.

\begin{table*}[tp]
  \caption{Margins off the development grid: AMM minus IPOP-CMA-ES and minus restarted SLSQP, pooled over $D=4$, 7, 15 and 30 (96 cells per budget), and minus IPOP-CMA-ES at $D=4$ alone (fresh ids 870--879).}
  \Description[Margins off the development grid]{A table of the mean margin of AMM over IPOP-CMA-ES and over restarted SLSQP at nine budgets from 10 to 1000 times the dimension, pooled over the unseen dimensions 4, 7, 15 and 30, and over IPOP-CMA-ES at dimension 4 alone; every margin is positive and the dimension-4 row is smallest at 300 to 600 times the dimension.}
  \label{tab:unseen}
  \centering\footnotesize
\begin{tabular}{@{}l rrrrrrrrr@{}}
\toprule
$k$ (\xD{k}) & 10 & 20 & 30 & 100 & 200 & 300 & 350 & 600 & 1000 \\
\midrule
AMM $-$ IPOP-CMA-ES & $+.118$ & $+.157$ & $+.195$ & $+.295$ & $+.267$ & $+.295$ & $+.268$ & $+.242$ & $+.208$ \\
AMM $-$ SLSQP & $+.018$ & $+.023$ & $+.043$ & $+.104$ & $+.117$ & $+.202$ & $+.201$ & $+.244$ & $+.293$ \\
$D=4$ only, $-$ IPOP-CMA-ES & $+.120$ & $+.191$ & $+.243$ & $+.282$ & $+.146$ & $+.078$ & $+.076$ & $+.090$ & $+.109$ \\
\bottomrule
\end{tabular}
\end{table*}

Every pooled margin is above zero under Holm (13 comparisons per competitor, the four new budgets at the old dimensions included: $+.160$ to $+.272$ over IPOP-CMA-ES, $+.021$ to $+.251$ over SLSQP). The margin against SLSQP at \xD{200} for the new dimensions ($+.117$) fell below its predicted range, with the sign held; every other prediction held (43 of 44). At $D=4$ the margin over IPOP-CMA-ES has intervals reaching zero at \xD{300}, \xD{350} and \xD{600}, the only such settings of all 81, and $D=4$ holds 16 of the 54 cells in which AMM is credibly behind IPOP-CMA-ES (f7, f8, f13, f14, f16, f17 and f18); whether a gap in the $D$-keyed gates causes it was not tested. The grid-fit index, the margin at the unseen settings minus the interpolation of this batch's own grid points, is $-.018$ [$-.031$, $-.005$] overall and $-.060$ [$-.099$, $-.026$] at $D=4$; both are inside their predicted ranges. On the fresh instances the final test's margins over IPOP-CMA-ES and SLSQP replicate within $.014$ at all five budgets.

\section{An exploratory check on other functions}\label{app:fresh}
After the final test, an unregistered check ran the released package (version 0.5.0, which reproduces every final-test run of the benchmarked configuration bit for bit; see Software availability) on 20 functions that are not bbob functions. The functions were written for the check by the AI assistant (Claude), among them jump, Cauchy-loss, ripple, needle, lattice and pinball-loss functions. Each ran at $D=2$, 3 and 4 on 12 instances with a random shift, scale and offset (and a random rotation where the function defines one), at budgets of \xD{30}--\xD{1000}, from scratch per budget, against restarted SLSQP and COBYQA, IPOP-CMA-ES, differential evolution, NADE and ERADS. It used its own driver, not the harness of Section~\ref{sec:design}, and scored each run by the log$_{10}$ of its error relative to the median error of random points, floored at $-10$. Over the 720 problem instances per budget, AMM's mean error was lower than every other solver's at \xD{100}; at \xD{30} it was level with restarted SLSQP ($0.00$ [$-0.24$, $+0.25$] decades) and behind the best other solver by more than half a decade in 20 of the 60 function--dimension cells. At \xD{300} and \xD{1000} it was level with IPOP-CMA-ES ($-0.15$ [$-0.36$, $+0.06$] and $-0.10$ [$-0.28$, $+0.07$] decades), and at \xD{1000} it was behind NADE, which trails it on bbob (Appendix~\ref{app:winners}), by $0.43$ [$0.28$, $0.59$] decades. No range was predicted in advance, the dimensions lie at the bottom of the final test's range, and the result is reported for completeness, not as a transfer test.

\csname @ACM@balancefalse\endcsname

\end{document}